\documentclass{article}

     \PassOptionsToPackage{numbers, compress}{natbib}

     \usepackage[preprint]{neurips_2025}

\usepackage[utf8]{inputenc} 
\usepackage[T1]{fontenc}    
\usepackage{url}            
\usepackage{booktabs}       
\usepackage{amsfonts}       
\usepackage{nicefrac}       
\usepackage{microtype}      
\usepackage{xcolor}         
\usepackage{graphicx}
\usepackage{subfigure}
\usepackage{multirow}       
\usepackage{algorithmic}
\usepackage{makecell}
\usepackage[linesnumbered,ruled,vlined]{algorithm2e}
\usepackage{diagbox}

\usepackage{hyperref}       

\usepackage{amsmath}
\usepackage{amssymb}
\usepackage{mathtools}
\usepackage{amsthm}

\usepackage[capitalize,noabbrev]{cleveref}

\theoremstyle{plain}
\newtheorem{theorem}{Theorem}[section]

\theoremstyle{definition}
\newtheorem{definition}[theorem]{Definition}
\newtheorem{assumption}[theorem]{Assumption}
\theoremstyle{remark}
\newtheorem{remark}[theorem]{Remark}

\title{Adverseness vs. Equilibrium: Exploring Graph Adversarial Resilience through Dynamic Equilibrium}

\author{%
 Xinxin Fan$^{1,2}$ \quad Wenxiong Chen$^{1,3}$ \quad Mengfan Li$^{1,2}$ \quad Wenqi Wei$^4$ \quad Ling Liu$^5$ \\
$^1$Institute of Computing Technology, Chinese Academy of Sciences \\
\quad $^2$University of the Chinese Academy of Sciences 
\quad $^3$Dalian University of Technology \\
\quad $^4$Fordham University 
\quad $^5$Georgia Institute of Technology\\
\texttt{\{fanxinxin,limengfan22z\}@ict.ac.cn}\\
\texttt{cwx1581015@mail.dlut.edu.cn} \quad \texttt{wenqiwei@fordham.edu} \quad \texttt{lingliu@cc.gatech.edu}
}

\begin{document}

\maketitle

\begin{abstract}
 Adversarial attacks to graph analytics are gaining increased attention. To date, two lines of countermeasures have been proposed to resist various graph adversarial attacks from the perspectives of either graph per se or graph neural networks. Nevertheless, a fundamental question lies in whether there exists an intrinsic adversarial resilience state within a graph regime and how to find out such a critical state if exists. This paper contributes to tackle the above research questions from three unique perspectives: i) we regard the process of adversarial learning on graph as a complex multi-object dynamic system, and model the behavior of adversarial attack; ii) we propose a generalized theoretical framework to show the existence of critical adversarial resilience state; and iii) we develop a condensed one-dimensional function to capture the dynamic variation of graph regime under perturbations, and pinpoint the critical state through solving the equilibrium point of dynamic system. Multi-facet experiments are conducted to show our proposed approach can significantly outperform the state-of-the-art defense methods under five commonly-used real-world datasets and three representative attacks. 
\end{abstract}

\section{Introduction} \label{Intro}
Graph adversarial attacks (GAAs) are launched through executing subtle perturbations on edges, weights and even features to the original clean graph \cite{MuWang21}, such as injecting/removing edges/nodes, or even modifying features/weights. To date, two main categories of GAAs emerges, one is injection attack \cite{WangLuo20,ZhengFei20,ZouZheng21}, i.e. inject new nodes into the original graph rather than straightly revise the existing characteristics. The other is modification attack \cite{Daniel18, Daniel19, Waniek16, SunWang20}, i.e. directly change the original graph in terms of edges/nodes and/or features/weights. 

Accordingly, to defend against GAAs, a set of adversarial defense methods are proposed from two perspectives: graph per se and graph neural networks. For the former, preprocessing countermeasures are utilized, for instance, GCN-SVD \cite{Entezari20} utilizes singular value decomposition (SVD) to decompose the adjacency matrix of perturbed graph and obtains a low-rank approximation matrix with the purpose of cleaning perturbation. GCN-Jaccard \cite{WuChen19} removes those edges that connected nodes have low Jaccard similarity of features. Analogously, GNNGUARD \cite{ZhangZitnik20} assigns higher weights to edges over similar nodes, but prunes edges over dissimilar nodes. Moreover, there also exist optimization countermeasures, e.g., Pro-GNN \cite{JinMa20} treats adjacency matrix as learning parameters to implement optimization manipulation under the requirements of being close to the original adjacency matrix, low-rank of learned adjacency matrix and feature smoothness. Referring to the variance-based attention, RGCN \cite{ZhuZhang19} dispenses differential weights to distinct neighbors during convolution. Additionally, towards the inductive graph-learning tasks, Wang et al. \cite{WangHuai23} propose secure graph-learning mechanism in terms of graph partitioning with fairness and balance. These methods implement adversarial defense through handling graph structure, i.e. graph per se.

For the latter, the neural ordinary differential equation (ODE) \cite{ChenRubanova18} and partial differential equation (PDE) \cite{Chamberlain21,ChamberlainJames21} have been attempted to defend against GAAs. Song et al. \cite{SongKang22} state that graph PDE can validly resist topology perturbation by heat-diffusion model on a general Riemannian manifold. Zhao et al. \cite{ZhaoKang23} propose an energy-conservation graph Hamiltonian flow to promote adversarial resilience. This line of research tries to adopt diffusion models in physics to conduct graph neural flow w.r.t. the attack resilience, and generates time-evolved embedding vectors for graph nodes, then utilizes backpropagation to minimize loss function for the parameters optimization. 

Upon the two lines of studies above, we can summarize a few highlights: i) GAAs are usually launched through node/edge revision and/or weight/feature modification; ii) towards the graph per se, current defense methods almost concentrate on low-rank approximation of adjacency matrix of the perturbed graph, deletion of edges over dissimilar nodes, smoothness of node features, bias of assigned weights; and iii) towards neural networks, ODE and PDE are employed to infer robust embedding vectors to represent nodes. Our work falls into the category of graph per se. 

We believe the inherent property of graph regime is the most influential factor for graph robustness, only a proper graph structure is built, can the adversarial resilience of graph regime exist. Then, a fundamental research question appears, that is, whether there exists a critical state of adversarial resilience within each graph regime, if yes, how to find out it. To surround this basic question, we in theory explore a generalized \textbf{Equili}brium point-based adversarial \textbf{Res}ilience approach \textbf{EquiliRes} to resist various graph adversarial attacks. 
In a nutshell, four main contributions are involved:

\begin{itemize}
	\item A hypothesis is put forward to bridge the connection between adversarial (deep) learning and complex dynamic system, by which GAAs can be modeled appropriately. 
	\item A generalized theoretic framework is proposed to showcase the existence of critical adversarial resilience state of graph regime, and explicitly exhibit in what prerequisites and in what perturbation domains the equilibrium point can be located.  
	\item A condensed one-dimensional function is developed to demonstrate the dynamics of graph under adversarial perturbations, 
	upon which an attack-resilient graph topology is built.   
	\item Multi-facet experiments are executed to validate the effectiveness of our proposed defense approach and the rationality of our proposed hypothesis using five real-world datasets. 
\end{itemize}


\section{Theoretical framework for graph adversarial resilience} \label{Sec: TheoreticalFramework}
\subsection{Adversarial resilience hypothesis} \label{Sec: ResiAssume}
\begin{figure}[tbhp] 
	\centering
	\includegraphics[height=1.65in, width=4.6in]{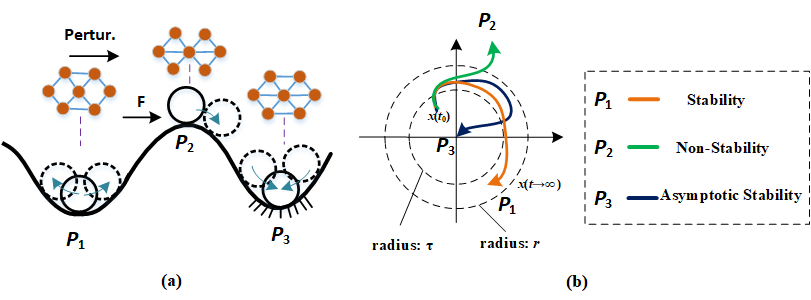}
	\caption{Three balls' motion states and equilibrium points.}
	\label{fig:BallPerturbation}
\end{figure}

We first utilize a vanilla physical phenomenon to present the intuitive understanding as shown in Fig. \ref{fig:BallPerturbation}(a), wherein three balls settle on three positions in a curved orbit. The difference between position \textbf{$P_1$} and position \textbf{$P_3$} is that the former has no friction while the latter has. At first, the three balls are settled in still, i.e. mathematically \(\left. {\frac{{dp(t)}}{{dt}}} \right|_{t = 0}\) =0, it means the first-order derivative of displacement $p(t)$ is zero, that is to say, the balls would keep stationary as the time $t$ changes if no extra action exists, these three positions are what is called equilibrium points. Nevertheless, when a force $F$ acts on the three balls, for position \textbf{$P_1$} the ball would move back-and-forth repeatedly in a bounded range; for position \textbf{$P_2$} the ball would move away and cannot go back any more; for position \textbf{$P_3$} the ball would first move back-and-forth and finally become still due to the action of friction. Inspired by this phenomenon, we inspect whether a graph regime also has such equilibrium point(s) representing its adversarial resilience state under different-level adversarial perturbations (attacks). 

\begin{center}
\fcolorbox{black}{gray!10}{\parbox{1.0\linewidth}{
\begin{remark}
The three balls' states respectively correspond to \textbf{Stability}, \textbf{Non-Stability}, and \textbf{Asymptotic Stability} as shown in Fig. \ref{fig:BallPerturbation}(b). Stability denotes the ball's trajectory varies along a boundary (e.g. between radius $\tau$ and radius $r$); Non-Stability indicates the ball's trajectory departs from the original position and moves towards infinity (e.g. exceed radius $r$); Asymptotic Stability implies the ball's trajectory first departs from the original position, and finally stops at the original position (e.g. the origin) even if it may experience a long-time motion. Note that there may exist multiple equilibrium points, also called fixed points, but it does not mean all of them are asymptotically stable.
\end{remark}}}
\end{center}

As well known, GAA aims to breach graph's resilience by continuously perturbing graph structure (topology) or/and node feature. From the viewpoint of dynamic system, each time the perturbation tries to make graph deviate from the original stable state to non-stability state. Hence, we think each graph ought to have such an equilibrium point to keep robust, then propose the following hypothesis. 

\begin{assumption}
\textbf{(Adversarial Resilience of Graph Regime)} For each graph regime, there exists an asymptotically-stable equilibrium point, corresponding to its critical adversarial resilience state, to keep it attack-resilient under continuous and bounded perturbations by adversarial attack, and such equilibrium point can be acquired as time goes.
\end{assumption}

GAAs generally deceive target model through dynamically adding/deleting important nodes/edges, in addition to modifying weights/features. Thus, we can abstractly mirror such continuous perturbations as the dynamic-variation process of complex multi-object physical systems, such as electromagnetic field, charged particles, etc. As analyzed in Fig. \ref{fig:BallPerturbation}(b), a dynamic system can fall into the three states: stable, non-stable, or asymptotically stable. Of course, staring from the stability state, we anticipate the dynamic system can finally converge into the asymptotically-stable state, this implies it can always go back to the original equilibrium point in the end even if a long-time dynamic undergoes. Therefore, referring to the converged trajectory of the ball at position $P_3$, we give the definition of asymptotic stability. The symbol "$\rightharpoonup$" denotes a vector representing the set of objects.

\begin{definition} \label{Definition: AsymptStability}
	\textbf{(Asymptotic Stability \cite{KhalilBook15})} Let $
	\frac{{d\mathord{\buildrel{\lower3pt\hbox{$\scriptscriptstyle\rightharpoonup$}} 
				\over x} }}{{dt}} = v(\mathord{\buildrel{\lower3pt\hbox{$\scriptscriptstyle\rightharpoonup$}} 
		\over x} ) $ be a set of locally Lipschitz functions defined over a domain $\xi \subset \mathbb{R}^n$, and it contains the origin\footnote{The origin always can be obtained via coordinate transformation.}, i.e. $v(0)$=0. Then, the equilibrium point $x$=0 of $\frac{{d\mathord{\buildrel{\lower3pt\hbox{$\scriptscriptstyle\rightharpoonup$}} \over x} }}{{dt}} $ is asymptotically stable if there exists $\tau$ $>$ 0, s. t. $ \left\| {\mathord{\buildrel{\lower3pt\hbox{$\scriptscriptstyle\rightharpoonup$}} 
			\over x} \left( {t_0 } \right)} \right\| < \tau  \Rightarrow \mathop {\lim }\limits_{t \to \infty } \mathord{\buildrel{\lower3pt\hbox{$\scriptscriptstyle\rightharpoonup$}} \over x} \left( t \right) = 0 $.
\end{definition}

\begin{center}
\fcolorbox{black}{gray!10}{\parbox{1.0\linewidth}{
\begin{remark} For a dynamic system falling in the state of asymptotic stability, it means there exists asymptotically-stable equilibrium point and the system's dynamic variation trajectory will finally converges into such equilibrium point, no matter how long the dynamic may undergo. This procedure well-matches the behavior of adversarial attacks, that is to say, we can equivalently map the graph adversarial perturbation as dynamic system's oscillation. In view of this, regarding the adversarial perturbations on graph as dynamic variation of multi-object dynamic system is persuasive.
\end{remark}}}
\end{center}

To make the asymptotically-stable equilibrium point (ASEP) fall into the origin, we give Lyapunov Criterion as stated in Appendix \ref{Sec:LyapunovCriterion}.	
In the light of Lyapunov criterion, the origin must be an ASEP. 

\subsection{GAA modeling}
Graph dynamics \cite{Kundu22,Gao16} are in general represented as a linear weighted average of the $N$ state variables each of which is associated with a node $i$, and the nonlinear influence of other nodes on node $i$. Hence, we can divide the adversarial perturbation on graph regime into two parts: i) linear perturbation, reflecting the graph is disturbed by a linear action; ii) nonlinear perturbation, implying the graph is disturbed by a nonlinear action. As time $t$ goes, the dynamic-variation function under continuous perturbations can be defined as
\begin{equation} \label{Equ-1}
	\frac{d}{{dt}}\mathord{\buildrel{\lower3pt\hbox{$\scriptscriptstyle\rightharpoonup$}} 
		\over x} \left( t \right) = A\vec x\left( t \right) + B\left( { - \mathord{\buildrel{\lower3pt\hbox{$\scriptscriptstyle\rightharpoonup$}} 
			\over \phi } \left(  \cdot  \right)} \right),
\end{equation}
where $ \vec x\left( t \right)$ is an $N$-dimensional vector denoting the states of $N$ nodes, reflecting the linear perturbation, and $\phi ( \cdot )$ is a nonlinear-mapping function representing the nonlinear perturbation. Equation \ref{Equ-1} illuminates the dynamic variation of graph nodes under the continuous perturbations (edge removal/injection) by adversarial attack. Accordingly, each node may have a different states at different time $t$. It is also in concert with the training process of deep learning, namely, during each-round backpropagation-based parameter update, the pre-round output is adopted to compute gradient for the next-round parameter update. In the light of deep-learning principle, we here deem $\phi (\cdot)$ as the input to infer next-round dynamic update, and it should be presented by the output of pre-round perturbation result, in view of this, we can naturally define the output $\mathord{\buildrel{\lower3pt\hbox{$\scriptscriptstyle\rightharpoonup$}} \over y} \left( t \right) = C\mathord{\buildrel{\lower3pt\hbox{$\scriptscriptstyle\rightharpoonup$}} \over x} \left( t \right) $, which represents the pre-round perturbed states of $N$ nodes, thus we have $ \phi \left(  \cdot  \right) = \phi \left({\mathord{\buildrel{\lower3pt\hbox{$\scriptscriptstyle\rightharpoonup$}} \over y} \left( t \right)} \right) $ accordingly. Matrices $A$, $B$, $C$ are three general parameters, whose elements depend on the particular adversarial attack, the subsequent Equations \ref{Equ-parameter}-\ref{Equ-trajectory} and associated parameters in Table \ref{Table:ResiTraPara} can interpret this point.

The integral from time $t_0$ to $t$ reflects a period of time aggregation of dynamic variation, which can be equivalently recognized as the process of adversarial learning epoch by epoch, thus the utilization of integral form is adequate to illuminate the effect of continuous adversarial perturbations on graph in a time frame. Therefore, we straightly define the time-aware 
aggregation as $\int_{t_0 }^t {x\left( \kappa  \right)d\kappa}$.
Assume a persistent long-time adversarial learning process, i.e. $t \to \infty$, and simultaneously for the sake of derivative calculation, we introduce Laplace Transform for the above time-aware integral
\begin{equation} \label{Equ-4}
	L\left[ {x\left( t \right)} \right] = X\left( s \right) = \int_{t_0  = 0}^{t \to \infty } {x\left( t \right)e^{ - st} dt}, 
\end{equation}
where $	s = \sigma  + \omega j,_{} j = \sqrt { - 1} $ is a complex number. Given the property of linear transformation of Laplace Transform, we have $Y\left( s \right) = C \cdot X\left( s \right)$, and the first derivative's Laplace Transform is 
\begin{equation} \label{Equ-LapFirstDeri}
	\begin{array}{l}
		L\left[ {\frac{d}{{dt}}x\left( t \right)} \right] = \int_{t_0  = 0}^{t \to \infty } {\left( {\frac{d}{{dt}}x\left( t \right)} \right)e^{ - st} dt} 
	= \mathop {\lim }\limits_{t \to \infty } x\left( t \right)e^{ - st}  - x\left( 0 \right) + sX\left( s \right) \\ 
	\end{array}.
\end{equation} 

In general, to achieve the convergence, the real part of complex number $s$ must satisfy $	{{\mathop{\rm Re}\nolimits} \left( s \right) > 0}$, furthermore, the initial condition meets ${x\left( 0 \right) = 0}$, thus $L\left[ {\frac{d}{{dt}}x\left( t \right)} \right]$=$ sX\left( s \right)$. Substituting Equations \ref{Equ-4} and \ref{Equ-LapFirstDeri} into Equation \ref{Equ-1}, an intermediate function $G\left( s \right)$ can be obtained between output and input,   
\begin{equation} \label{Equ-TransferFunction}
\begin{split}
\mathord{\buildrel{\lower3pt\hbox{$\scriptscriptstyle\rightharpoonup$}} 
\over X} \left( s \right) = \left( {sI - A} \right)^{ - 1}  \cdot B \cdot L\left[ {\mathord{\buildrel{\lower3pt\hbox{$\scriptscriptstyle\rightharpoonup$}} 
\over \phi } \left( {\mathord{\buildrel{\lower3pt\hbox{$\scriptscriptstyle\rightharpoonup$}} 
\over y} \left( t \right)} \right)} \right]  \\
\begin{array}{l}
G\left( s \right) = \frac{{\mathord{\buildrel{\lower3pt\hbox{$\scriptscriptstyle\rightharpoonup$}} 
\over Y} \left( s \right)}}{{L\left[ {\mathord{\buildrel{\lower3pt\hbox{$\scriptscriptstyle\rightharpoonup$}} 
\over \phi } \left( {\mathord{\buildrel{\lower3pt\hbox{$\scriptscriptstyle\rightharpoonup$}} 
\over y} \left( t \right)} \right)} \right]}} = \frac{{C \cdot \mathord{\buildrel{\lower3pt\hbox{$\scriptscriptstyle\rightharpoonup$}} 
\over X} \left( s \right)}}{{L\left[ {\mathord{\buildrel{\lower3pt\hbox{$\scriptscriptstyle\rightharpoonup$}} 
\over \phi } \left( {\mathord{\buildrel{\lower3pt\hbox{$\scriptscriptstyle\rightharpoonup$}} 
\over y} \left( t \right)} \right)} \right]}} 
= C\left( {sI - A} \right)^{ - 1} B \\ 
\end{array}
\end{split}.
\end{equation}

\begin{center}
\fcolorbox{black}{gray!10}{\parbox{1.0\linewidth}{
\begin{remark} The function $G(s)$ reflects the intermediate variation process of information-passing from an input $L\left[ {\phi \left( {\mathord{\buildrel{\lower3pt\hbox{$\scriptscriptstyle\rightharpoonup$}} \over y} } \right)} \right] $ to an output $Y(s)$ in Laplace form, thus, we can control the coefficients $\{A, B, C\}$ to obtain the result that we anticipate the system to output. From the viewpoint of adversarial resilience, we anticipate the graph finally converges into critical state of adversarial resilience (asymptotically-stable equilibrium point) to remain attack resilient, no mater how long it takes. In fact, you can image function $G(s)$ represents the intermediate process of adversarial learning.
\end{remark}}}
\end{center}

We propose the following theorem to warrant Equation \ref{Equ-1} converges into the ASEP.

\begin{theorem} \label{Theorem-2}
	\textbf{(Existence of Asymptotically-Stable Equilibrium point)} The dynamic-variance function (Equation \ref{Equ-1}) has asymptotically-stable equilibrium point(s) if matrix $A$ is Hurwitz, the output $\mathord{\buildrel{\lower3pt\hbox{$\scriptscriptstyle\rightharpoonup$}}\over y} \left( t \right)$ and its associated nonlinear-mapping input $\phi \left( {\mathord{\buildrel{\lower3pt\hbox{$\scriptscriptstyle\rightharpoonup$}} \over y} \left( t \right)} \right) $ satisfy the condition that $ \mathord{\buildrel{\lower3pt\hbox{$\scriptscriptstyle\rightharpoonup$}} 
		\over \phi } \left( {\mathord{\buildrel{\lower3pt\hbox{$\scriptscriptstyle\rightharpoonup$}} 
			\over y} \left( t \right)} \right)^T  \cdot \left[ {\mathord{\buildrel{\lower3pt\hbox{$\scriptscriptstyle\rightharpoonup$}} 
			\over y} \left( t \right) - M\mathord{\buildrel{\lower3pt\hbox{$\scriptscriptstyle\rightharpoonup$}} 
			\over \phi } \left( {\mathord{\buildrel{\lower3pt\hbox{$\scriptscriptstyle\rightharpoonup$}} 
				\over y} \left( t \right)} \right)} \right] > 0 $, where matrix $ M = diag\left( {{\raise0.7ex\hbox{$1$} \!\mathord{\left/{\vphantom {1 {k_1 }}}\right.\kern-\nulldelimiterspace}	\!\lower0.7ex\hbox{${k_1 }$}}, \cdots ,{\raise0.7ex\hbox{$1$} \!\mathord{\left/ {\vphantom {1 {k_p }}}\right.\kern-\nulldelimiterspace}
			\!\lower0.7ex\hbox{${k_p }$}}} \right) $, $k_i$$>$0. Furthermore, given $ \psi  = diag\left( {\gamma _1 , \cdots ,\gamma _p } \right) $, there exists constant $\gamma_i  \ge 0 $, such that $ (1 + \lambda _k \gamma _i ) \ne 0 $ for each eigenvalue $ \lambda _k $ of matrix $A$, and meanwhile $ M + \left( {I + s\psi } \right)G\left( s \right) $ is strictly positive real.
\end{theorem}

\textit{Proof.} See Appendix \ref{Sec:ProofTheorem2}

\section{Adversarial resilience inference}	\label{Sec:Practice}
\subsection{One-dimensional perturbation mapping}
The work \cite{Kundu22} studies single-node autonomous behavior (SNAB) and pairwise-node interaction behavior (PNIB), taking as a referral, our mentioned adversarial perturbation is investigated along this thought. In detail, we refer to the dynamic variation of graph under adversarial attack as coupled linear function $ \chi \left( {\mathord{\buildrel{\lower3pt\hbox{$\scriptscriptstyle\rightharpoonup$}} 	\over x} } \right)	$ and nonlinear function $ \varphi \left({\mathord{\buildrel{\lower3pt\hbox{$\scriptscriptstyle\rightharpoonup$}} \over x},\mathord{\buildrel{\lower3pt\hbox{$\scriptscriptstyle\rightharpoonup$}} \over x} } \right) $. Therefore, for each node $i$, we have 
\begin{equation} \label{Equ-Mapping}
	\frac{{dx_i }}{{dt}} = \tilde A_{\left( i \right)} \chi \left( {x_i } \right) + \sum\nolimits_{j = 1}^N {\tilde B_{\left( {ij} \right)} \varphi (x_i ,x_j )},
\end{equation}
where $	\tilde A_{\left( i \right)} $ denotes the linear SNAB-caused dynamic-variation coefficient of node $i$, and $ \tilde B_{\left( {ij} \right)} $ stands for the nonlinear PNIB-caused dynamic-variation coefficient. 

Currently, the node is in general represented in a high-dimension space, to achieve lightweight computation, some dimension-reduction methods are adopted to accomplish low-dimension mapping. Considering the complexity of linkage/interactive relationship over nodes, we here employ one-dimensional condensed method to map the whole graph's dynamic variation. Resorting to Heterogeneous Mean-Field (HMF) approximation theory \cite{Kundu22}, pursuant to the deduction step by step in Appendix \ref{Sec:oneDimenFunctionDeduce}, we obtain the one-dimensional perturbation mapping function  

\begin{equation}\label{Equ-OneDimen}
	\frac{{d\tilde x}}{{dt}} = \tilde A_{\left(  \cdot  \right)} \chi \left( {\tilde x} \right) + \tilde \beta \varphi \left( {\tilde x,\tilde x} \right).
\end{equation}

From Equation \ref{Equ-OneDimen}, we know the mapping function of the whole graph's dynamics 
is condensed as a single variable $\tilde \beta $, which can also be regarded as the reflection of robustness of the graph as a whole.  
\begin{center}
\fcolorbox{black}{gray!10}{\parbox{1.0\linewidth}{
\begin{remark} The condensed one-dimensional mapping function may lead to somewhat loss of graph-structure information. Alternatively, the sophisticated two-dimensional or multi-dimensional mapping functions are workable as well. The principle to develop an appropriate mapping function lies in the capability of faithfully reflecting the caused dynamics/perturbations to the most extent.
\end{remark}}}
\end{center}

\subsection{Equilibrium-point exploration}
Given that $ \chi \left( {x_i } \right) $ is linear and $\varphi (x_i ,x_j )$ is nonlinear, we can simply and generally define a dynamic-variation equation to concretize the process of adversarial perturbations as, 
\begin{equation} \label{Equ-instantiation}
	\frac{{dx_i }}{{dt}} =  - \tilde A_{\left( i \right)} x_i  + \sum\nolimits_{j = 1}^N {\tilde B_{\left( {ij} \right)} \frac{{Hx_j^2 }}{{\theta x_j^2  + 1}}}.
\end{equation}
The first term on the right hand of Equation \ref{Equ-instantiation} presents node $i$'s self-dynamics, for detail, we utilize the degree centrality of node $i$ to reflect the probability of being attacked. The second stands for the interactively-affected perturbation by its neighboring nodes. This simple yet effective one-dimensional function is developed to map the dynamics of graph regime as a whole, in fact, two-dimensional mapping or multi-dimensional mapping can be explored as well for distinct graph topologies, since our proposed asymptotically-stable theoretic framework fully supports multi-dimensional dynamic-variation function. The more accurate the modeling on GAAs, the higher the adversarial resilience would be. To get an accurate function to capture the dynamics, usually, some extra variable $\theta$ $>$ 0 is needed, and $H$ denotes the average in-degree weight of node $i$, implying the affected extent of perturbation by neighbors. Referring to Equation \ref{Equ-OneDimen} in the condensed one-dimensional form, we can concretize Equation \ref{Equ-instantiation} as 
\begin{equation} \label{Equ-parameter}
	\frac{{d\tilde x}}{{dt}} =  - a\tilde x + \tilde \beta \frac{{H\tilde x_{}^2 }}{{\theta \tilde x_{}^2  + 1}},
\end{equation}
where $a$ is a successor constant condensed from vector $\tilde A_{\left( i \right)}$. The equation above exhibits the detailed settings regrading the concrete graph attributes: degree centrality, in-degree weight, etc. Here what we need emphasize is that Equation \ref{Equ-OneDimen} is a theoretical guidance, Equations \ref{Equ-instantiation} and \ref{Equ-parameter} are concretized deductions w.r.t. specified graph topology. 
From Equation \ref{Equ-parameter}, we infer the following theorems.

\begin{theorem} \label{Theorem-5}
	Upon the condensed one-dimensional mapping function, a graph regime disturbed by graph adversarial attack can be finally converged into an asymptotically-stable equilibrium point if the constant $k$ in Theorem \ref{Theorem-2} belongs to the interval $ \left[ {H \cdot \left( {2\sqrt \theta  } \right)^{ - 1} ,\infty } \right) $.       
\end{theorem}

\textit{Proof.} See Appendix \ref{Sec:ProofTheorem5}.

Theorem \ref{Theorem-5} gives the perturbation domain only within which the asymptotically-stable equilibrium point can be finally found out, and the following Theorem \ref{Theorem-6} warrants the existence of such point. 

\begin{theorem} \label{Theorem-6}
The condensed one-dimensional dynamic-variation function (Equation \ref{Equ-parameter}) is asymptotically stable.
\end{theorem}

\textit{Proof.} See Appendix \ref{Sec:ProofTheorem6}.

\begin{center}
\fcolorbox{black}{gray!10}{\parbox{1.0\linewidth}{
\begin{remark} \textbf{Theorem \ref{Theorem-5}} and \textbf{Theorem \ref{Theorem-6}} guarantee our proposed one-dimensional dynamic-variation function has asymptotically-stable equilibrium point, i.e. there exist a critical state of adversarial resilience to keep graph regime attack-resilient. The parameter $\omega \in \mathbb{R} $ in proof of Theorem \ref{Theorem-6} means the equilibrium point can be found out in the real number field, simultaneously with the constraint of perturbation scope of $k$. In other words, the existence of equilibrium point can be only warranted within a bounded domain. Once the perturbation jumps out of the constraint domain, 
the critical state of adversarial resilience would disappear.
\end{remark}}}
\end{center}

Seen from Equation \ref{Equ-parameter}, we know $ \frac{{d\tilde x}}{{dt}} $=0 if $\tilde x $ = 0, which means the entire graph is completely broken with no connection between nodes (Equation \ref{Equ-nodePerturbation}), this is also a so-called equilibrium point, but not what we anticipate although this case indeed happens in reality. Thereby, we need calculate other equilibrium points divided by $\tilde x $, and at the same time regard $\tilde x $ as the variable of $\tilde \beta $, 
\begin{equation} \label{Equ-trajectory}
	\tilde \beta \left( {\tilde x} \right) = a \cdot \frac{{\theta \tilde x_{res}^2  + 1}}{{H\tilde x}}.
\end{equation}
Pursuant to this function, we know a trajectory consisting of infinite equilibrium points can be drawn in theory. 
Since our work focuses on enhancing adversarial resilience from the angle of graph topology, i.e. resist GAAs from the radical graph structure. In order to endow appropriate variables for Equation \ref{Equ-parameter}, we choose the basic graph property--Degree Centrality, which reflects the “hub” role in the connectivity of entire graph, and defined as $ C_{deg} (i) = {\raise0.7ex\hbox{${\left| {d_i } \right|}$} \!\mathord{\left/
		{\vphantom {{\left| {d_i } \right|} {\left( {N - 1} \right)}}}\right.\kern-\nulldelimiterspace}
	\!\lower0.7ex\hbox{${\left( {N - 1} \right)}$}} $ where $d_i$ denotes the set of neighbors of node $i$. Accordingly, we can separately obtain in-degree centrality $
C_{Indeg} (i) = {\raise0.7ex\hbox{${\left| {d_{Inneib} \left( i \right)} \right|}$} \!\mathord{\left/
		{\vphantom {{\left| {d_{Inneib} \left( i \right)} \right|} {\left( {N - 1} \right)}}}\right.\kern-\nulldelimiterspace}
	\!\lower0.7ex\hbox{${\left( {N - 1} \right)}$}} $ where $ d_{Inneib} \left( i \right) $ denotes the set of node $i$'s in-going neighbors, and out-degree centrality $ C_{Outdeg} (i) = {\raise0.7ex\hbox{${\left| {d_{Outneib} (i)} \right|}$} \!\mathord{\left/
		{\vphantom {{\left| {d_{Outneib} (i)} \right|} {\left( {N - 1} \right)}}}\right.\kern-\nulldelimiterspace} \!\lower0.7ex\hbox{${\left( {N - 1} \right)}$}} $ where $ d_{Outneib} \left( i \right) $ indicates the set of node $i$'s out-going neighbors. As mentioned previously, we set $ a = \eta  \cdot \frac{{\sum\nolimits_i {C_{Indeg} (i)}  + \sum\nolimits_i {C_{Outdeg} (i)} }}{N}$, $ H = {\raise0.7ex\hbox{${\sum\nolimits_i {d_{Inneib} \left( i \right)} }$} \!\mathord{\left/ {\vphantom {{\sum\nolimits_i {d_{Inneib} \left( i \right)} } N}}\right.\kern-\nulldelimiterspace} \!\lower0.7ex\hbox{$N$}} $, and use the deduced Equation \ref{Equ-nodePerturbation} in Appendix \ref{Sec:oneDimenFunctionDeduce} to calculate $\tilde x$.

Although the degree centrality is the basic and simple attribute in graph regime, the subsequent experiment showcases it plays an important role to boost adversarial resilience under the guidance of equilibrium-point trajectory, that is to say, the attack-resilient graph topology can be established resorting to this basic degree-related properties.						

\section{Implementation} \label{Sec: Defense}
The asymptotically-stable equilibrium point is in accordance with the critical state of adversarial resilience for graph regime, thus we take such equilibrium point as the referral to build a new attack-resilient graph topology for perturbed graph. In detail, resorting to the adjacency matrix of undisturbed graph, we first compute $\tilde \beta $ using Equation \ref{Equ-trajectory} to obtain an asymptotically-stable equilibrium point as shown "Original" in Fig. \ref{Fig: PolblogsTrajec}-\ref{Fig:AmazonPhotoTrajec}. Section \ref{Sec:trajectoryComputation} will detail the computation of equilibrium-point trajectory. Then, we set up an iterative-optimization procedure to generate a robust adjacency matrix referring to this asymptotically-stable equilibrium point, i.e.,  
enforcing the modified adjacency matrix-based coordinator point to be close to the "Original" point. 
For easy understanding, we state the procedures in Algorithm \ref{Algorithm-1} and Algorithm \ref{Algorithm-2} in Appendix \ref{Sec:Algorithms1-2}.

\section{Experiment evaluation} \label{Sec: Experiments}
Apart from regular performance analytics, our experiments need first to answer three fundamental research questions: 
i) \textbf{RQ1:} Whether there is a correlation between adversarial learning and complex dynamic system; 
ii) \textbf{RQ2:} Whether Laplace Transfer is adaptable to embody the process of continuous adversarial perturbations; 
and iii) \textbf{RQ3:} Whether the condensed one-dimensional mapping function suffices to appropriately present the dynamics of entire graph regime. 

\subsection{Configuration} 
The configurations on the statistics of five commonly-used moderate- and large-scale datasets\footnote{https://pytorch-geometric.readthedocs.io/en/latest/cheatsheet/data\_cheatsheet.html}: 
Polblogs, Cora, Cora\_ML, Citeseer, and Amazon Photo, five representative baselines: GCN \cite{KipfWelling17}, GAT \cite{Velickovic18}, GCN-SVD \cite{Entezari20}, HANG \cite{ZhaoKang23}, 
and Mid-GCN \cite{HuangJin25}, three non-targeted poisoning attacks: Metattack \cite{Daniel19}, CE-PGD \cite{XuChen19}, and DICE\cite{Waniek16}, and execution settings are detailed in Appendix \ref{Sec:Configuration}.  

\subsection{Equilibrium-point trajectory} \label{Sec:trajectoryComputation}
\begin{figure} [tbhp]
\centering
\subfigure[Metattack]{
\begin{minipage}[t]{0.3\columnwidth}
\centering
\includegraphics[height=1.2in, width=1.78in]{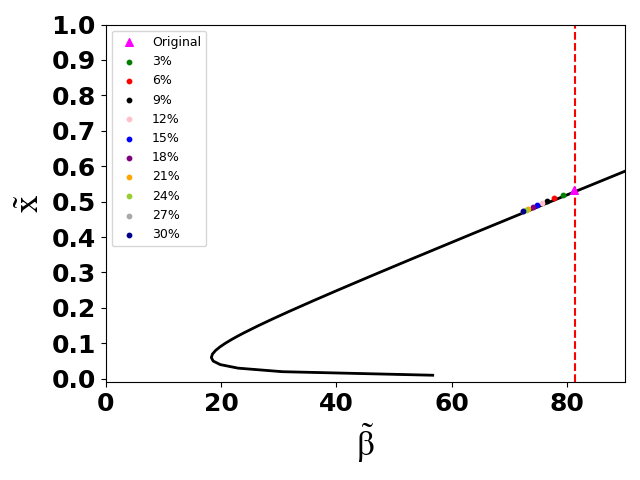}
\label{Fig: Metattack}
\end{minipage}
}
\vspace{0.02cm}
\subfigure[CE-PGD]{
\begin{minipage}[t]{0.3\columnwidth}
\centering
\includegraphics[height=1.2in, width=1.78in]{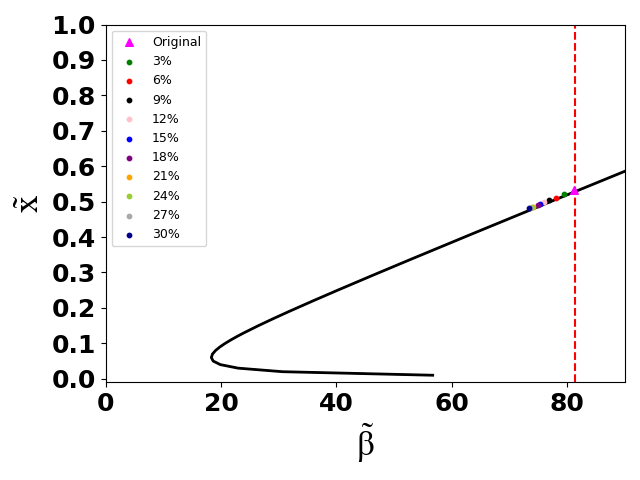}
\label{Fig: PGD} 
\end{minipage}
}
\vspace{0.02cm}
\subfigure[DICE]{
\begin{minipage}[t]{0.3\columnwidth}
\centering
\includegraphics[height=1.2in, width=1.78in]{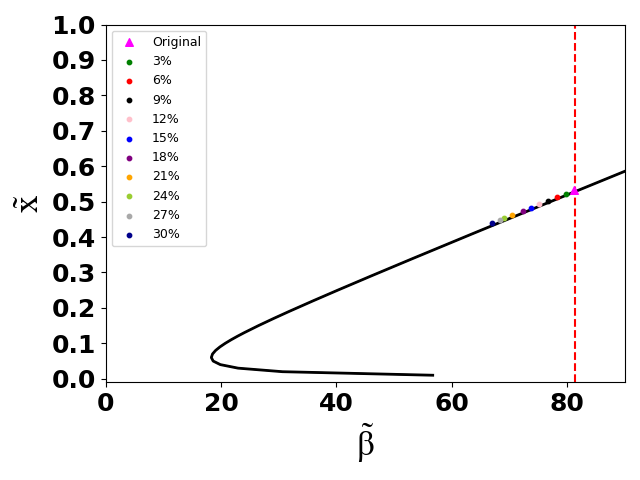}
\label{Fig:DICE}
\end{minipage}
}
\caption{Adversarial effect with Polblogs.}
\label{Fig: PolblogsTrajec}
\end{figure}

As aforementioned, different graph topologies may have distinct trajectories of equilibrium points. In order to well predict the adverseness of the three representative GAAs: Metattack \cite{Daniel19}, CE-PGD \cite{XuChen19}, and DICE \cite{Waniek16}, 
we use the parameters in Table \ref{Table:ResiTraPara} in Appendix \ref{Sec:EquiliPointTraj} to draw equilibrium-point trajectories with Polblogs in Fig. \ref{Fig: PolblogsTrajec}, Cora\_ML in Fig. \ref{Fig:Cora_MLTrajec}, Cora in Fig. \ref{Fig:CoraTrajec}, Citeseer in Fig. \ref{Fig:CiteseerTrajec}, and Amazon Photo in Fig. \ref{Fig:AmazonPhotoTrajec}, wherein the horizontal-longitudinal coordinate converts. 

From Fig. \ref{Fig: PolblogsTrajec}-\ref{Fig:AmazonPhotoTrajec}, we observe the adversarial resilience degrades almost along the trajectory curves, this phenomenon unveils three-aspect hints: i) the concordance of equilibrium-point trajectory and adverseness of adversarial attacks discloses the existence of correlation between adversarial learning and complex dynamic system, which answers the first question \textbf{RQ1}; ii) note that for each three curves drawn from the five datasets, the variation tendency of inferred adversarial resilience under different-level adversarial perturbations ranging from 3\% to 30\% goes exactly along the theoretical equilibrium-point curve which is deduced by Laplace Transfer, this unveils the rationality of introduction of Laplace Transfer for the modeling on graph adversarial attacks, which replies the second question \textbf{RQ2}; and iii) the condensed one-dimensional function-based equilibrium-point trajectory is appropriate to capture the dynamics of graph regime as a whole under the continuous adversarial perturbations, which answers the third question \textbf{RQ3}.
				
\subsection{Performance analytics}
\subsubsection{Adversarial resilience promotion} \label{Sec:AdverResiPro}
Two groups of experiments are presented in this subsection, one group showcases the effectiveness of our work compared to another two similar (adjacency-matrix based modification) approaches, i.e. GCN-SVD \cite{Entezari20} and GCN \cite{KipfWelling17}; the other exhibits the significant enhancement of our approach on another three non-adjacency-matrix based approaches, i.e. GAT \cite{Velickovic18}, HANG \cite{ZhaoKang23}, and Mid-GCN \cite{HuangJin25}, wherein the graph's adjacency matrix is at first optimized referring to the asymptotically-stable equilibrium point, then, the node classification task is performed by GCN. 
				
\textbf{Asymptotically-Stable Equilibrium-Point Trajectory Conducts Attack-Resilient Topology.} The results in Table \ref{Table: Polblogs}, Tables \ref{Table: CiteseerUp}-\ref{Table: AmazonPhoto} in Appendix \ref{Sec: ExtraPerformanceResults} showcase the remarkable robustness improvement of our EquiliRes compared to GCN-SVD and GCN. 
The adversarial perturbation rate (APR) 0\% means there does not exist adversarial attack, we execute matrix-modification operation with a little percentage 1\%, from which, we can deduce two-facet opinions: i) the accuracy of node classification is high without adversarial perturbation, e.g. approximately 94\% with the three methods on Polblogs dataset; and ii) our EquiliRes gains larger accuracy compared to the two baselines, which implies our proposed asymptotic stability theory can validly locate the critical state of adversarial resilience to keep the graph more robust. Even for clean graph, our method still can further boost it. 
\begin{table*}[tbp]  
\caption{Accuracy (\%) of node classification on Polblogs (\textbf{bold}-the best).}
\label{Table: Polblogs}
\vskip 0.15in
\begin{small}
\centering
	\begin{tabular}{|p{1.1cm}|p{1.45cm}|>{\centering\arraybackslash}p{1.32cm}|>{\centering\arraybackslash}p{1.32cm}|>{\centering\arraybackslash}p{1.32cm}|>{\centering\arraybackslash}p{1.32cm}|>{\centering\arraybackslash}p{1.32cm}|>{\centering\arraybackslash}p{1.32cm}|}	
		\hline
		\textbf{GAA} & \diagbox[dir=NW]{\textbf{Def.}}{\textbf{Per.}} & \textbf{0\%} & \textbf{5\%} & \textbf{10\%} & \textbf{15\%} & \textbf{20\%} & \textbf{25\%}\\
		\hline
		\multirow{3}{*}{Metattack} & GCN & 94.26$\pm$1.24 & 76.35$\pm$1.39 & 70.61$\pm$1.51 & 67.66$\pm$1.39 &	67.45$\pm$1.45 & 66.34$\pm$1.54 \\
		\cline{2-8}
		& GCN-SVD & 93.98$\pm$0.52 &\textbf{92.52$\pm$1.09} &  87.99$\pm$4.35 & 71.13$\pm$2.93 & 67.25$\pm$1.69 & 65.56$\pm$1.49 \\
		\cline{2-8}
		& EquiliRes & 
		\begin{tabular} [c]{@{}l@{}} \textbf{94.34$\pm$0.81} \\ (\textbf{\textcolor{red}{8\%$\uparrow$}}) \end{tabular} &  
		\begin{tabular} [c]{@{}l@{}} 90.11$\pm$2.62 \\ (\textbf{\textcolor{red}{2.41\%$\downarrow$}}) \end{tabular} & 
		\begin{tabular} [c]{@{}l@{}} \textbf{89.70$\pm$2.22} \\ (\textbf{\textcolor{red}{1.71\%$\uparrow$}}) \end{tabular} &  \begin{tabular} [c]{@{}l@{}} \textbf{86.53$\pm$1.47} \\ (\textbf{\textcolor{red}{15.40\%$\uparrow$}}) \end{tabular} & 
		\begin{tabular} [c]{@{}l@{}} \textbf{81.41$\pm$1.94} \\ (\textbf{\textcolor{red}{13.96\%$\uparrow$}}) \end{tabular} &  \begin{tabular} [c]{@{}l@{}} \textbf{73.11$\pm$2.60} \\ (\textbf{\textcolor{red}{6.97\%$\uparrow$}}) \end{tabular} \\		         		           
		\hline\hline
		\multirow{3}{*}{CE-PGD} & GCN & 94.26$\pm$1.24 & 84.07$\pm$0.65 & 79.19$\pm$1.08 & 76.86$\pm$0.91 &	75.37$\pm$1.05 & 74.47$\pm$1.14 \\
		\cline{2-8}
		& GCN-SVD & 93.98$\pm$0.52 & 85.31$\pm$1.06 & 79.43$\pm$1.04 & 76.82$\pm$1.06 & 75.59$\pm$1.00 & 74.80$\pm$1.26 \\
		\cline{2-8}
		& EquiliRes & 
		\begin{tabular} [c]{@{}l@{}} \textbf{94.34$\pm$0.81} \\ (\textbf{\textcolor{red}{0.08\%$\uparrow$}}) \end{tabular} &  
		\begin{tabular} [c]{@{}l@{}} \textbf{86.69$\pm$1.19} \\ (\textbf{\textcolor{red}{1.38\%$\uparrow$}})  \end{tabular} &  
		\begin{tabular} [c]{@{}l@{}} \textbf{82.67$\pm$1.12} \\ (\textbf{\textcolor{red}{3.24\%$\uparrow$}}) \end{tabular} & 
		\begin{tabular} [c]{@{}l@{}} \textbf{79.93$\pm$1.39} \\ (\textbf{\textcolor{red}{3.07\%$\uparrow$}}) \end{tabular} &  
		\begin{tabular} [c]{@{}l@{}} \textbf{76.82$\pm$1.01} \\ (\textbf{\textcolor{red}{1.23\%$\uparrow$}}) \end{tabular} &
		\begin{tabular} [c]{@{}l@{}} \textbf{75.59$\pm$1.41} \\ (\textbf{\textcolor{red}{0.79\%$\uparrow$}}) \end{tabular} 
		\\			
		\hline\hline			
		\multirow{3}{*}{DICE} & GCN & 94.26$\pm$1.24 & 86.58$\pm$1.22 & 80.83$\pm$0.96 & 77.42$\pm$1.07 & 75.44$\pm$1.13 &	73.37$\pm$0.99 \\
		\cline{2-8}
		& GCN-SVD & 93.98$\pm$0.52 &\textbf{90.83$\pm$0.78} & 88.29$\pm$0.50 & 85.56$\pm$0.66 & 82.92$\pm$1.20 & 81.29$\pm$0.96 \\
		\cline{2-8}
		& EquiliRes & 
		\begin{tabular} [c]{@{}l@{}} \textbf{94.34$\pm$0.81} \\ (\textbf{\textcolor{red}{0.08\%$\uparrow$}}) \end{tabular} &  
		\begin{tabular} [c]{@{}l@{}} 89.47$\pm$0.95 \\ (\textbf{\textcolor{red}{1.36\%$\downarrow$}})  \end{tabular} &  
		\begin{tabular} [c]{@{}l@{}} \textbf{88.43$\pm$2.32} \\ (\textbf{\textcolor{red}{0.14\%$\uparrow$}}) \end{tabular} & 
		\begin{tabular} [c]{@{}l@{}} \textbf{86.85$\pm$2.11}\\ (\textbf{\textcolor{red}{1.29\%$\uparrow$}}) \end{tabular} &  
		\begin{tabular} [c]{@{}l@{}} \textbf{84.75$\pm$1.54} \\ (\textbf{\textcolor{red}{1.83\%$\uparrow$}}) \end{tabular} &		
		\begin{tabular} [c]{@{}l@{}} \textbf{83.45$\pm$1.63} \\ (\textbf{\textcolor{red}{2.16\%$\uparrow$}}) \end{tabular} \\	
		\hline				
	\end{tabular}
\end{small}
\vskip -0.1in
\end{table*}
					
For Metattack attack, the accuracy raises by 15.40\%, 13.96\%, 6.97\% than the second best when the APRs are 15\%, 20\% and 25\% on Polblogs, and by 5.67\%, 6.51\% and 4.93\% on Citeseer when APRs are 10\%, 15\% and 20\% respectively, which shows our proposed approach can effectively defend against such non-targeted gradient-based attack. For CE-PGD attack, although it perturbs clean graph in a maximum probability through enlarging the cross-entropy loss, pursuant to Theorem \ref{Theorem-6}, we know that, under a bounded perturbation, our proposed dynamic-variation function can finally ends up at asymptotically-stable equilibrium point to make the graph attack-resilient, thus our EquiliRes possesses 3.24\%, 3.07\% and 1.23\% improvement on Polblogs, and 2.65\%, 2.53\% and 2.30\% on Citeseer when APRs are 10\%, 15\% and 20\% respectively. 
Compared to Metattack and CE-PGD, the attack strength of modularity-based DICE is weaker, due to the straight manipulation of reducing the connection density within same community while increasing the linkage over different communities, thus both GCN-SVD and our EquiliRes have relatively good performance. 

\begin{table*}[hbtp]  
\caption{Accuracy (\%) of node classification under combination with EquiliRes using Metattack.}
\label{Table: EquiliResCombine-1stPart}
\begin{small}
\vskip 0.1in
\centering
\begin{tabular}{|p{0.85cm}|p{1.5cm}|>{\centering\arraybackslash}p{1.3cm}|>{\centering\arraybackslash}p{1.3cm}|>{\centering\arraybackslash}p{1.45cm}|>{\centering\arraybackslash}p{1.45cm}|>{\centering\arraybackslash}p{1.45cm}|>{\centering\arraybackslash}p{1.45cm}|}
\hline
\textbf{Dataset} & \diagbox[dir=NW]{\textbf{Def.}}{\textbf{Per.}} & \textbf{0\%} & \textbf{5\%} & \textbf{10\%} & \textbf{15\%} & \textbf{20\%} & \textbf{25\%}\\
\hline
\multirow{6}{*}{Polblogs} 
& GAT & 94.88$\pm$0.42 & 86.93$\pm$2.39 & 76.89$\pm$2.14 &	71.78$\pm$1.46 & 68.29$\pm$1.83 & 66.99$\pm$1.64 \\
\cline{2-8}
& Ours\textbf{+}GAT & 
\begin{tabular} [c]{@{}l@{}} \textbf{95.12$\pm$0.52} \\ (\textbf{\textcolor{red}{0.24\%$\uparrow$}}) \end{tabular} &  
\begin{tabular} [c]{@{}l@{}} \textbf{92.21$\pm$1.20} \\ (\textbf{\textcolor{red}{5.28\%$\uparrow$}})\end{tabular} & 
\begin{tabular} [c]{@{}l@{}} \textbf{89.94$\pm$1.89} \\ (\textbf{\textcolor{red}{13.05\%$\uparrow$}}) \end{tabular}&  \begin{tabular} [c]{@{}l@{}} \textbf{87.96$\pm$2.09} \\ (\textbf{\textcolor{red}{16.18\%$\uparrow$}}) \end{tabular} & 
\begin{tabular} [c]{@{}l@{}} \textbf{84.17$\pm$2.26} \\ (\textbf{\textcolor{red}{15.88\%$\uparrow$}}) \end{tabular}&  \begin{tabular} [c]{@{}l@{}} \textbf{76.08$\pm$4.80} \\ (\textbf{\textcolor{red}{9.09\%$\uparrow$}}) \end{tabular} \\	
\cline{2-8}
& HANG & 93.72$\pm$1.26 & 88.88$\pm$1.83 & 87.45$\pm$1.21 & 86.53$\pm$1.47 & 81.41$\pm$1.94 & 73.11$\pm$2.60  \\
\cline{2-8}
& Ours\textbf{+}HANG & 
\begin{tabular} [c]{@{}l@{}} \textbf{93.85$\pm$0.75} \\ (\textbf{\textcolor{red}{0.13\%$\uparrow$}}) \end{tabular} &  
\begin{tabular} [c]{@{}l@{}} \textbf{90.73$\pm$1.14} \\ (\textbf{\textcolor{red}{1.85\%$\uparrow$}}) \end{tabular}& 
\begin{tabular} [c]{@{}l@{}} \textbf{89.65$\pm$2.13} \\ (\textbf{\textcolor{red}{2.20\%$\uparrow$}}) \end{tabular} &  \begin{tabular} [c]{@{}l@{}} \textbf{86.67$\pm$2.08} \\ (\textbf{\textcolor{red}{0.14\%$\uparrow$}}) \end{tabular} & 
\begin{tabular} [c]{@{}l@{}} \textbf{82.71$\pm$2.66} \\ (\textbf{\textcolor{red}{1.30\%$\uparrow$}}) \end{tabular} &  \begin{tabular} [c]{@{}l@{}} \textbf{76.11$\pm$4.53} \\ (\textbf{\textcolor{red}{3.00\%$\uparrow$}}) \end{tabular} 
\\
\cline{2-8}
& Mid-GCN & 91.11$\pm$1.82 & 89.44$\pm$3.35 & 84.71$\pm$4.94 & 72.59$\pm$4.10 & 67.16$\pm$1.38 & 64.70$\pm$1.08  \\
\cline{2-8}
& Ours\textbf{+}Mid-GCN & 
\begin{tabular} [c]{@{}l@{}} \textbf{92.15$\pm$2.26} \\ (\textbf{\textcolor{red}{1.04\%$\uparrow$}}) \end{tabular} &  
\begin{tabular} [c]{@{}l@{}} \textbf{90.82$\pm$2.49} \\ (\textbf{\textcolor{red}{1.38\%$\uparrow$}}) \end{tabular}& 
\begin{tabular} [c]{@{}l@{}} \textbf{85.27$\pm$4.67} \\ (\textbf{\textcolor{red}{0.56\%$\uparrow$}}) \end{tabular} &  
\begin{tabular} [c]{@{}l@{}} \textbf{74.29$\pm$1.96} \\ (\textbf{\textcolor{red}{1.70\%$\uparrow$}}) \end{tabular} & 
\begin{tabular} [c]{@{}l@{}} \textbf{70.88$\pm$1.51} \\ (\textbf{\textcolor{red}{3.72\%$\uparrow$}}) \end{tabular} &  
\begin{tabular} [c]{@{}l@{}} \textbf{67.67$\pm$1.39} \\ (\textbf{\textcolor{red}{2.97\%$\uparrow$}}) \end{tabular} \\			         		           
\hline\hline			
\multirow{6}{*}{\makecell[c] {Amazon \\ Photo}} 
& GAT & 93.54$\pm$0.18 & 91.73$\pm$0.39 & 76.53$\pm$31.14 & 69.99$\pm$32.95 & 51.57$\pm$38.85 & 62.53$\pm$33.59 \\
\cline{2-8}
& Ours\textbf{+}GAT & 
\begin{tabular} [c]{@{}l@{}} \textbf{93.70$\pm$0.16} \\ (\textbf{\textcolor{red}{0.16\%$\uparrow$}}) \end{tabular} &  
\begin{tabular} [c]{@{}l@{}} \textbf{91.74$\pm$0.18} \\ (\textbf{\textcolor{red}{0.01\%$\uparrow$}})\end{tabular} & 
\begin{tabular} [c]{@{}l@{}} \textbf{83.65$\pm$20.01} \\ (\textbf{\textcolor{red}{7.12\%$\uparrow$}}) \end{tabular}&  
\begin{tabular} [c]{@{}l@{}} \textbf{76.19$\pm$29.97} \\ (\textbf{\textcolor{red}{6.20\%$\uparrow$}}) \end{tabular} & 
\begin{tabular} [c]{@{}l@{}} \textbf{74.30$\pm$31.89} \\ (\textbf{\textcolor{red}{22.73\%$\uparrow$}}) \end{tabular}&  
\begin{tabular} [c]{@{}l@{}} \textbf{81.05$\pm$25.55} \\ (\textbf{\textcolor{red}{18.52\%$\uparrow$}}) \end{tabular} \\	
\cline{2-8}
& HANG & 93.47$\pm$0.32 & 91.62$\pm$0.54 & 91.51$\pm$0.39 & 90.93$\pm$0.68 & 90.16$\pm$0.56 & 89.92$\pm$0.61	\\
\cline{2-8}
& Ours\textbf{+}HANG & 
\begin{tabular} [c]{@{}l@{}} \textbf{93.63$\pm$0.43} \\ (\textbf{\textcolor{red}{0.16\%$\uparrow$}}) \end{tabular} &  
\begin{tabular} [c]{@{}l@{}} \textbf{92.64$\pm$0.44} \\ (\textbf{\textcolor{red}{1.02\%$\uparrow$}})\end{tabular}&  
\begin{tabular} [c]{@{}l@{}} \textbf{91.72$\pm$0.35} \\ (\textbf{\textcolor{red}{0.21\%$\uparrow$}}) \end{tabular} & 
\begin{tabular} [c]{@{}l@{}} \textbf{91.97$\pm$0.42}\\ (\textbf{\textcolor{red}{1.04\%$\uparrow$}}) \end{tabular} &  
\begin{tabular} [c]{@{}l@{}} \textbf{91.04$\pm$0.39}\\(\textbf{\textcolor{red}{0.88\%$\uparrow$}})\end{tabular}&		
\begin{tabular} [c]{@{}l@{}} \textbf{90.85$\pm$0.73} \\ (\textbf{\textcolor{red}{0.93\%$\uparrow$}}) \end{tabular} 
\\
\cline{2-8}
& Mid-GCN & 77.25$\pm$0.48 & 74.39$\pm$0.49 & 65.95$\pm$0.82 & 57.08$\pm$1.70 & 48.75$\pm$0.88 & 46.75$\pm$1.13	\\
\cline{2-8}
& Ours\textbf{+}Mid-GCN & 
\begin{tabular} [c]{@{}l@{}} \textbf{78.45$\pm$0.50} \\ (\textbf{\textcolor{red}{1.20\%$\uparrow$}}) \end{tabular} &  
\begin{tabular} [c]{@{}l@{}} \textbf{74.81$\pm$0.82} \\ (\textbf{\textcolor{red}{0.42\%$\uparrow$}})\end{tabular}&  
\begin{tabular} [c]{@{}l@{}} \textbf{66.15$\pm$1.14} \\ (\textbf{\textcolor{red}{0.20\%$\uparrow$}}) \end{tabular} & 
\begin{tabular} [c]{@{}l@{}} \textbf{59.47$\pm$1.18}\\ (\textbf{\textcolor{red}{2.39\%$\uparrow$}}) \end{tabular} &  
\begin{tabular} [c]{@{}l@{}} \textbf{51.02$\pm$1.35}\\(\textbf{\textcolor{red}{2.27\%$\uparrow$}})\end{tabular}&		
\begin{tabular} [c]{@{}l@{}} \textbf{47.37$\pm$1.43} \\ (\textbf{\textcolor{red}{0.62\%$\uparrow$}}) \end{tabular} 
\\		
\hline				
\end{tabular}
\end{small}
\vskip -0.1in
\end{table*}
					
\textbf{Taking Equilibrium State of Graph Regime as Booster to Promote Adversarial Resilience.}	Besides the adjacency matrix optimization, other robustness-enhancing mechanisms exist, like graph attention network, ODE-based graph neural flow, mid-frequency signals, etc. To furtherly explore the effectiveness of our work, we at first find out the asymptotically-stable equilibrium point and generate associated topology, then utilize the existing methods to execute node-classification task. Table \ref{Table: EquiliResCombine-1stPart} and Table \ref{Table: EquiliResCombine-2ndPart} (Appendix \ref{Sec: ExtraPerformanceResults}) state the experimental results, wherein the sign "\textbf{+}" indicates our EquiliRes as the preliminary is first executed and then followed by GAT, HANG, and Mid-GCN. 
			
Overall, referring to our proposed EquiliRes as a booster can significantly enhance the performance of existing defense mechanisms, upon which the different-level advancement on the five datasets at APR 0\% unveils that even a clean graph can be enhanced through constructing a stable/equilibrium-state graph topology at first. In the case of existing adversarial perturbations, GAT improves the accuracy by on-average 11.90\% on Polblogs, 10.916\% on Amazon Photo, 4.68\% on Cora\_ML, 2.02\% on Cora, 2.58\% on Citeseer, HANG raises by on-average 1.70\%, 0.82\%, 5.02\%, 1.99\%, 2.55\%, and Mid-GCN promotes by on-average 2.07\%, 1.18\%, 0.42\%, 2.20\%, 0.61\% respectively.             
						
To sum up, we can highlight two aspects from the experimental results: i) our proposed method can significantly promote the adversarial resilience from the perspective of graph topology; and ii) our work also enables to benefit the present defense methods to resist adversarial attacks through building an equilibrium state in advance for the protected graph.
						
\subsubsection{Analytics on rank/singular variation and time overhead}
The studies \cite{JinMa20, ZhaoKang23}  pointed out the attack Metattack could enlarge the rank and singular values of adjacency matrix, to resist that, our experiments showcase EquiliRes can indeed decease them when the graph topology matches the equilibrium-point state. The analytics are detailed in Appendices \ref{Sec:Rank} and \ref{Sec:SingularValue}. At same time, the computation complexity is analyzed in Appendix \ref{Sec:computationOverhead}. 
						
\section{Related work and discussion} \label{Sec: RelatedWork}
We review the related work in Appendix \ref{RelatedWork}. Simultaneously, we also discuss the behind motivation why to bridge the connection between adversarial learning and dynamic systems, along with three open issues from the perspectives of condensed one-dimensional mapping function, adversarial resilience on multi-modal data, and parameters of resilience-declining function in Appendix \ref{Sec: OpenIssue}.     
						
\section{Conclusion} \label{Sec: conclusion}
Upon our proposed hypothesis that each graph regime has a critical state to warrant its adversarial resilience, we first model graph adversarial learning as complex dynamic system, then, resort to the stability theory to find out asymptotically-stable equilibrium point corresponding to the graph's critical state of adversarial resilience. Furthermore, we propose a generalized theoretic framework to articulate the prerequisites that guarantee the existence of such equilibrium point. To satisfy the prerequisites, we embody the graph as a whole into a condensed one-dimensional function to reflect its dynamics under continuous adversarial perturbations, and deduce the equilibrium-point trajectory. In addition, we figure out a set of parameters for the five realistic datasets to appropriately mirror the adversarial effects at different perturbation rates. Finally, referring to such equilibrium-point trajectory, we iteratively modify the adjacency matrix to infer an attack-resilient topology. Multi-facet experiments validate the effectiveness of our approach and the rationality of our hypothesis. Compared to SOTA methods, our work has a remarkable improvement on the robustness. To us, the most important innovation lies in bridging the connection between adversarial learning and complex dynamic system, and provide a provable solution to pinpoint the critical state of adversarial resilience. We believe our work will stimulate more exploration on the intrinsic robustness of graph in future. 
						
%
%


\bibliography{example_paper}
\bibliographystyle{neurips_2025}

%
%
%
%
%
%
%
%


\appendix

\section {Lyapunov criterion} \label{Sec:LyapunovCriterion}
In the domain of system stability, Lyapunov criterion as a classic methodology provides the strict condition to guarantee the origin must be an asymptotically-stable equilibrium point for multi-object dynamic systems. For the detailed proof, refer to the reference \cite{KhalilBook15}.
\begin{theorem} \label{Theorem-1}
	\textbf{(Lyapunov Criterion \cite{KhalilBook15})} Let $ v(\mathord{\buildrel{\lower3pt\hbox{$\scriptscriptstyle\rightharpoonup$}} 
		\over x} ) $ be a locally Lipschitz function over a domain $ \xi  \subset \mathbb{R}^n $, which contains the origin. Define a continuously differentiable function $ V\left( {\mathord{\buildrel{\lower3pt\hbox{$\scriptscriptstyle\rightharpoonup$}} 
			\over x} } \right) $ over domain $\xi$, such that
	\begin{equation}
		\begin{aligned}
			V\left( 0 \right) = 0,_{} V\left( {\mathord{\buildrel{\lower3pt\hbox{$\scriptscriptstyle\rightharpoonup$}} 
					\over x} } \right) > 0 \;\; for \; all \;\; \mathord{\buildrel{\lower3pt\hbox{$\scriptscriptstyle\rightharpoonup$}} 
				\over x}  \in \xi  \;\; and \;\; \mathord{\buildrel{\lower3pt\hbox{$\scriptscriptstyle\rightharpoonup$}} \over x}  \ne 0 
			\\
			\frac{d}{{dt}}V\left( {\mathord{\buildrel{\lower3pt\hbox{$\scriptscriptstyle\rightharpoonup$}} 
					\over x} } \right) < 0 \;\; for \;\; all \;\;   		\mathord{\buildrel{\lower3pt\hbox{$\scriptscriptstyle\rightharpoonup$}} \over x}  \in \xi \;\; and \;\; \mathord{\buildrel{\lower3pt\hbox{$\scriptscriptstyle\rightharpoonup$}} \over x}  \ne 0
		\end{aligned}.   	
	\end{equation}	 
\end{theorem}

This Lyapunov criterion gives a general methodology to prove whether an asymptotically-stable equilibrium point exists or not, upon which, we infer the subsequent theorems in this paper. 

\section{Proof of Theorem \ref{Theorem-2}} \label{Sec:ProofTheorem2}
Pursuant to Theorem \ref{Theorem-1}, a candidate Lyapunov function is defined at first, then the quadratic-form is employed for the linear-part perturbation. For the conciseness, variable $t$ is omitted hereafter, 
\begin{equation}
	V_{Linear} \left( {\mathord{\buildrel{\lower3pt\hbox{$\scriptscriptstyle\rightharpoonup$}} 
			\over x} } \right) = \frac{1}{2}\mathord{\buildrel{\lower3pt\hbox{$\scriptscriptstyle\rightharpoonup$}} 
		\over x} ^T P\mathord{\buildrel{\lower3pt\hbox{$\scriptscriptstyle\rightharpoonup$}}
		\over x}.
\end{equation}

For the nonlinear-part perturbation, a reasonable and valid manner is to use the integral form to present the aggregated effect of continuous perturbations,
\begin{equation}
	V_{Nonlinear} \left( {\mathord{\buildrel{\lower3pt\hbox{$\scriptscriptstyle\rightharpoonup$}} 
			\over x} } \right) = \sum\nolimits_{i = 1}^m {\gamma _i \int_0^{y_i } {\phi _i (\sigma )} d\sigma }. 
\end{equation}

According to Theorem \ref{Theorem-1} and Equation \ref{Equ-1}, we know $V_{Linear}$ ought to be positive definite, this implies the matrix must be $P$ $>$0 and symmetric. The first-order derivative of linear-part perturbation and nonlinear-part perturbation can be respectively calculated as:
\begin{equation} \label{Equ-linearPart}
	\frac{d}{{dt}}V_{Linear} \left( {\mathord{\buildrel{\lower3pt\hbox{$\scriptscriptstyle\rightharpoonup$}} 
			\over x} } \right) = \frac{1}{2}\mathord{\buildrel{\lower3pt\hbox{$\scriptscriptstyle\rightharpoonup$}} 
		\over x} ^T \left( {PA + A^T P} \right)\mathord{\buildrel{\lower3pt\hbox{$\scriptscriptstyle\rightharpoonup$}} 
		\over x} {\rm{ + }}\mathord{\buildrel{\lower3pt\hbox{$\scriptscriptstyle\rightharpoonup$}} 
		\over x} ^T PB\left( { - \phi \left( {\mathord{\buildrel{\lower3pt\hbox{$\scriptscriptstyle\rightharpoonup$}} 
				\over y} } \right)} \right).
\end{equation}

\begin{equation}
	\frac{d}{{dt}}V_{Nonlinear} \left( {\mathord{\buildrel{\lower3pt\hbox{$\scriptscriptstyle\rightharpoonup$}} 
			\over x} } \right) = \phi ^T \left( {\mathord{\buildrel{\lower3pt\hbox{$\scriptscriptstyle\rightharpoonup$}} 
			\over y} } \right)\psi C\left( {A\mathord{\buildrel{\lower3pt\hbox{$\scriptscriptstyle\rightharpoonup$}} 
			\over x}  + B\left( { - \phi \left( {\mathord{\buildrel{\lower3pt\hbox{$\scriptscriptstyle\rightharpoonup$}} 
					\over y} } \right)} \right)} \right). 
\end{equation}

Referring to the aforementioned Lyapunov Criterion, $(PA + A^T P)$ is anticipated to be negative definite, to prove this point, the following theorem is provided.

\begin{theorem} \label{Theorem-3}
	For any given positive definite and symmetric matrix $\tilde N$, and its decomposed nonsingular matrix $ \hat{N}$, $ PA + A^T P = - \hat{N}^T \hat{N} $ has a unique positive definite and symmetric solution $P$ if all eigenvalues of matrix $A$ have negative-real parts. 
\end{theorem}

\textit{Proof.} See Appendix \ref{Sec:ProofTheorem3}.

Theorem \ref{Theorem-3} guarantees the negative definite of first part of Equation \ref{Equ-linearPart}. Next, turn to the rest part $      \mathord{\buildrel{\lower3pt\hbox{$\scriptscriptstyle\rightharpoonup$}} 
	\over x} ^T PB\left( { - \phi \left( {\mathord{\buildrel{\lower3pt\hbox{$\scriptscriptstyle\rightharpoonup$}} 
			\over y} } \right)} \right) $ and $
\frac{d}{{dt}}V_{Nonlinear} \left( {\mathord{\buildrel{\lower3pt\hbox{$\scriptscriptstyle\rightharpoonup$}} 
		\over x} } \right) $. Similarly, we also attempt to match a quadratic form,
\begin{equation} \label{Equ-quadratic}
	\begin{array}{l}
		\frac{d}{{dt}}V_{} \left( {\mathord{\buildrel{\lower3pt\hbox{$\scriptscriptstyle\rightharpoonup$}} 
				\over x} } \right) = \frac{1}{2}\mathord{\buildrel{\lower3pt\hbox{$\scriptscriptstyle\rightharpoonup$}} 
			\over x} ^T \left( { - \hat{N}^T \hat{N}} \right)\mathord{\buildrel{\lower3pt\hbox{$\scriptscriptstyle\rightharpoonup$}} 
			\over x} {\rm{ + }}\mathord{\buildrel{\lower3pt\hbox{$\scriptscriptstyle\rightharpoonup$}} 
			\over x} ^T PB\left( { - \phi \left( {\mathord{\buildrel{\lower3pt\hbox{$\scriptscriptstyle\rightharpoonup$}} 
					\over y} } \right)} \right) \\ 
		\begin{array}{*{20}c}
			{}  \\
		\end{array}\begin{array}{*{20}c}
			{} & {} & {}  \\
		\end{array} + \phi \left( {\mathord{\buildrel{\lower3pt\hbox{$\scriptscriptstyle\rightharpoonup$}} 
				\over y} } \right)^T \psi C\left( {A\mathord{\buildrel{\lower3pt\hbox{$\scriptscriptstyle\rightharpoonup$}} 
				\over x}  + B\left( { - \phi \left( {\mathord{\buildrel{\lower3pt\hbox{$\scriptscriptstyle\rightharpoonup$}} 
						\over y} } \right)} \right)} \right) \\ 
		\begin{array}{*{20}c}
			{} & {} & {}  \\
		\end{array} =  - \frac{1}{2}\left( {\hat{N}\mathord{\buildrel{\lower3pt\hbox{$\scriptscriptstyle\rightharpoonup$}} 
				\over x}  - W\phi \left( {\mathord{\buildrel{\lower3pt\hbox{$\scriptscriptstyle\rightharpoonup$}} 
					\over y} } \right)} \right)^T \left( {\hat{N}\mathord{\buildrel{\lower3pt\hbox{$\scriptscriptstyle\rightharpoonup$}} 
				\over x}  - W\phi \left( {\mathord{\buildrel{\lower3pt\hbox{$\scriptscriptstyle\rightharpoonup$}} 
					\over y} } \right)} \right) \\ 
		\begin{array}{*{20}c}
			{}  \\
		\end{array}\begin{array}{*{20}c}
			{} & {} & {}  \\
		\end{array} - \phi \left( {\mathord{\buildrel{\lower3pt\hbox{$\scriptscriptstyle\rightharpoonup$}} 
				\over y} } \right)^T \left[ {\mathord{\buildrel{\lower3pt\hbox{$\scriptscriptstyle\rightharpoonup$}} 
				\over y}  - M\phi \left( {\mathord{\buildrel{\lower3pt\hbox{$\scriptscriptstyle\rightharpoonup$}} 
					\over y} } \right)} \right] \\
	\end{array}.
\end{equation}

On account of the negative semidefinite of  $
\frac{d}{{dt}}V_{} \left( {\mathord{\buildrel{\lower3pt\hbox{$\scriptscriptstyle\rightharpoonup$}} \over x} } \right) $, we deduce the following theorem.

\begin{theorem} \label{Theorem-4}
	There exist a diagonal matrix $M$, such that the perturbation output $      \mathord{\buildrel{\lower3pt\hbox{$\scriptscriptstyle\rightharpoonup$}} 
		\over y} $ and its associated nonlinear-mapping input $
	\phi \left( {\mathord{\buildrel{\lower3pt\hbox{$\scriptscriptstyle\rightharpoonup$}} \over y} } \right) $ satisfying $
	\mathord{\buildrel{\lower3pt\hbox{$\scriptscriptstyle\rightharpoonup$}} 
		\over \phi } \left( y \right)^T  \cdot \left[ {\mathord{\buildrel{\lower3pt\hbox{$\scriptscriptstyle\rightharpoonup$}} 
			\over y}  - M\mathord{\buildrel{\lower3pt\hbox{$\scriptscriptstyle\rightharpoonup$}} 
			\over \phi } \left( y \right)} \right] \ge  0 $, a matrix $W$ and a diagonal matrix $\psi$, such that $ PB =  - \hat{N}^T W + A^T C^T \psi  + C^T $, $W^T W = \psi CB + 2M + B^T C^T \psi $.
\end{theorem}

\textit{Proof.} See Appendix \ref{Sec:ProofTheorem4}.

Theorem \ref{Theorem-3} and Theorem \ref{Theorem-4} warrant the realization of Equation \ref{Equ-quadratic}, namely the first-order derivative of candidate Lyapunov function is negative semidefinite. To finally warrant it converges into the asymptotically-stable equilibrium point, we anticipate it to be negative definite categorically,
\begin{equation}
	\frac{d}{{dt}}V_{} \left( {\mathord{\buildrel{\lower3pt\hbox{$\scriptscriptstyle\rightharpoonup$}} 
			\over x} } \right) <  - \frac{1}{2}\varepsilon \mathord{\buildrel{\lower3pt\hbox{$\scriptscriptstyle\rightharpoonup$}} 
		\over x} ^T P\mathord{\buildrel{\lower3pt\hbox{$\scriptscriptstyle\rightharpoonup$}} 
		\over x} ,
\end{equation}
where $\varepsilon$ is an arbitrarily small positive number. To achieve this requirement, Equation \ref{Equ-quadratic} must satisfy $	PA + A^T P =  - \hat N^T \hat N - \varepsilon P$. Now, we need prove the existence of such an $\varepsilon$. According to Theorem \ref{Theorem-3}, we know if ($PA + A^T P$) is positive definite and symmetric, then it can be equal to the multiplication of a nonsingular matrix by its transpose. That is to say, we need prove $ \hat N^T \hat N + \varepsilon P $ is positive definite (obviously symmetric), thus we have
\begin{equation}
	\begin{array}{l}
		\mathord{\buildrel{\lower3pt\hbox{$\scriptscriptstyle\rightharpoonup$}} 
			\over x} ^T \hat N^T \hat N\mathord{\buildrel{\lower3pt\hbox{$\scriptscriptstyle\rightharpoonup$}} 
			\over x}  + \varepsilon \int_0^t {\mathord{\buildrel{\lower3pt\hbox{$\scriptscriptstyle\rightharpoonup$}} 
				\over x} ^T e^{\left( {A^T \sigma } \right)} \hat N^T \hat Ne^{\left( {A\sigma } \right)} \mathord{\buildrel{\lower3pt\hbox{$\scriptscriptstyle\rightharpoonup$}} 
				\over x} d\sigma }  > 0 \\ 
		\begin{array}{*{20}c}
			{} & {}  \\
		\end{array}\varepsilon  >  - \frac{{\left\| {\mathord{\buildrel{\lower3pt\hbox{$\scriptscriptstyle\rightharpoonup$}} 
						\over x} \hat N} \right\|_2 ^2 }}{{\int_{t_0 }^t {\left\| {\mathord{\buildrel{\lower3pt\hbox{$\scriptscriptstyle\rightharpoonup$}} 
							\over x} e^{\left( {A\sigma } \right)} \hat N} \right\|_2 ^2 d\sigma } }} \\ 
	\end{array}.
\end{equation}
Visibly, there indeed exists such a positive number $ \varepsilon  \in \left( {0,_{} \infty } \right)	$. Of course, the interval can be extended to $\left( {{\raise0.7ex\hbox{${ - \left\| {\mathord{\buildrel{\lower3pt\hbox{$\scriptscriptstyle\rightharpoonup$}} \over x} \hat N} \right\|_2 ^2 }$} \!\mathord{\left/
			{\vphantom {{ - \left\| {\mathord{\buildrel{\lower3pt\hbox{$\scriptscriptstyle\rightharpoonup$}} 
								\over x} \hat N} \right\|_2 ^2 } {\int_{t_0 }^t {\left\| {\mathord{\buildrel{\lower3pt\hbox{$\scriptscriptstyle\rightharpoonup$}} 
									\over x} e^{\left( {A\sigma } \right)} \hat N} \right\|_2 ^2 d\sigma } }}}\right.\kern-\nulldelimiterspace}
		\!\lower0.7ex\hbox{${\int_{t_0 }^t {\left\| {\mathord{\buildrel{\lower3pt\hbox{$\scriptscriptstyle\rightharpoonup$}} 
							\over x} e^{\left( {A\sigma } \right)} \hat N} \right\|_2 ^2 d\sigma } }$}},_{} \infty } \right) $, concretely, it is subject to the eigenvalues of matrix $A$.

\section{Proof of Theorem \ref{Theorem-3}} \label{Sec:ProofTheorem3}
At first, establish the connection to matrix $A$ with time variable $t$ and assume an integral form for matrix $P$, 
\begin{equation}
	P = \int_{\rm{0}}^t {e^{\left( {A^T \sigma } \right)} \tilde Ne^{\left( {A\sigma } \right)} d\sigma }. 
\end{equation}

Then, we have that
\begin{equation}
	\begin{array}{l}
		PA + A^T P \\ 
		= \int_{\rm{0}}^t {e^{\left( {A^T \sigma } \right)} \tilde Ne^{\left( {A\sigma } \right)} Ad\sigma }  + \int_{\rm{0}}^t {A^T e^{\left( {A^T \sigma } \right)} \tilde Ne^{\left( {A\sigma } \right)} d\sigma }  \\ 
		= \int_{\rm{0}}^t {\frac{d}{{dt}}\left( {e^{\left( {A^T \sigma } \right)} \tilde Ne^{\left( {A\sigma } \right)} } \right)d\sigma }  \\ 
		= \left. {e^{\left( {A^T \sigma } \right)} \tilde Ne^{\left( {A\sigma } \right)} } \right|_{\rm{0}}^t  \\ 
	\end{array}.
\end{equation}

Let $\lambda _i$ ($i$=1, $\cdots$, $n$) be the eigenvalues of matrix $A$ with associated eigenvector $v_i$, thus $A$ can be diagonalized as: $\Lambda  = \hat M^{ - 1} A \hat M$, where $ \hat M = \left[ {v{}_1,_{} v_2 ,_{}  \cdots ,_{} v_n } \right]$, $\Lambda  = diag\left\{ {\lambda _1 ,_{}  \cdots ,_{} \lambda _n } \right\}$.

Next, calculate $e^{\left( {At} \right)}$ using Taylor series, 
\begin{equation}
	\begin{array}{l}
		e^{\left( {At} \right)}  = I + tA + \frac{{t^2 }}{{2!}}A^2  +  \cdots  = \sum\limits_{k = 0}^\infty  {\frac{1}{{k!}}t^k A^k }  \\ 
		\begin{array}{*{20}c}
			{} & {}  \\
		\end{array} = MM^{ - 1}  + tM\Lambda M^{ - 1}  + \frac{{t^2 }}{{2!}}M\Lambda ^2 M^{ - 1}  +  \cdots  \\ 
	\end{array},
\end{equation}

thus,
\begin{equation}
	\begin{array}{l}
		M^{ - 1} e^{\left( {At} \right)} M = I + t\Lambda  + \frac{{t^2 }}{{2!}}\Lambda ^2  +  \cdots  \\ 
		\begin{array}{*{20}c}
			{} & {}  \\
		\end{array}\begin{array}{*{20}c}
			{} & {}  \\
		\end{array} = \left( {\begin{array}{*{20}c}
				{e^{\lambda _1 t} } & {} & 0  \\
				{} &  \ddots  & {}  \\
				0 & {} & {e^{\lambda _n t} }  \\
		\end{array}} \right) = e^{\left( {\Lambda t} \right)}  \\ 
	\end{array}.
\end{equation}

Since all the eigenvalues of $A$ have negative-real parts, thus we have that $e^{\left( {At} \right)}  \to 0$ when $t \to \infty$, that is to say, through a long-time adversarial learning, $PA + A^T P$ will equal to $- \tilde N$. Due to the symmetry and positive-definite property of matrix $\tilde N$, it can be decomposed into the multiplication of one nonsingular matrix by its transpose\footnote{$[$\cite{ChenBook99}, Theorem 3.7$]$}, i.e. $\tilde N = \hat{N}^T \hat{N}$, which further indicates the symmetry of matrix $P$. Next, judge its definite property,
\begin{equation}
	\begin{array}{l}
		x^T Px = \int_{\rm{0}}^t {x^T e^{\left( {A^T \sigma } \right)} \hat{N}^T \hat{N}e^{\left( {A\sigma } \right)} xdt}  \\ 
		\begin{array}{*{20}c}
			{} & {}  \\
		\end{array} = \int_{\rm{0}}^t {\left\| {x^T e^{\left( {A^T \sigma } \right)} \hat{N}^T } \right\|_2 ^2 d\sigma  > 0}  \\ 
	\end{array}.
\end{equation}

Thus, there exists a unique positive definite solution $P$.

\section{Proof of Theorem \ref{Theorem-4}} \label{Sec:ProofTheorem4}
We first introduce a theorem\footnote{$[$\cite{Youla61}, Theorem 2$]$}, namely, for a $p \times p$	rational positive-real and Hurwitz\footnote{All the eigenvalues are strictly in the left half plane.} matrix $ Z\left( s \right)$, there exists an $r \times p$ Hurwitz matrix $Q\left( s \right)$ such that
\begin{equation} \label{Equ-A1}
	\tilde Z\left( s \right) = Z(s) + Z^T ( - s) = Q^T ( - s)Q(s),
\end{equation}
where $\hat r$ is the rank of $ Z(s) + Z^T ( - s)$ over the field of rational functions of $s$, and rank $Q\left( s \right)$=$\hat r$ for $Res[s] > 0$.

To prove the Theorem \ref{Theorem-4}, we need extend $\mathord{\buildrel{\lower3pt\hbox{$\scriptscriptstyle\rightharpoonup$}} \over y}  = C\mathord{\buildrel{\lower3pt\hbox{$\scriptscriptstyle\rightharpoonup$}} \over x}  + D\mathord{\buildrel{\lower3pt\hbox{$\scriptscriptstyle\rightharpoonup$}} \over u} $, where $ \mathord{\buildrel{\lower3pt\hbox{$\scriptscriptstyle\rightharpoonup$}} \over u} $ can be seen as a flexible adjustment to the output $\mathord{\buildrel{\lower3pt\hbox{$\scriptscriptstyle\rightharpoonup$}} \over y}$ by design, it can be degraded to the original if $D$=0.   

Let $\left\{ {{\rm{\bar A, \bar B, \bar C, \bar D}}} \right\}$ be a minimal realization\footnote{For a given $ Z\left( s \right)$, there exist many sets  $\left\{ {{\rm{\bar A, \bar B, \bar C, \bar D}}} \right\}$ to realize, however, there must be a minimal-dimension of matrix ${\rm{\bar A}}$, such a realization associated with minimal dimension ${\rm{\bar A}}$ is termed a minimal realization.} of positive-real function matrix $ Z\left( s \right)$. Pursuant to Equation \ref{Equ-A1}, there exists an $\hat r \times p$ function matrix $Q\left( s \right)$ satisfying it, and let $ \left\{ {F,_{} G,_{} L,_{} J} \right\}$ be a minimal realization of $Q\left( s \right)$, thus, the minimal realization of $ Q^T \left( s \right)$ is $\left\{ { - F^T ,_{} L^T ,_{}  - G^T ,_{} J^T } \right\}$. Then, the realization of $ Q^T \left( { - s} \right) \cdot Q\left( s \right)$ can be furtherly inferred as:
\begin{equation}
	\begin{array}{l}
		\left\{ {{\rm{\hat A, \hat B, \hat C, \hat D}}} \right\} =  
		\left\{ {\left( {\begin{array}{*{20}c}
					F & 0  \\
					{L^T L} & { - F^T }  \\
			\end{array}} \right),\left( \begin{array}{l}
				G \\ 
				L^T J \\ 
			\end{array} \right),\left( {J^T L, - G^T } \right),J^T J} \right\} \\ 
	\end{array}.
\end{equation}

Since both $Z\left( s \right)$ and $Z^T \left( { - s} \right)$ are Hurwitz, they have no common poles. Referring to the claim $[$\cite{Duffin63}, Theorem 7$]$, we have $	\delta \left[ {Z\left( s \right) + Z^T ( - s)} \right] = \delta [Z\left( s \right)] + \delta [Z^T \left( s \right)] = 2\delta [Z\left( s \right)]$, where $\delta [Z\left(  \cdot  \right)]$ denotes the degree of matrix $Z\left(  \cdot  \right)$. Therefore, $\left\{ {{\rm{\hat A, \hat B, \hat C, \hat D}}} \right\}$ is the minimal realization of $Q^T (- s)Q(s)$. From the claim $[$\cite{ChenBook99}, Theorem 7.2$]$, we know $\left( {{\rm{\hat A, \hat B}}} \right)$ is controllable and $\left( {{\rm{\hat A, \hat C}}} \right)$ is observable.

Given $(F, L)$ is observable and $F$ is Hurwitz, referring to Theorem \ref{Theorem-3}, there must exist a unique symmetric solution $K$, such that $ KF + F^T K =  - L^T L$. According to the proposition $[$\cite{HoKalman66}, Proposition 3$]$, there must exist a nonsingular $T$ to achieve a similar transformation from a minimal realization $\left\{ {{\rm{\hat A, \hat B, \hat C}}{\rm{ \hat D}}} \right\}$ to another minimal realization $\left\{ {{\rm{T\hat AT}}^{{\rm{ - 1}}} {\rm{, T\hat B, \hat CT}}^{{\rm{ - 1}}} {\rm{, \hat D}}} \right\}$. Use
$\begin{matrix}
	T = \left( {\begin{array}{*{20}c}
			I & 0  \\
			K & I  \\
	\end{array}} \right)
\end{matrix}$
to make a new minimal realization,
\begin{equation}
	\begin{array}{l}
		\left\{ {{\rm{\hat A}}_{\rm{1}} {\rm{, \hat B}}_{\rm{1}} {\rm{, \hat C}}_{\rm{1}} {\rm{, \hat D}}_{\rm{1}} } \right\} =  
		\left\{ {\left( {\begin{array}{*{20}c}
					F & 0  \\
					0 & { - F^T }  \\
			\end{array}} \right),\left( \begin{array}{l}
				G \\ 
				KG + L^T J \\ 
			\end{array} \right),} \right. 
		\left. {\left( {J^T L + G^T K, - G^T } \right),J^T J} \right\} \\ 
	\end{array}.
\end{equation}

Equivalently, we directly calculate the minimal realization of $ Z(s) + Z^T ( - s)$,
\begin{equation}
	\begin{array}{l}
		\left\{ {{\rm{\hat A}}_{\rm{2}} {\rm{, \hat B}}_{\rm{2}} {\rm{, \hat C}}_{\rm{2}} {\rm{, \hat D}}_{\rm{2}} } \right\} =  
		\left\{ {\left( {\begin{array}{*{20}c}
					{\bar A} & 0  \\
					0 & { - \bar A^T }  \\
			\end{array}} \right),_{} \left( \begin{array}{l}
				{\bar B} \\ 
				\bar C^T  \\ 
			\end{array} \right),_{} \left( {\bar C, - \bar B^T } \right),_{} \bar D + \bar D^T } \right\}. \\ 
	\end{array}
\end{equation}

Given the eigenvalues of matrix $A$ are all in the left-half plane (Hurwitz), and the eigenvalues of $-A^T$ is all in the right-half plane, indicating no common pole exists. Hence, we conclude the realization $ \left\{ {{\rm{\hat A}}_{\rm{2}} {\rm{, \hat B}}_{\rm{2}} {\rm{, \hat C}}_{\rm{2}} {\rm{, \hat D}}_{\rm{2}} } \right\}$ is minimal as well. Therefore, there must exist a nonsingular matrix $R$ such that $\left\{ {{\rm{\hat A}}_{\rm{1}} {\rm{ = R\hat A}}_{\rm{2}} {\rm{R}}^{{\rm{ - 1}}} {\rm{, \hat B}}_{\rm{1}} {\rm{ = R\hat B}}_{\rm{2}} {\rm{, \hat C}}_{\rm{1}} {\rm{ = \hat C}}_{\rm{2}} {\rm{R}}^{{\rm{ - 1}}} {\rm{, \hat D}}_{\rm{1}} {\rm{ = \hat D}}_{\rm{2}} } \right\}$. Suppose $R$ is a block matrix as follows:
\begin{equation}
\begin{aligned}
\left( {\begin{array}{*{20}c}
	{R_{11} } & {R_{12} }  \\
	{R_{21} } & {R_{22} }  \\
\end{array}} \right) \nonumber
\end{aligned}	
\end{equation}
	
Thus, we can deduce that $ {\rm{ - R}}_{{\rm{12}}} {\rm{\bar A}}^{\rm{T}} {\rm{ = FR}}_{12} $. Then premultiplying by $\exp \left( {Ft} \right)$ and postmultiplying by $\exp \left( {A^T t} \right)$, then we have 
\begin{equation}
\begin{array}{l}
\frac{d}{{dt}}\left( {\exp \left( {Ft} \right)R_{12} \exp (\bar A^T t)} \right) =  \\ 
\exp \left( {Ft} \right)\left[ {FR_{12}  + R_{12} \bar A^T } \right]\exp (\bar A^T t) = 0 \\ 
\end{array}.
\end{equation}
	
Thus, $\exp \left( {Ft} \right)R_{12} \exp (\bar A^T t)$ is a constant for all $t$. Furthermore, $R_{12}  = \exp \left( {Ft} \right)R_{12} \exp (\bar A^T t)$ at $t$=0, also $ R_{12}  = \exp \left( {Ft} \right)R_{12} \exp (\bar A^T t) \to 0$ at $ t \to \infty$, thus $R_{12}$=0. The sequel equations can be inferred, i.e., $F = R_{11} \bar AR_{11}^{ - 1}$, $ G = R_{11} \bar B$, $J^T L + G^T K = \bar CR_{11} ^{ - 1}$, $R_{22} ^T G = \bar B$. 
	
Assume $P = R_{11} ^T KR_{11}^{}$, $\hat{N}= LR_{11}$, $W = J$, then we can infer that
\begin{equation}  \label{Equ-A3}
\begin{split}
P\bar B = \bar C^T  - \hat{N}^T W \\
W^T W = \bar D + \bar D^T
\end{split}.
\end{equation}
	
On the basis of Equation \ref{Equ-TransferFunction}, we redefine a new transfer function $\mathord{\buildrel{\lower3pt\hbox{$\scriptscriptstyle\smile$}} \over G} \left( s \right)$ as,
\begin{equation} \label{Equ-A2}
	\begin{array}{l}
		\mathord{\buildrel{\lower3pt\hbox{$\scriptscriptstyle\smile$}} 
			\over G} \left( s \right) = M + \left( {I + s\psi } \right)G\left( s \right) \\ 
		= M + C\left( {sI - A} \right)^{ - 1} B + \psi C\left( {sI - A + A} \right)\left( {sI - A} \right)^{ - 1} B \\ 
		= M + \left( {C + \psi CA} \right)\left( {sI - A} \right)^{ - 1} B + \psi CB \\ 
	\end{array}.
\end{equation}
	
Let $\lambda _k $ be an eigenvalue of matrix $A$ and $v_k$ be the associated eigenvector. Then, 
\begin{equation}
	\left( {C + \psi CA} \right)v_k  = \left( {C + \psi C\lambda _k } \right)v_k  = \left( {I + \lambda _k \psi } \right)Cv_k. 
\end{equation}
	
The condition implies ($A$, $C$) is observable. Naturally, Equation \ref{Equ-A2} can be recognized as a quadri-tuple $\left\{ {A',B',C',D'} \right\}$, namely $A' = A$, $B' = B$, $C' = \left( {C + \psi CA} \right)$, $D' = M + \psi CB$. If $\mathord{\buildrel{\lower3pt\hbox{$\scriptscriptstyle\smile$}} 
	\over G} \left( s \right)$ is strictly positive real and Hurwitz, referring to the conclusive Equation \ref{Equ-A3}, we can easily obtain the results $PB =  - \hat{N}^T W + A^T C^T \psi  + C^T$, $ W^T W = \psi CB + 2M + B^T C^T \psi$. That is to say, there exist a matrix $W$ and a diagonal matrix $\psi$ that make Equation \ref{Equ-quadratic} correct. 

\section{Deduction of one-dimensional mapping function} \label{Sec:oneDimenFunctionDeduce}
Let $
\Gamma ^{in}  = \left\{ {\left. {\eta _i^{in} } \right|\forall i \in [1,N],_{} \eta _i^{in}  = \sum\nolimits_{j = 1}^N {\tilde B_{\left( {ij} \right)} } } \right\} $ and $ \vartheta _j \left( {x_i } \right) = \varphi \left( {x_i ,x_j } \right) $, where $
\tilde B $ is the adjacency matrix, then 
\begin{equation}
	\sum\nolimits_{j = 1}^N {\tilde B_{\left( {ij} \right)} \varphi \left( {x_i ,x_j } \right)}  = \eta _i^{in} \left\langle {\vartheta _j \left( {x_i } \right)} \right\rangle, 
\end{equation}
where $	\left\langle {\vartheta _j \left( {x_i } \right)} \right\rangle $ denotes the average of PNIB-caused perturbation by node $i$'s neighbor $j$. Pursuant to Heterogeneous Mean-Field (HMF) approximation theory \cite{Kundu22}, node $i$'s in-neighbor $j$ provides weight proportional to node $j$'s out-degree, then we have the weighted average of $\vartheta _j \left( {x_i } \right) $ for node $i$ 
\begin{equation}
	\left\langle {\vartheta _j \left( {x_i } \right)} \right\rangle _{HMF}  = \frac{{\frac{1}{N}\sum\nolimits_{j = 1}^N {\eta _j^{out} \vartheta _j \left( {x_i } \right)} }}{{\frac{1}{N}\sum\nolimits_{j = 1}^N {\eta _j^{out} } }},
\end{equation}
where $\Gamma ^{out}  = \left\{ {\left. {\eta _j^{out} } \right|\forall j \in [1,N]_{}, \eta _j^{out}  = \sum\nolimits_{i = 1}^N {\tilde B_{\left( {ij} \right)} } } \right\} $ is the weighted out-degree of node $j$. As reported in \cite{Kundu22}, we can derive all the nodes have statistically identical perturbation affection if the degree correlation is small. Thus, the vector of adversarial perturbation can be approximated as
\begin{equation} \label{Equ-nodePerturbation}
	\Psi \left( \vartheta  \right) = \frac{{\textbf{1}^T \tilde B\vartheta }}{{\textbf{1}^T \tilde B1}} = \frac{{\frac{1}{N}\sum\nolimits_{j = 1}^N {\eta _j^{out} \vartheta _j } }}{{\frac{1}{N}\sum\nolimits_{j = 1}^N {\eta _j^{out} } }} = \frac{{\left\langle {\Gamma ^{out} \vartheta } \right\rangle }}{{\left\langle {\Gamma ^{out} } \right\rangle }}.
\end{equation}

From the angle of graph structure, we can presume the variance of components of node set (vectorized as $ \mathord{\buildrel{\lower3pt\hbox{$\scriptscriptstyle\rightharpoonup$}} \over x} $) is small. Using HMF approximation theory, we rewrite Equation \ref{Equ-Mapping} as 
\begin{equation}
	\frac{{d\mathord{\buildrel{\lower3pt\hbox{$\scriptscriptstyle\rightharpoonup$}} 
				\over x} }}{{dt}} = \tilde A_{\left(  \cdot  \right)} \chi \left( {\mathord{\buildrel{\lower3pt\hbox{$\scriptscriptstyle\rightharpoonup$}} 
			\over x} } \right) + \overrightarrow {\Gamma ^{in} }  \odot \varphi \left( {\mathord{\buildrel{\lower3pt\hbox{$\scriptscriptstyle\rightharpoonup$}} 
			\over x} ,\Psi \left( {\mathord{\buildrel{\lower3pt\hbox{$\scriptscriptstyle\rightharpoonup$}} 
				\over x} } \right)} \right),
\end{equation}
where the sign $\odot$ denotes Hadamard product, i.e. multiply two vectors term by term. 
There are ($N$+1) variables with the join of $ \Psi \left( {\mathord{\buildrel{\lower3pt\hbox{$\scriptscriptstyle\rightharpoonup$}} \over x} } \right) $,
\begin{equation}
	\begin{array}{l}
		\frac{{d\Psi \left( {\mathord{\buildrel{\lower3pt\hbox{$\scriptscriptstyle\rightharpoonup$}} 
						\over x} } \right)}}{{dt}} = \Psi \left( {\tilde A_{\left(  \cdot  \right)} \chi \left( {\mathord{\buildrel{\lower3pt\hbox{$\scriptscriptstyle\rightharpoonup$}} 
					\over x} } \right) + \mathord{\buildrel{\lower3pt\hbox{$\scriptscriptstyle\rightharpoonup$}} 
				\over \Gamma } ^{in}  \odot \varphi \left( {\mathord{\buildrel{\lower3pt\hbox{$\scriptscriptstyle\rightharpoonup$}} 
					\over x} ,\Psi \left( {\mathord{\buildrel{\lower3pt\hbox{$\scriptscriptstyle\rightharpoonup$}} 
						\over x} } \right)} \right)} \right) \\ 
		\approx \tilde A_{\left(  \cdot  \right)} \chi \left( {\Psi \left( {\mathord{\buildrel{\lower3pt\hbox{$\scriptscriptstyle\rightharpoonup$}} 
					\over x} } \right)} \right) + \Psi \left( {\mathord{\buildrel{\lower3pt\hbox{$\scriptscriptstyle\rightharpoonup$}} 
				\over \Gamma } ^{in} } \right)\varphi \left( {\Psi \left( {\mathord{\buildrel{\lower3pt\hbox{$\scriptscriptstyle\rightharpoonup$}} 
					\over x} } \right),\Psi \left( {\mathord{\buildrel{\lower3pt\hbox{$\scriptscriptstyle\rightharpoonup$}} 
					\over x} } \right)} \right) \\ 
	\end{array}.
\end{equation} 

Thus, let $ \tilde x = \Psi \left( {\mathord{\buildrel{\lower3pt\hbox{$\scriptscriptstyle\rightharpoonup$}} 
		\over x} } \right)$, $
\tilde \beta  = \Psi \left( {\mathord{\buildrel{\lower3pt\hbox{$\scriptscriptstyle\rightharpoonup$}} 
		\over \Gamma } ^{in} } \right)$, we can infer the one-dimensional condensed Equation \ref{Equ-OneDimen}.

\section{Proof of Theorem \ref{Theorem-5}} \label{Sec:ProofTheorem5}
Deriving from Theorem \ref{Theorem-2}, we know that $ \phi \left( {\tilde x} \right) = \frac{{H\tilde x_{}^2 }}{{\theta \tilde x_{}^2  + 1}} $, and $ \tilde x = \Psi \left( {\mathord{\buildrel{\lower3pt\hbox{$\scriptscriptstyle\rightharpoonup$}} 
\over x} } \right) $ is positive, thus $ \phi \left( {\tilde x} \right)^T  > 0 $, $	\tilde x - M\phi \left( {\tilde x} \right) \ge 0 $. Furtherly, $ \frac{{\phi (\tilde x)}}{{\tilde x}} = \frac{{H\tilde x}}{{\theta \tilde x_{}^2  + 1}} \le k $, therefore, let the first-order derivative equal to zero and compute the maximum value, we have $ \frac{{H\tilde x}}{{\theta \tilde x_{}^2  + 1}} \le H \cdot \left( {2\sqrt \theta  } \right)^{ - 1} $, thus $ k \ge H \cdot \left( {2\sqrt \theta  } \right)^{ - 1} $.
	
\section{Proof of Theorem \ref{Theorem-6}} \label{Sec:ProofTheorem6}
According to Theorem \ref{Theorem-2}, we have $\mathord{\buildrel{\lower3pt\hbox{$\scriptscriptstyle\smile$}} 
\over G} \left( s \right) = \frac{1}{k} + \left( {1 + s\gamma } \right)G\left( s \right) = \frac{1}{k} + \left( {1 - \gamma a} \right)\left( {s + a} \right)^{ - 1} \tilde \beta  + \gamma \tilde \beta $. First, the $ \det \left[ {\mathord{\buildrel{\lower3pt\hbox{$\scriptscriptstyle\smile$}} 
\over G} (s) + \mathord{\buildrel{\lower3pt\hbox{$\scriptscriptstyle\smile$}} 
\over G} ^T ( - s)} \right] $ is obviously not identically zero. Since $a$ is positive, then $ \mathord{\buildrel{\lower3pt\hbox{$\scriptscriptstyle\smile$}} \over G} (s) $ is Hurwitz, that is, the poles of all elements of $ \mathord{\buildrel{\lower3pt\hbox{$\scriptscriptstyle\smile$}} \over G} (s) $ have negative real parts. Pursuant to the claim $[$\cite{KhalilBook02}, Theorem 6.1$]$, we calculate
\begin{equation}
\mathord{\buildrel{\lower3pt\hbox{$\scriptscriptstyle\smile$}} 
\over G} \left( {j\omega } \right) + \mathord{\buildrel{\lower3pt\hbox{$\scriptscriptstyle\smile$}} 
\over G} ^T \left( { - j\omega } \right){\rm{ = }}\frac{2}{k} + \frac{{{\rm{2}}\tilde \beta \left( {a + \gamma \omega ^2 } \right)}}{{a^2  + \omega ^2 }} . 
\end{equation}  
Thereby, it is positive definite for all $\omega  \in \mathbb{R}$, and $\mathord{\buildrel{\lower3pt\hbox{$\scriptscriptstyle\smile$}} \over G} \left( s \right) $ is strictly positive real, which further implies Equation \ref{Equ-parameter} has asymptotically-stable equilibrium point. 	
	
\section {Implementation algorithms}  \label{Sec:Algorithms1-2}
\begin{algorithm} [hbtp]
\caption{Graph Parameters Calculation}
\label{Algorithm-1}
\KwIn{Adjacency matrix ${\mathord{\buildrel{\lower3pt\hbox{$\scriptscriptstyle\frown$}} \over A}}$} 
\KwOut{Graph state parameters $\tilde \beta$, $\tilde x$}
\textbf{Procedure calculate$\_$parameters (${\mathord{\buildrel{\lower3pt\hbox{$\scriptscriptstyle\frown$}} \over A}}$)} \\
inDeg $\leftarrow$ sum(${\mathord{\buildrel{\lower3pt\hbox{$\scriptscriptstyle\frown$}} \over A}}$, axis=0) \\
outDeg $\leftarrow$ sum(${\mathord{\buildrel{\lower3pt\hbox{$\scriptscriptstyle\frown$}} \over A}}$, axis=1) \\
degCentrality $\leftarrow$ (inDeg + outDeg) / (len(${\mathord{\buildrel{\lower3pt\hbox{$\scriptscriptstyle\frown$}} \over A}}$)-1) \\
$\tilde \beta$ $\leftarrow$ sum(outDeg $\times$ inDeg) / sum(outDeg) \\
$\tilde x$ $\leftarrow$ sum(outDeg $\times$ degCentrality) /sum(outDeg) \\
\Return $\tilde \beta$, $\tilde x$
\end{algorithm}  
	
\begin{algorithm} [hbtp]
\caption{Adjacency Matrix Modification}
\label{Algorithm-2}
\KwIn{Adjacency matrix ${\mathord{\buildrel{\lower3pt\hbox{$\scriptscriptstyle\frown$}} \over A}}$, rate of adversarial perturbation ${\mathord{\buildrel{\lower3pt\hbox{$\scriptscriptstyle\frown$}} \over {RoP}}}$} 
\KwOut{Resilience$\_$matrix ${\mathord{\buildrel{\lower3pt\hbox{$\scriptscriptstyle\frown$}} \over {RA}}}$} 
\textbf{Procedure defense ()} \\
simi$\_$list $\leftarrow$ calculate$\_$edge$\_$simi(${\mathord{\buildrel{\lower3pt\hbox{$\scriptscriptstyle\frown$}} \over A}}$) \\
${\mathord{\buildrel{\lower3pt\hbox{$\scriptscriptstyle\frown$}} \over {RA}}}$ $\leftarrow$ ${\mathord{\buildrel{\lower3pt\hbox{$\scriptscriptstyle\frown$}} \over A}}$ \\
critical$\_\tilde\beta$ $\leftarrow$ 0 \\
critical$\_\tilde x$ $\leftarrow$ 0   \\
Initial$\_$lenth $\leftarrow$ len(simi$\_$list) \\
\While {True} {
modified$\_$matrix, simi$\_$list $\leftarrow$ remove$\_$low$\_$simi$\_$edges(${\mathord{\buildrel{\lower3pt\hbox{$\scriptscriptstyle\frown$}} \over {RA}}}$.copy(), simi$\_$list) \\
$\tilde \beta$, $\tilde x$ $\leftarrow$ calculate$\_$parameters(modified$\_$matrix) \\
\If{$\tilde \beta$ $>$ critical$\_\tilde \beta$ and $\tilde x$ $>$ critical$\_\tilde x$}{
critical$\_\tilde \beta$ $\leftarrow$ $\tilde \beta$ \\
critical$\_\tilde x$ $\leftarrow$ $\tilde x$   \\
${\mathord{\buildrel{\lower3pt\hbox{$\scriptscriptstyle\frown$}} \over {RA}}}$ $\leftarrow$ modified$\_$matrix
}
\If{len(simi$\_$list) $\le$ Initial$\_$lenth $\times$ ${\mathord{\buildrel{\lower3pt\hbox{$\scriptscriptstyle\frown$}} \over {RoP}}}$}{
break
}
}	 
\Return{${\mathord{\buildrel{\lower3pt\hbox{$\scriptscriptstyle\frown$}} \over {RA}}}$}
\end{algorithm}     
	
The inferred asymptotically-stable equilibrium point (ASEP) servers as the referral to purifying the perturbed graph by enforcing adjacency matrix-projection coordinator to be close to such equilibrium point. The implementation mainly involves two phases: i) graph parameters calculation (Algorithm 1). This part aims to compute the two important parameters $\tilde \beta$ and $\tilde x$ as the referral to modify the adjacency matrix of perturbed graph; and ii) adjacency matrix modification (Algorithm 2). Referring to the two calculated critical-state parameters, this part mainly achieves the modification on adversarial edges. Specifically, we first define function \textit{calculate$\_$edge$\_$simi()} to empirically search the neighboring nodes within $h$-hop ($h$=2), then use Jaccard metric to calibrate the pairwise similarity for each two nodes. Based on the similarity, we define function \textit{remove$\_$low$\_$simi$\_$edges()} to remove those edges with low similarity during the "while" loop, until the inferred coordinator point from the iteratively-modified adjacency matrix approaches to the "Original" point (i.e. ASEP). 
	
\section{Experiment \& performance}     
\subsection{Configuration} \label{Sec:Configuration}
\textbf{Datasets.} Our experiments are performed on five commonly-used scalable realistic graph datasets: Polblogs, Cora, Cora\_ML, Citeseer, and Amazon Photo. The statistics are sketched in Table \ref{Table:dataset}. Given our method aims to promote the adversarial resilience from the perspective of graph topology, thus, the dataset Polblogs is completely adaptive since the node feature is unavailable. To sufficiently verify the effectiveness of our proposed method, we also introduce another four feature-involved datasets, however, our method and baselines only concentrate the graph topology information but disregard the node feature in the experiments. Concretely, we set the discrepancy of pairwise-node features as the same in our experimental procedure. 
	
\begin{table}[bth] 
\caption{Dataset statistics.}
\label{Table:dataset}
\centering
\newcolumntype{C}[1]{>{\centering\let\newline\\\arraybackslash\hspace{0pt}}m{#1}}
\begin{tabular}{C{0.15\textwidth} C{0.13\textwidth} C{0.13\textwidth} C{0.15\textwidth} C{0.18\textwidth}}
	\toprule
	\textbf{Dataset} & \textbf{\# Node} & \textbf{\# Edge} & \textbf{\# Feature} & \textbf{\# Classification}  \\
	\midrule
	PolBlogs & 1222 & 16714 & / & 2 \\
	Cora\_ML & 2810 & 7981 & 2879 & 7 \\
	Cora & 2485 & 5069 & 1433 & 7 \\
	Citeseer & 2110 & 3668 & 3703 & 6 \\
	Amazon Photo & 7650 & 238162 & 745 & 8 \\
	\bottomrule
\end{tabular}
\end{table}

\textbf{Baselines.} Methodologically, our work aims to improve the graph's adversarial resilience through modifying its adjacency matrix, same as ours, GCN-SVD \cite{Entezari20} and GCN \cite{KipfWelling17} also focus on the adjacency-matrix optimization as well. On the other hand, for other non-adjacency-matrix based defense approaches, such as GAT \cite{Velickovic18}, HANG \cite{ZhaoKang23}, and Mid-GCG \cite{HuangJin25}, our approach can also boost them from the lens of graph topology, that is, our work can significantly enhance the adversarial resilience by constructing a proper graph structure in advance.  
		
\begin{itemize}
\item \textbf{GCN \cite{KipfWelling17}:} Graph Convolutional Network (GCN), as a basic and prevalent deep learning model, is good at handling network/graph-specified data.   
\item \textbf{GAT \cite{Velickovic18}:} Graph Attention Network (GAT) equipped with attention layers aims to learn differential weights to different neighbors for individuals. It is usually utilized to defend against various graph adversarial attacks.
\item \textbf{GCN-SVD \cite{Entezari20}:} It is a preprocessing method to withstand adversarial attacks by decomposing adjacency matrix of the perturbed graph and then yielding a low-rank approximate matrix to purify the adversarial perturbations (edges).  
\item \textbf{HANG \cite{ZhaoKang23}:} It is a recently-proposed graph neural flow method, and resorting to the ordinary differential equation to learn a robust representation embeddings for graph nodes.
\item \textbf{Mid-GCN \cite{HuangJin25}:} It is a novel method recently proposed to leverage the mid-frequency signals (Laplacian eigenvalue around 1) on graphs for the promotion of robustness through designing a mid-pass filtering GCN model.
\end{itemize}

\textbf{Graph Adversarial Attacks.} As we know, the distinction between evasion attack and poisoning attack lies in different phases, i.e. the former works in inference process by modifying the testing samples, while the latter works in training phase by modifying the training samples. Thus, the three non-targeted GAAs used in this paper belong to poisoning attacks: 
\begin{itemize}
\item \textbf{Metattack \cite{Daniel19}:} It utilizes meta-learning to generate adversarial perturbations by computing gradients to determine how to modify the graph structure in order to achieve the attack objective. The gradient computation is typically based on the loss function of the target nodes, aiming to identify which edges should be modified to maximally affect the GNN's prediction results. Based on the computed gradient information, the attacker makes small changes to the graph, such as adding new edges or deleting existing ones. The modified graph structure is then used to retrain the GNN model and evaluate the success of the attack. 
\item \textbf{CE-PGD \cite{XuChen19}:} It uses the negative cross-entropy loss function to measure the difference between the model's predictions and the true labels. The attack starts from the original graph by randomly adding or deleting edges to initialize the adversarial graph. In each iteration, it computes the gradient of the loss function with respect to the current adversarial graph, then updates the graph using the gradient descent method. The updated graph structure is then projected back into the feasible region defined by the attack constraints. 
\item \textbf{DICE \cite{Waniek16}:} It is a heuristic adversarial attack inspired by modularity, which is used to measure the quality of any given community structure. In other words, it promotes structures where communities have dense connections within themselves and sparse connections between different communities. The algorithm mainly consists of two steps. The first step reduces the connection density among nodes within the same region. The second step increases the connections between nodes in that region and nodes in other regions, by which the algorithm can ignore that region, meaning it will not identify it as a separate community, but instead assign the nodes within that region to other communities.
\end{itemize}
	
\textbf{Execution Settings.} For each dataset, 10\% nodes are randomly chosen for training, 10\% nodes for verifying and the rest 80\% nodes for testing. For GCN, GAT, HANG, and Mid-GCN, the parameters are kept same as the default settings in their original papers. 
For GCN-SVD, the reduced rank is tuned from \{20, 40, 60, 80, 100\}. For our EquiliRes, the ratio of edge modification (deletion/addition) over dissimilar noes is tuned from \{1\%, 5\%, 10\%, 15\%\}, here 1\% just for the clean graph, i.e. adversarial perturbation rate is 0\%. The experimental results are calculated on average for 10 times. Additionally, the running environment is: ubuntu 20.04.2, GPU: GeForce RTX 3090. 
		
\subsection{Equilibrium-point trajectory} \label{Sec:EquiliPointTraj}
\begin{figure} [htbp]
	\centering
	\subfigure[Metattack]{
		\begin{minipage}[t]{0.3\columnwidth}
			\centering
			\includegraphics[height=1.25in, width=1.78in]{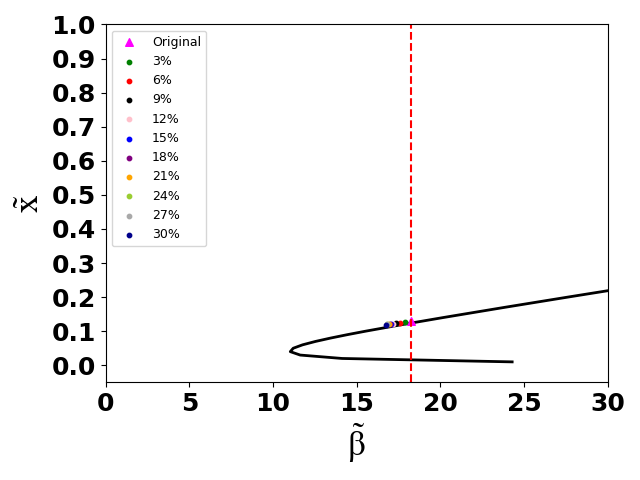}
			\label{Fig: Metattack}
		\end{minipage}
	}
	\vspace{0.02cm}
	\subfigure[CE-PGD]{
		\begin{minipage}[t]{0.3\columnwidth}
			\centering
			\includegraphics[height=1.25in, width=1.78in]{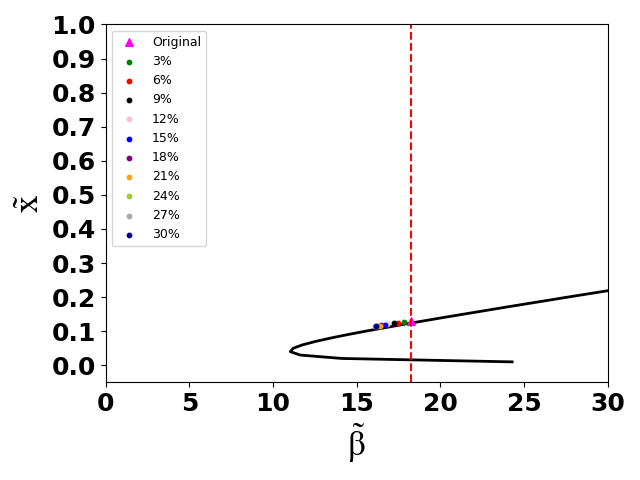}
			\label{Fig: PGD}
		\end{minipage}
	}
	\vspace{0.02cm}
	\subfigure[DICE]{
			\begin{minipage}[t]{0.3\columnwidth}
					\centering
					\includegraphics[height=1.25in, width=1.78in]{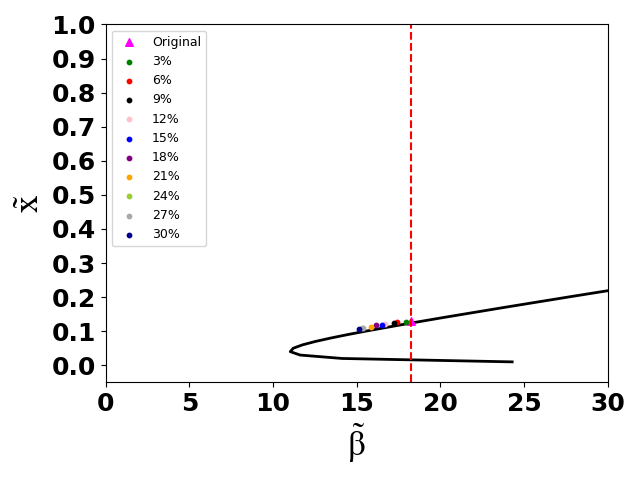}
					\label{Fig: DICE}
				\end{minipage}
		}
	\caption{Adversarial effect with Cora\_ML.}
	\label{Fig:Cora_MLTrajec}
\end{figure}

\begin{figure*} [htbp]
	\centering
	\subfigure[Metattack]{
		\begin{minipage}[t]{0.3\columnwidth}
			\centering
			\includegraphics[height=1.25in, width=1.78in]{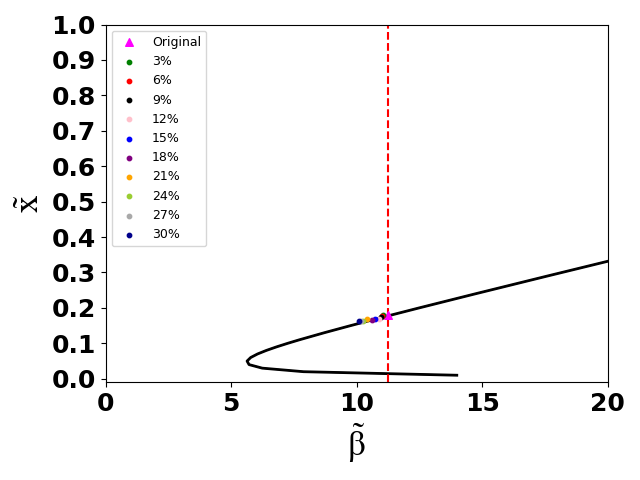}
			\label{Fig: Metattack}
		\end{minipage}
	}
	\vspace{0.02cm}
	\subfigure[CE-PGD]{
		\begin{minipage}[t]{0.3\columnwidth}
			\centering
			\includegraphics[height=1.25in, width=1.78in]{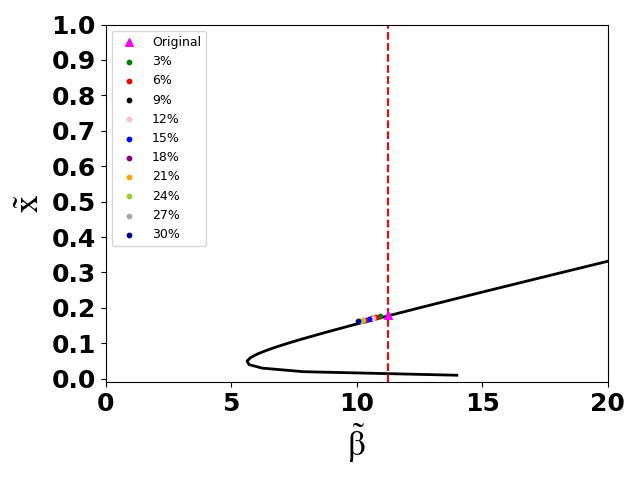}
			\label{Fig: PGD} 
		\end{minipage}
	}
	\vspace{0.02cm}
	\subfigure[DICE]{
			\begin{minipage}[t]{0.3\columnwidth}
					\centering
					\includegraphics[height=1.25in, width=1.78in]{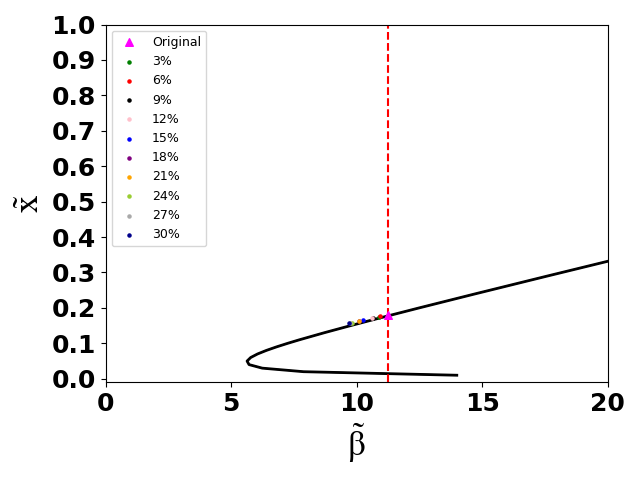}
					\label{Fig:DICE}
				\end{minipage}
		}
	\caption{Adversarial effect with Cora.}
	\label{Fig:CoraTrajec}
\end{figure*}

\begin{figure*} [htbp]
	\centering
	\subfigure[Metattack]{
		\begin{minipage}[t]{0.3\columnwidth}
			\centering
			\includegraphics[height=1.25in, width=1.78in]{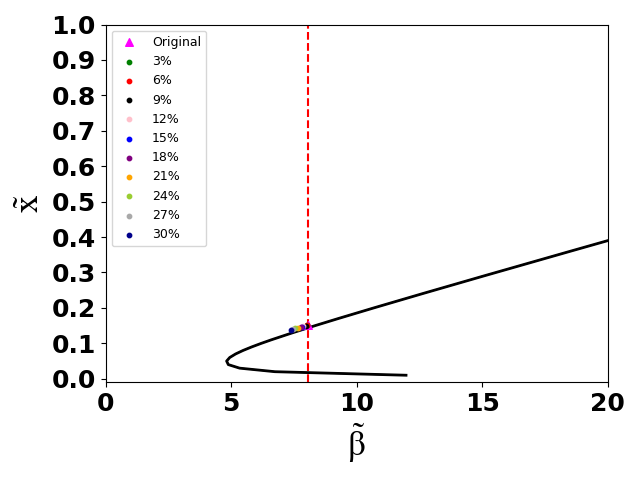}
			\label{Fig: Metattack}
		\end{minipage}
	}
	\vspace{0.02cm}
	\subfigure[CE-PGD]{
		\begin{minipage}[t]{0.3\columnwidth}
			\centering
			\includegraphics[height=1.25in, width=1.78in]{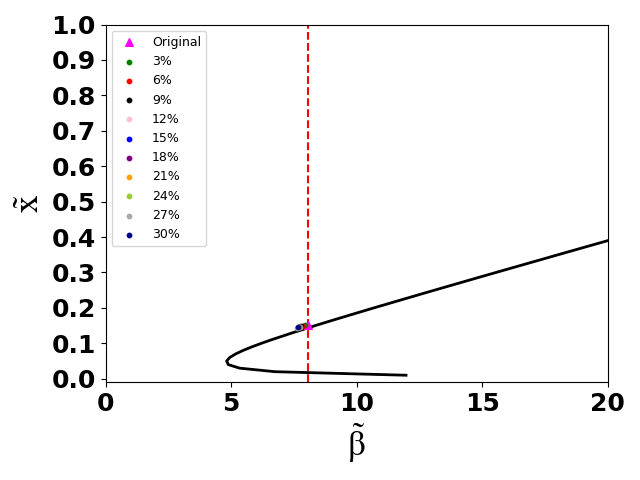}
			\label{Fig: PGD}
		\end{minipage}
	}
	\vspace{0.02cm}
	\subfigure[DICE]{
			\begin{minipage}[t]{0.3\columnwidth}
					\centering
					\includegraphics[height=1.25in, width=1.78in]{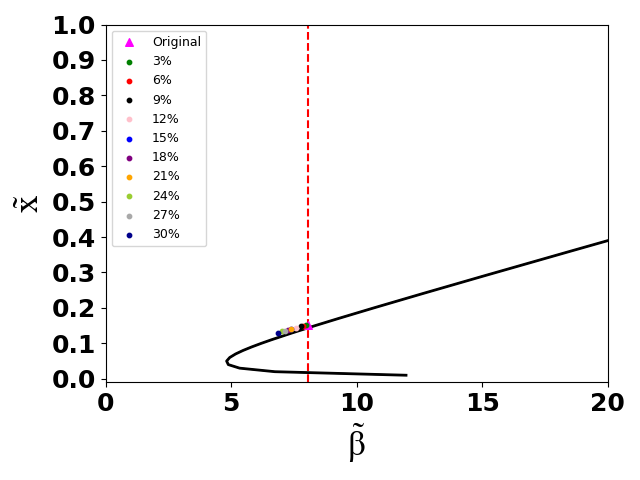}
					\label{Fig: DICE}
				\end{minipage}
		}
	\caption{Adversarial effect with Citeseer.}
	\label{Fig:CiteseerTrajec}
\end{figure*}

\begin{figure*} [htbp]
	\centering
	\subfigure[Metattack]{
		\begin{minipage}[t]{0.3\columnwidth}
			\centering
			\includegraphics[height=1.25in, width=1.78in]{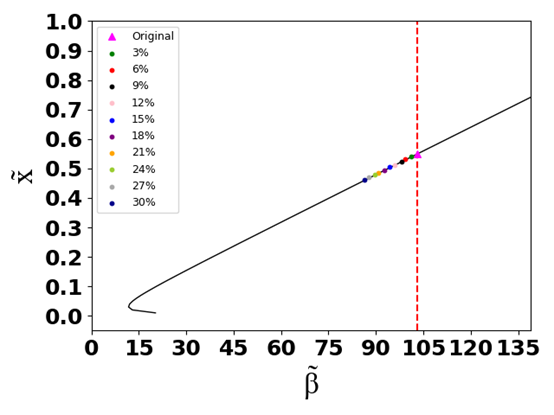}
			\label{Fig: Metattack}
		\end{minipage}
	}
	\vspace{0.02cm}
	\subfigure[CE-PGD]{
		\begin{minipage}[t]{0.3\columnwidth}
			\centering
			\includegraphics[height=1.2in, width=1.78in]{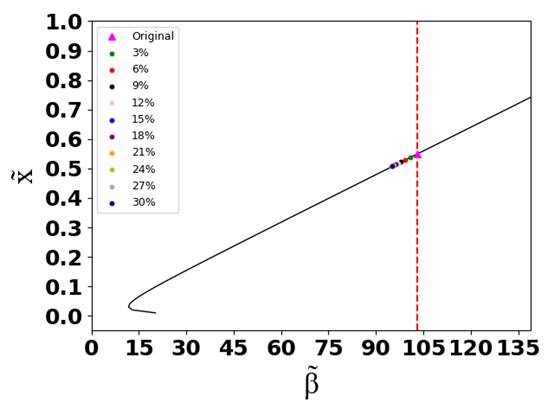}
			\label{Fig: PGD}
		\end{minipage}
	}
	\vspace{0.02cm}
	\subfigure[DICE]{
		\begin{minipage}[t]{0.3\columnwidth}
			\centering
			\includegraphics[height=1.25in, width=1.78in]{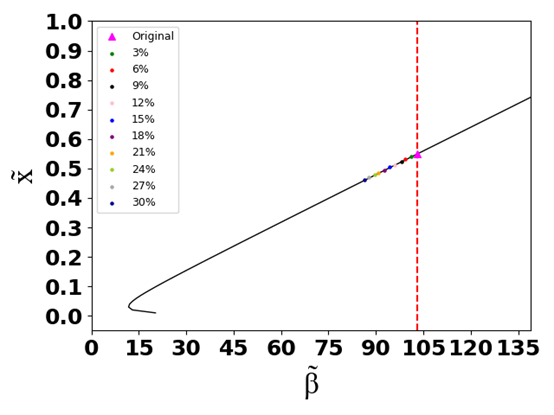}
			\label{Fig: DICE}
		\end{minipage}
	}
	\caption{Adversarial effect with Amazon Photo.}
	\label{Fig:AmazonPhotoTrajec}
\end{figure*}

Pursuant to the parameters as listed in Table \ref{Table:ResiTraPara}, we draw the trajectories of asymptotically-stable equilibrium-points on datasets Polblogs, Cora\_ML, Cora, Citeseer and Amazon Photo as displayed in Fig. \ref{Fig: PolblogsTrajec}-\ref{Fig:AmazonPhotoTrajec}. From the trajectories, besides the analogous analytics and reviews as presented in Section \ref{Sec:trajectoryComputation}, we also obtain the following observations: i) although the plot of adversarial resilience can well-match the attack effect under different rates of adversarial perturbation, the graph's adversarial resilience indeed declines as the strength of perturbation enlarges, which makes sense in intuition, you can image an extreme case that if all edges are removed, the nodes would become isolated and cannot be classified without the connections to each other; and ii) different datasets have distinct trajectories of adversarial resilience, stemming from the difference of graph's internal topology. Thereby, we need carefully figure out such an equilibrium-point trajectory for each unique graph. 
		
\begin{table}[hbt] 
\caption{Parameters for generating equilibrium-point trajectory.}
\label{Table:ResiTraPara}
\centering
\newcolumntype{C}[1]{>{\centering\let\newline\\\arraybackslash\hspace{0pt}}m{#1}}
\begin{tabular}{C{0.22\textwidth} C{0.35\textwidth} C{0.25\textwidth} }		
\toprule
\textbf{Dataset} & \textbf{GAA} &\textbf{Parameter} \\
\midrule
Polblogs & Metattack/CE-PGD/DICE &  $\theta=16.6^2$   $\eta$=0.45 \\
\midrule
Cora\_ML & Metattack/CE-PGD/DICE   &  $\theta=24^2$    $\eta$=10 \\
\midrule
{Cora} & Metattack/CE-PGD/DICE   &  $\theta=21^2$    $\eta$=10 \\
\midrule
Citeseer & Metattack/CE-PGD/DICE   &  $\theta=21^2$    $\eta$=10 \\
\midrule
Amazon Photo & Metattack/CE-PGD/DICE   &  $\theta=31.8^2$    $\eta$=8 \\
\bottomrule
\end{tabular}
\end{table}
			
\subsection{Performance on attack resilience} \label{Sec: ExtraPerformanceResults}
Table \ref{Table: CiteseerUp} and Table \ref{Table: AmazonPhoto} show the comparative results separately between our approach and another two adjacency-matrix optimization  defense methods GCN and GCN-SVD. Table \ref{Table: EquiliResCombine-2ndPart} exhibits the results of GAT, HANG and Mid-GCN from angle of combination with our EquiliRes, from which we can gain the analogous comparative analysis and conclusion as stated in Section \ref{Sec:AdverResiPro}. 
\begin{table*}[htbp]  
\caption{Accuracy (\%) of node classification on Citeseer.}
\label{Table: CiteseerUp}
\begin{small}
\vskip 0.1in
\centering
\begin{tabular}{|p{1.1cm}|p{1.45cm}|>{\centering\arraybackslash}p{1.32cm}|>{\centering\arraybackslash}p{1.32cm}|>{\centering\arraybackslash}p{1.32cm}|>{\centering\arraybackslash}p{1.32cm}|>{\centering\arraybackslash}p{1.32cm}|>{\centering\arraybackslash}p{1.32cm}|}		
	\hline
	\textbf{GAA} & \diagbox[dir=NW]{\textbf{Def.}}{\textbf{Per.}} & \textbf{0\%} & \textbf{5\%} & \textbf{10\%} & \textbf{15\%} & \textbf{20\%} & \textbf{25\%}\\
	\hline
	\multirow{3}{*}{Metattack} & GCN & 54.85$\pm$1.71 & 52.30$\pm$2.13 & 46.92$\pm$1.58 & 43.30$\pm$1.42 & 41.18$\pm$1.27 & 39.40$\pm$0.73 \\
	\cline{2-8}
	& GCN-SVD & 46.12$\pm$1.54  & 45.30$\pm$1.64 &  44.89$\pm$1.44 & 44.28$\pm$1.84 & 43.26$\pm$1.45 & 42.84$\pm$1.48 \\
	\cline{2-8}
	& EquiliRes & 
	\begin{tabular} [c]{@{}l@{}} \textbf{55.76$\pm$2.07} \\ (\textbf{\textcolor{red}{0.91\%$\uparrow$}}) \end{tabular} &  
	\begin{tabular} [c]{@{}l@{}} \textbf{55.05$\pm$1.21} \\ (\textbf{\textcolor{red}{2.75\%$\uparrow$}})\end{tabular} & 
	\begin{tabular} [c]{@{}l@{}} \textbf{52.59$\pm$1.04} \\ (\textbf{\textcolor{red}{5.67\%$\uparrow$}}) \end{tabular} &  \begin{tabular} [c]{@{}l@{}} \textbf{50.79$\pm$1.78} \\ (\textbf{\textcolor{red}{6.51\%$\uparrow$}}) \end{tabular} & 
	\begin{tabular} [c]{@{}l@{}} \textbf{48.19$\pm$0.34} \\ (\textbf{\textcolor{red}{4.93\%$\uparrow$}}) \end{tabular} &  \begin{tabular} [c]{@{}l@{}} \textbf{45.15$\pm$2.19} \\ (\textbf{\textcolor{red}{2.31\%$\uparrow$}}) \end{tabular} \\		         	 	           
	\hline\hline
	\multirow{3}{*}{CE-PGD} & GCN & 54.85$\pm$1.71 & 52.76$\pm$1.95 & 50.46$\pm$2.44 & 50.34$\pm$1.79 &	49.25$\pm$1.51 & 49.68$\pm$1.46 \\
	\cline{2-8}
	& GCN-SVD & 46.12$\pm$1.54 & 46.22$\pm$1.78 & 46.16$\pm$1.69 & 45.73$\pm$1.93 & 45.80$\pm$3.20 & 47.70$\pm$3.15 \\
	\cline{2-8}
	& EquiliRes & 
	\begin{tabular} [c]{@{}l@{}} \textbf{55.76$\pm$2.07} \\ (\textbf{\textcolor{red}{0.91\%$\uparrow$}}) \end{tabular} &  
	\begin{tabular} [c]{@{}l@{}} \textbf{54.67$\pm$2.44} \\ (\textbf{\textcolor{red}{1.91\%$\uparrow$}}) \end{tabular} &  
	\begin{tabular} [c]{@{}l@{}} \textbf{53.11$\pm$2.41} \\ (\textbf{\textcolor{red}{2.65\%$\uparrow$}}) \end{tabular} & 
	\begin{tabular} [c]{@{}l@{}} \textbf{52.87$\pm$2.35} \\ (\textbf{\textcolor{red}{2.53\%$\uparrow$}}) \end{tabular} &  
	\begin{tabular} [c]{@{}l@{}} \textbf{51.55$\pm$2.32} \\ (\textbf{\textcolor{red}{2.30\%$\uparrow$}}) \end{tabular} &
	\begin{tabular} [c]{@{}l@{}} \textbf{52.89$\pm$2.18} \\ (\textbf{\textcolor{red}{3.21\%$\uparrow$}}) \end{tabular} 
	\\			
	\hline\hline			
	\multirow{3}{*}{DICE} & GCN & 54.85$\pm$1.71 & 51.48$\pm$1.48 & 51.16$\pm$1.75 &	49.00$\pm$1.43 & 47.71$\pm$1.48 & 45.94$\pm$1.53 \\
	\cline{2-8}
	& GCN-SVD & 46.12$\pm$1.54 & 44.70$\pm$1.96 & 41.65$\pm$3.07 & 41.58$\pm$2.43 & 39.61$\pm$2.84 &
	37.71$\pm$1.75 \\
	\cline{2-8}
	& EquiliRes & 
	\begin{tabular} [c]{@{}l@{}} \textbf{55.76$\pm$2.07} \\ (\textbf{\textcolor{red}{0.91\%$\uparrow$}}) \end{tabular} &  
	\begin{tabular} [c]{@{}l@{}} \textbf{53.30$\pm$2.09} \\ (\textbf{\textcolor{red}{1.82\%$\uparrow$}}) \end{tabular} &  
	\begin{tabular} [c]{@{}l@{}} \textbf{51.42$\pm$1.73} \\ (\textbf{\textcolor{red}{0.26\%$\uparrow$}}) \end{tabular} & 
	\begin{tabular} [c]{@{}l@{}} \textbf{50.81$\pm$2.16}\\ (\textbf{\textcolor{red}{1.81\%$\uparrow$}}) \end{tabular} &  
	\begin{tabular} [c]{@{}l@{}} \textbf{49.46$\pm$1.78}\\ (\textbf{\textcolor{red}{1.75\%$\uparrow$}}) \end{tabular}&
	\begin{tabular} [c]{@{}l@{}} \textbf{47.01$\pm$1.68} \\ (\textbf{\textcolor{red}{1.07\%$\uparrow$}}) \end{tabular} \\	
	\hline				
\end{tabular}
\end{small}
\vskip -0.1in
\end{table*}

\begin{table*}[htbp]  
\caption{Accuracy (\%) of node classification on Amazon Photo.}
\label{Table: AmazonPhoto}
\begin{small}
\vskip 0.1in
\centering
		\begin{tabular}{|p{0.6cm}|p{1.43cm}|>{\centering\arraybackslash}p{1.32cm}|>{\centering\arraybackslash}p{1.32cm}|>{\centering\arraybackslash}p{1.45cm}|>{\centering\arraybackslash}p{1.45cm}|>{\centering\arraybackslash}p{1.45cm}|>{\centering\arraybackslash}p{1.45cm}|}
			\hline
			\textbf{GAA} & \diagbox[dir=NW]{\textbf{Def.}}{\textbf{Per.}} & \textbf{0\%} & \textbf{5\%} & \textbf{10\%} & \textbf{15\%} & \textbf{20\%} & \textbf{25\%}\\
			\hline
			\multirow{3}{*}{\makecell{Meta\\ttack}} & GCN & 92.15$\pm$0.34 & 82.42$\pm$7.74 & 82.81$\pm$5.81 & 79.85$\pm$13.16 & 62.95$\pm$14.96 & 61.08$\pm$15.26 \\
			\cline{2-8}
			& GCN-SVD & 87.13$\pm$0.47  & 84.32$\pm$7.94 &  82.01$\pm$9.81 & 72.39$\pm$12.46 & 71.21$\pm$12.29 & 63.57$\pm$16.78 \\
			\cline{2-8}
			& EquiliRes & 
			\begin{tabular} [c]{@{}l@{}} \textbf{92.87$\pm$0.46} \\ (\textbf{\textcolor{red}{0.72\%$\uparrow$}}) \end{tabular} &  
			\begin{tabular} [c]{@{}l@{}} \textbf{91.65$\pm$0.21} \\ (\textbf{\textcolor{red}{7.33\%$\uparrow$}})\end{tabular} & 
			\begin{tabular} [c]{@{}l@{}} \textbf{85.11$\pm$20.05} \\ (\textbf{\textcolor{red}{2.30\%$\uparrow$}}) \end{tabular} &  \begin{tabular} [c]{@{}l@{}} \textbf{82.25$\pm$30.00} \\ (\textbf{\textcolor{red}{2.40\%$\uparrow$}}) \end{tabular} & 
			\begin{tabular} [c]{@{}l@{}} \textbf{74.35$\pm$31.92} \\ (\textbf{\textcolor{red}{3.14\%$\uparrow$}}) \end{tabular} &  \begin{tabular} [c]{@{}l@{}} \textbf{81.14$\pm$25.58} \\ (\textbf{\textcolor{red}{17.57\%$\uparrow$}}) \end{tabular} \\		         	 	           
			\hline\hline
			\multirow{3}{*}{\makecell{CE-\\PGD}} & GCN & 92.15$\pm$0.34 & 90.42$\pm$0.37 & 90.15$\pm$0.31 & 90.12$\pm$0.29 &	88.41$\pm$0.56 & 88.29$\pm$0.34 \\
			\cline{2-8}
			& GCN-SVD & 87.13$\pm$0.47 & 85.82$\pm$4.83 & 87.75$\pm$0.36 & 86.78$\pm$2.47 & 86.75$\pm$2.17 & 86.17$\pm$0.49 \\
			\cline{2-8}
			& EquiliRes & 
			\begin{tabular} [c]{@{}l@{}} \textbf{92.87$\pm$0.46} \\ (\textbf{\textcolor{red}{0.72\%$\uparrow$}}) \end{tabular} &  
			\begin{tabular} [c]{@{}l@{}} \textbf{91.12$\pm$0.32} \\ (\textbf{\textcolor{red}{0.70\%$\uparrow$}}) \end{tabular} &  
			\begin{tabular} [c]{@{}l@{}} \textbf{90.62$\pm$0.45} \\ (\textbf{\textcolor{red}{0.47\%$\uparrow$}}) \end{tabular} & 
			\begin{tabular} [c]{@{}l@{}} \textbf{90.33$\pm$0.17} \\ (\textbf{\textcolor{red}{0.21\%$\uparrow$}}) \end{tabular} &  
			\begin{tabular} [c]{@{}l@{}} \textbf{89.39$\pm$0.46} \\ (\textbf{\textcolor{red}{0.98\%$\uparrow$}}) \end{tabular} &
			\begin{tabular} [c]{@{}l@{}} \textbf{88.58$\pm$0.25} \\ (\textbf{\textcolor{red}{0.29\%$\uparrow$}}) \end{tabular} 
			\\			
			\hline\hline			
			\multirow{3}{*}{DICE} & GCN & 92.15$\pm$0.34 & 90.96$\pm$0.65 & 89.03$\pm$5.30 & 89.00$\pm$1.01 & 86.60$\pm$3.17 & 84.90$\pm$3.18 \\
			\cline{2-8}
			& GCN-SVD & 87.13$\pm$0.47 & 85.03$\pm$0.56 & 82.74$\pm$0.89 & 81.54$\pm$0.41 & 75.75$\pm$2.72 &
			76.14$\pm$0.57 \\
			\cline{2-8}
			& EquiliRes & 
			\begin{tabular} [c]{@{}l@{}} \textbf{92.87$\pm$0.46} \\ (\textbf{\textcolor{red}{0.72\%$\uparrow$}}) \end{tabular} &  
			\begin{tabular} [c]{@{}l@{}} \textbf{91.09$\pm$0.33} \\ (\textbf{\textcolor{red}{0.13\%$\uparrow$}}) \end{tabular} &  
			\begin{tabular} [c]{@{}l@{}} \textbf{91.45$\pm$0.21} \\ (\textbf{\textcolor{red}{2.42\%$\uparrow$}}) \end{tabular} & 
			\begin{tabular} [c]{@{}l@{}} \textbf{90.84$\pm$0.12}\\ (\textbf{\textcolor{red}{1.84\%$\uparrow$}}) \end{tabular} &  
			\begin{tabular} [c]{@{}l@{}} \textbf{89.45$\pm$0.53}\\ (\textbf{\textcolor{red}{2.85\%$\uparrow$}}) \end{tabular}&
			\begin{tabular} [c]{@{}l@{}} \textbf{88.59$\pm$0.11} \\ (\textbf{\textcolor{red}{3.69\%$\uparrow$}}) \end{tabular} \\	
			\hline				
		\end{tabular}
	\end{small}
	\vskip -0.1in
\end{table*}

\begin{table*}[hbtp]  
\caption{Accuracy (\%) of node classification under combination with EquiliRes using Metattack.}
\label{Table: EquiliResCombine-2ndPart}
\begin{small}
\vskip 0.1in
\centering
	\begin{tabular}{|p{1.0cm}|p{1.5cm}|>{\centering\arraybackslash}p{1.32cm}|>{\centering\arraybackslash}p{1.32cm}|>{\centering\arraybackslash}p{1.32cm}|>{\centering\arraybackslash}p{1.32cm}|>{\centering\arraybackslash}p{1.32cm}|>{\centering\arraybackslash}p{1.32cm}|}
		\hline
		\textbf{Dataset} & \diagbox[dir=NW]{\textbf{Def.}}{\textbf{Per.}} & \textbf{0\%} & \textbf{5\%} & \textbf{10\%} & \textbf{15\%} & \textbf{20\%} & \textbf{25\%}\\
		\hline
		\hline\hline
		\multirow{6}{*}{Cora\_ML} 
		& GAT & 77.55$\pm$1.15 & 68.86$\pm$1.04 & 64.13$\pm$1.37 & 60.65$\pm$0.92 & 58.75$\pm$1.70 & 56.43$\pm$1.26 \\
		\cline{2-8}
		& Ours\textbf{+}GAT & 
		\begin{tabular} [c]{@{}l@{}} \textbf{77.69$\pm$1.30} \\ (\textbf{\textcolor{red}{0.14\%$\uparrow$}}) \end{tabular} &  
		\begin{tabular} [c]{@{}l@{}} \textbf{71.15$\pm$1.85} \\ (\textbf{\textcolor{red}{2.29\%$\uparrow$}})\end{tabular} & 
		\begin{tabular} [c]{@{}l@{}} \textbf{68.32$\pm$1.99} \\ (\textbf{\textcolor{red}{4.19\%$\uparrow$}}) \end{tabular}&  \begin{tabular} [c]{@{}l@{}} \textbf{66.11$\pm$1.61} \\ (\textbf{\textcolor{red}{5.46\%$\uparrow$}}) \end{tabular} & 
		\begin{tabular} [c]{@{}l@{}} \textbf{64.52$\pm$1.71} \\ (\textbf{\textcolor{red}{5.77\%$\uparrow$}}) \end{tabular}&  \begin{tabular} [c]{@{}l@{}} \textbf{62.12$\pm$1.59} \\ (\textbf{\textcolor{red}{5.69\%$\uparrow$}}) \end{tabular} \\	
		\cline{2-8}
		& HANG & 64.11$\pm$4.49 & 57.98$\pm$2.30 & 52.98$\pm$2.23 & 50.54$\pm$2.52 & 47.64$\pm$2.51 & 45.93$\pm$2.93 \\
		\cline{2-8}
		& Ours\textbf{+}HANG & 
		\begin{tabular} [c]{@{}l@{}} \textbf{64.34$\pm$5.03} \\ (\textbf{\textcolor{red}{0.23\%$\uparrow$}}) \end{tabular} &  
		\begin{tabular} [c]{@{}l@{}} \textbf{60.47$\pm$3.58} \\ (\textbf{\textcolor{red}{2.49\%$\uparrow$}}) \end{tabular} &  
		\begin{tabular} [c]{@{}l@{}} \textbf{57.08$\pm$2.68} \\ (\textbf{\textcolor{red}{4.10\%$\uparrow$}}) \end{tabular} & 
		\begin{tabular} [c]{@{}l@{}} \textbf{55.34$\pm$3.14} \\ (\textbf{\textcolor{red}{4.80\%$\uparrow$}}) \end{tabular} &  
		\begin{tabular} [c]{@{}l@{}} \textbf{54.23$\pm$3.63} \\ (\textbf{\textcolor{red}{6.59\%$\uparrow$}}) \end{tabular} &
		\begin{tabular} [c]{@{}l@{}} \textbf{53.06$\pm$3.25} \\ (\textbf{\textcolor{red}{7.13\%$\uparrow$}}) \end{tabular} 
		\\	
		\cline{2-8}
		& Mid-GCN & 83.40$\pm$0.21 & 80.01$\pm$0.37 & 78.95$\pm$0.37 & 77.80$\pm$0.55 & 75.66$\pm$0.57 & 74.17$\pm$0.72 \\
		\cline{2-8}
		& Ours\textbf{+}Mid-GCN & 
		\begin{tabular} [c]{@{}l@{}} \textbf{83.49$\pm$0.16} \\ (\textbf{\textcolor{red}{0.09\%$\uparrow$}}) \end{tabular} &  
		\begin{tabular} [c]{@{}l@{}} \textbf{80.22$\pm$0.46} \\ (\textbf{\textcolor{red}{0.21\%$\uparrow$}}) \end{tabular} &  
		\begin{tabular} [c]{@{}l@{}} \textbf{79.70$\pm$0.34} \\ (\textbf{\textcolor{red}{0.75\%$\uparrow$}}) \end{tabular} & 
		\begin{tabular} [c]{@{}l@{}} \textbf{78.03$\pm$0.63} \\ (\textbf{\textcolor{red}{0.23\%$\uparrow$}}) \end{tabular} &  
		\begin{tabular} [c]{@{}l@{}} \textbf{76.19$\pm$0.72} \\ (\textbf{\textcolor{red}{0.53\%$\uparrow$}}) \end{tabular} &
		\begin{tabular} [c]{@{}l@{}} \textbf{74.53$\pm$0.83} \\ (\textbf{\textcolor{red}{0.36\%$\uparrow$}}) \end{tabular} 
		\\			
		\hline\hline			
		\multirow{6}{*}{Cora} 
		& GAT & 73.28$\pm$1.43 & 67.53$\pm$1.39 &	61.20$\pm$1.77 & 58.09$\pm$1.21 & 55.07$\pm$1.59 & 52.16$\pm$1.75 \\
		\cline{2-8}
		& Ours\textbf{+}GAT & 
		\begin{tabular} [c]{@{}l@{}} \textbf{73.43$\pm$1.36} \\ (\textbf{\textcolor{red}{0.15\%$\uparrow$}}) \end{tabular} &  
		\begin{tabular} [c]{@{}l@{}} \textbf{67.65 $\pm$1.60} \\ (\textbf{\textcolor{red}{0.12\%$\uparrow$}})\end{tabular} & 
		\begin{tabular} [c]{@{}l@{}} \textbf{63.53$\pm$2.06} \\ (\textbf{\textcolor{red}{2.33\%$\uparrow$}}) \end{tabular}&  \begin{tabular} [c]{@{}l@{}} \textbf{60.08$\pm$2.29} \\ (\textbf{\textcolor{red}{1.99\%$\uparrow$}}) \end{tabular} & 
		\begin{tabular} [c]{@{}l@{}} \textbf{57.66$\pm$2.22} \\ (\textbf{\textcolor{red}{2.59\%$\uparrow$}}) \end{tabular}&  \begin{tabular} [c]{@{}l@{}} \textbf{55.23$\pm$2.05} \\ (\textbf{\textcolor{red}{3.07\%$\uparrow$}}) \end{tabular} \\	
		\cline{2-8}
		& HANG & 58.22$\pm$4.79 & 55.23$\pm$3.80 & 50.25$\pm$3.30 & 48.36$\pm$3.33 & 45.84$\pm$3.12 & 44.54$\pm$3.14	\\
		\cline{2-8}
		& Ours\textbf{+}HANG & 
		\begin{tabular} [c]{@{}l@{}} \textbf{58.40$\pm$4.82} \\ (\textbf{\textcolor{red}{0.18\%$\uparrow$}}) \end{tabular} &  
		\begin{tabular} [c]{@{}l@{}} \textbf{55.49$\pm$3.98} \\ (\textbf{\textcolor{red}{0.26\%$\uparrow$}})\end{tabular}&  
		\begin{tabular} [c]{@{}l@{}} \textbf{52.41$\pm$3.50} \\ (\textbf{\textcolor{red}{2.16\%$\uparrow$}}) \end{tabular} & 
		\begin{tabular} [c]{@{}l@{}} \textbf{50.22$\pm$3.58}\\ (\textbf{\textcolor{red}{1.86\%$\uparrow$}}) \end{tabular} &  
		\begin{tabular} [c]{@{}l@{}} \textbf{48.77$\pm$2.99}\\(\textbf{\textcolor{red}{2.93\%$\uparrow$}})\end{tabular}&		
		\begin{tabular} [c]{@{}l@{}} \textbf{47.29$\pm$2.85} \\ (\textbf{\textcolor{red}{2.75\%$\uparrow$}}) \end{tabular} 
		\\	
		\cline{2-8}
		& Mid-GCN & 83.10$\pm$0.41 & 75.81$\pm$15.82 & 78.88$\pm$0.93 & 75.43$\pm$1.11 & 75.50$\pm$1.35 & 74.43$\pm$0.83	\\
		\cline{2-8}
		& Ours\textbf{+}Mid-GCN & 
		\begin{tabular} [c]{@{}l@{}} \textbf{83.23$\pm$0.36} \\ (\textbf{\textcolor{red}{0.13\%$\uparrow$}}) \end{tabular} &  
		\begin{tabular} [c]{@{}l@{}} \textbf{81.27$\pm$0.72} \\ (\textbf{\textcolor{red}{5.46\%$\uparrow$}})\end{tabular}&  
		\begin{tabular} [c]{@{}l@{}} \textbf{80.20$\pm$1.10} \\ (\textbf{\textcolor{red}{1.32\%$\uparrow$}}) \end{tabular} & 
		\begin{tabular} [c]{@{}l@{}} \textbf{77.69$\pm$0.82}\\ (\textbf{\textcolor{red}{2.26\%$\uparrow$}}) \end{tabular} &  
		\begin{tabular} [c]{@{}l@{}} \textbf{76.74$\pm$0.64}\\(\textbf{\textcolor{red}{1.24\%$\uparrow$}})\end{tabular}&		
		\begin{tabular} [c]{@{}l@{}} \textbf{75.16$\pm$0.76} \\ (\textbf{\textcolor{red}{0.73\%$\uparrow$}}) \end{tabular} 
		\\	
		\hline\hline			
		\multirow{6}{*}{Citeseer} 
		& GAT & 64.04$\pm$1.85 & 60.21$\pm$1.91 & 54.00$\pm$1.81 & 49.92$\pm$2.02 & 46.96$\pm$1.70 & 44.65$\pm$0.97 \\
		\cline{2-8}
		& Ours\textbf{+}GAT & 
		\begin{tabular} [c]{@{}l@{}} \textbf{64.17$\pm$1.73} \\ (\textbf{\textcolor{red}{0.13\%$\uparrow$}}) \end{tabular} &  
		\begin{tabular} [c]{@{}l@{}} \textbf{61.14$\pm$1.97} \\ (\textbf{\textcolor{red}{0.93\%$\uparrow$}})\end{tabular} & 
		\begin{tabular} [c]{@{}l@{}} \textbf{54.60$\pm$1.65} \\ (\textbf{\textcolor{red}{0.60\%$\uparrow$}}) \end{tabular}&  \begin{tabular} [c]{@{}l@{}} \textbf{53.21$\pm$1.36} \\ (\textbf{\textcolor{red}{3.29\%$\uparrow$}}) \end{tabular} & 
		\begin{tabular} [c]{@{}l@{}} \textbf{51.30$\pm$1.64} \\ (\textbf{\textcolor{red}{4.34\%$\uparrow$}}) \end{tabular}&  \begin{tabular} [c]{@{}l@{}} \textbf{48.39$\pm$2.31} \\ (\textbf{\textcolor{red}{3.74\%$\uparrow$}}) \end{tabular} \\	
		\cline{2-8}
		& HANG & 56.48$\pm$2.85 & 52.92$\pm$1.91 & 48.05$\pm$2.72 & 45.12$\pm$0.83 & 44.29$\pm$1.83 & 40.61$\pm$1.96	\\
		\cline{2-8}
		& Ours\textbf{+}HANG & 
		\begin{tabular} [c]{@{}l@{}} \textbf{56.54$\pm$2.74} \\ (\textbf{\textcolor{red}{0.06\%$\uparrow$}}) \end{tabular} &  
		\begin{tabular} [c]{@{}l@{}} \textbf{53.17$\pm$2.37} \\ (\textbf{\textcolor{red}{0.25\%$\uparrow$}})\end{tabular}&  
		\begin{tabular} [c]{@{}l@{}} \textbf{49.62$\pm$2.17} \\ (\textbf{\textcolor{red}{1.57\%$\uparrow$}}) \end{tabular} & 
		\begin{tabular} [c]{@{}l@{}} \textbf{49.62$\pm$2.17}\\ (\textbf{\textcolor{red}{4.50\%$\uparrow$}}) \end{tabular} &  
		\begin{tabular} [c]{@{}l@{}} \textbf{46.64$\pm$1.60}\\(\textbf{\textcolor{red}{2.35\%$\uparrow$}})\end{tabular}&		
		\begin{tabular} [c]{@{}l@{}} \textbf{44.70$\pm$2.40} \\ (\textbf{\textcolor{red}{4.09\%$\uparrow$}}) \end{tabular} 
		\\
		\cline{2-8}
		& Mid-GCN & 74.52$\pm$0.32 & 72.37$\pm$0.67 & 71.05$\pm$0.66 & 65.84$\pm$1.06 & 67.00$\pm$0.52 & 63.06$\pm$0.99	\\
		\cline{2-8}
		& Ours\textbf{+}Mid-GCN & 
		\begin{tabular} [c]{@{}l@{}} \textbf{74.66$\pm$0.47} \\ (\textbf{\textcolor{red}{0.14\%$\uparrow$}}) \end{tabular} &  
		\begin{tabular} [c]{@{}l@{}} \textbf{72.70$\pm$0.37} \\ (\textbf{\textcolor{red}{0.33\%$\uparrow$}})\end{tabular}&  
		\begin{tabular} [c]{@{}l@{}} \textbf{71.52$\pm$0.71} \\ (\textbf{\textcolor{red}{0.47\%$\uparrow$}}) \end{tabular} & 
		\begin{tabular} [c]{@{}l@{}} \textbf{66.45$\pm$1.03}\\ (\textbf{\textcolor{red}{0.61\%$\uparrow$}}) \end{tabular} &  
		\begin{tabular} [c]{@{}l@{}} \textbf{67.84$\pm$0.50}\\(\textbf{\textcolor{red}{0.84\%$\uparrow$}})\end{tabular}&		
		\begin{tabular} [c]{@{}l@{}} \textbf{63.84$\pm$1.25} \\ (\textbf{\textcolor{red}{0.78\%$\uparrow$}}) \end{tabular} 
		\\		
		\hline				
	\end{tabular}
\end{small}
\vskip -0.1in
\end{table*}
			
\subsection{Rank variation of adjacency matrix} \label{Sec:Rank}
\begin{figure} [tbhp]
	\centering
	\subfigure[Rank growth under attacks]{
		\begin{minipage}[t]{0.45\columnwidth}
			\centering
			\includegraphics[width=1.9in, height=1.45in]{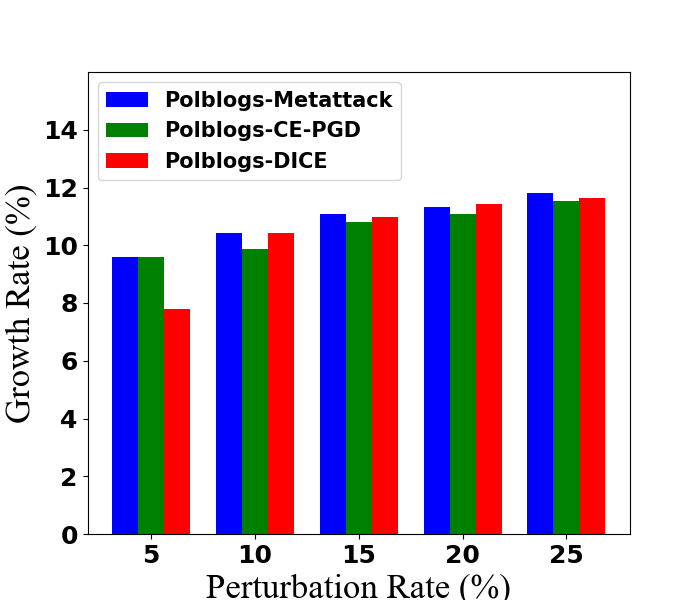}
			\label{Fig: Metattack}
		\end{minipage}
	}
	\vspace{0.02cm}
	\subfigure[Rank decrease by EquiliRes]{
		\begin{minipage}[t]{0.45\columnwidth}
			\centering
			\includegraphics[width=1.9in, height=1.45in]{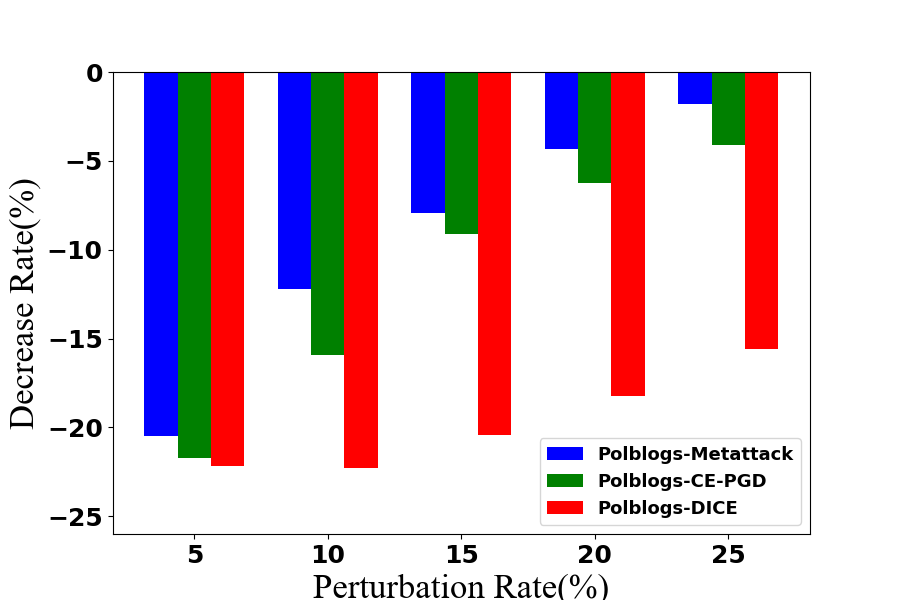}
			\label{Fig: PGD} 
		\end{minipage}
	}
	\caption{Rank variation of adjacency matrix of Polblogs.}
	\label{Fig: MatrixRankPolblogs}
\end{figure}

\begin{figure} [tbhp]
	\centering
	\subfigure[Rank growth under attacks]{
		\begin{minipage}[t]{0.45\columnwidth}
			\centering
			\includegraphics[width=2.0in, height=1.45in]{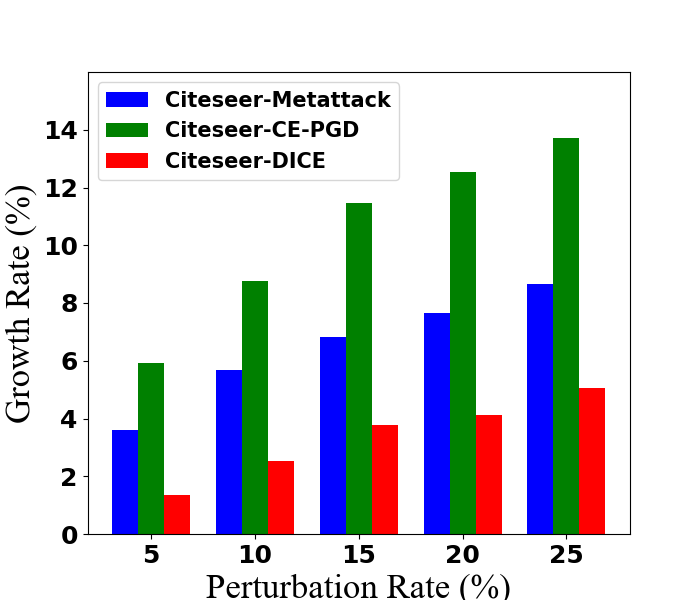}
			\label{Fig: Metattack}
		\end{minipage}
	}
	\vspace{0.02cm}
	\subfigure[Rank decrease by EquiliRes]{
		\begin{minipage}[t]{0.45\columnwidth}
			\centering
			\includegraphics[width=2.0in, height=1.45in]{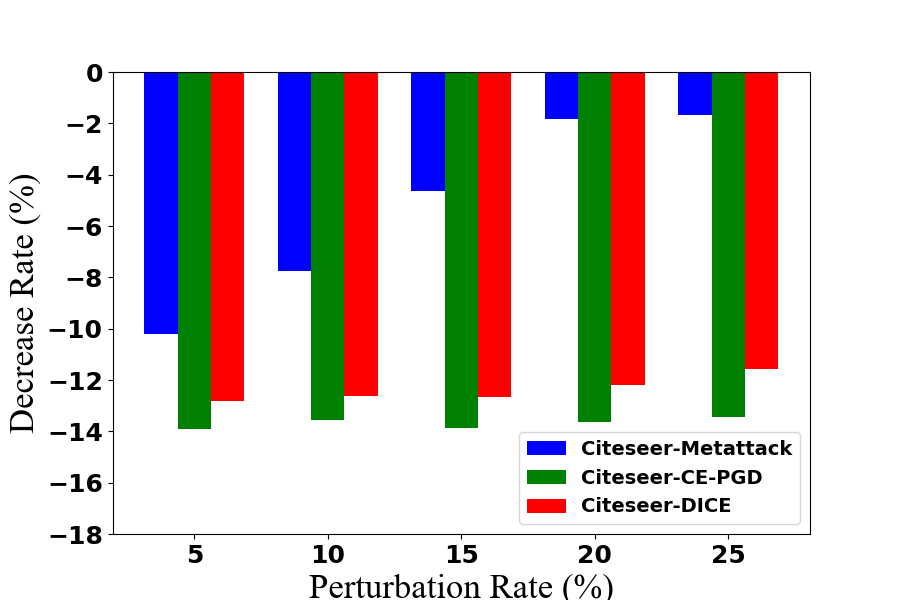}
			\label{Fig: PGD} 
		\end{minipage}
	}
	\caption{Rank variation of adjacency matrix of Citeseer.}
	\label{Fig: MatrixRankCiteseer}
\end{figure}

The recent research points out the GAAs could enlarge the rank of adjacency matrix quickly, thus, to validate the effectiveness of our EquiliRes, we produce two groups of comparative experiments under the three adversarial attacks, as shown in Fig. \ref{Fig: MatrixRankPolblogs} on Polblogs and Fig. \ref{Fig: MatrixRankCiteseer} on Citeseer, to observe the variation of adjacency matrix rank before and after our defense method is engaged. 
			
Seen from the experimental results, we have the following observations: i) as the APR increases, the rank value goes up gradually under the three adversarial attacks on dataset Polblogs, especially with visible lift for Citeseer dataset, this phenomenon shows that graph adversarial attack renders the rank growth to establish connection between dissimilar nodes; ii) correspondingly, our EquiliRes enables to dramatically decline the rank to remove the adversarial edges; and iii) the rank-declining ratio by our proposed EquiliRes has a descending tendency as the perturbation strength becomes larger, this matches the experimental results in Tables \ref{Table: Polblogs}-\ref{Table: EquiliResCombine-1stPart} and Tables \ref{Table: CiteseerUp}-\ref{Table: EquiliResCombine-2ndPart} , reflecting the accuracy of node-classification task would become lower under the gradually growing adversarial perturbations.   
			
\subsection{Singular-value variation} \label{Sec:SingularValue}
\begin{figure*} [hbtp]
\centering
\subfigure[S-V (Metattack)]{
	\begin{minipage}[t]{0.3\columnwidth}
		\centering
		\includegraphics[width=1.7in]{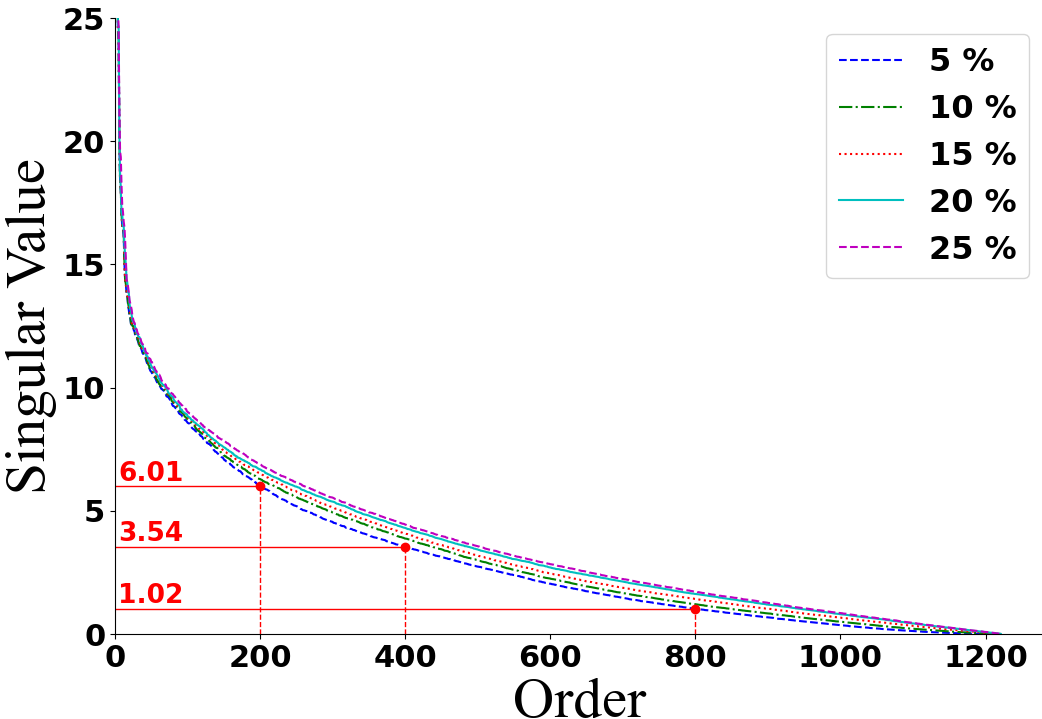}
		\label{Fig: Metattack}
	\end{minipage}
}
\vspace{0.02cm}
\subfigure[S-V (CE-PGD)]{
	\begin{minipage}[t]{0.3\columnwidth}
		\centering
		\includegraphics[width=1.7in]{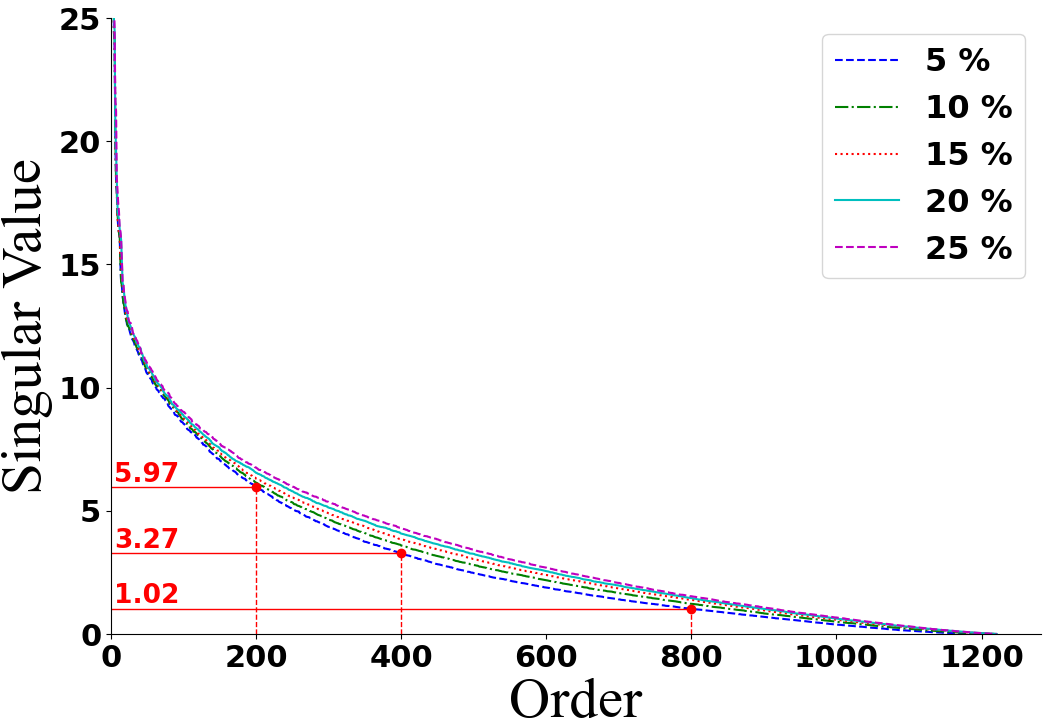}
		\label{Fig: PGD} 
	\end{minipage}
}
	\vspace{0.02cm}
	\subfigure[S-V (DICE)]{
			\begin{minipage}[t]{0.3\columnwidth}
					\centering
					\includegraphics[width=1.7in]{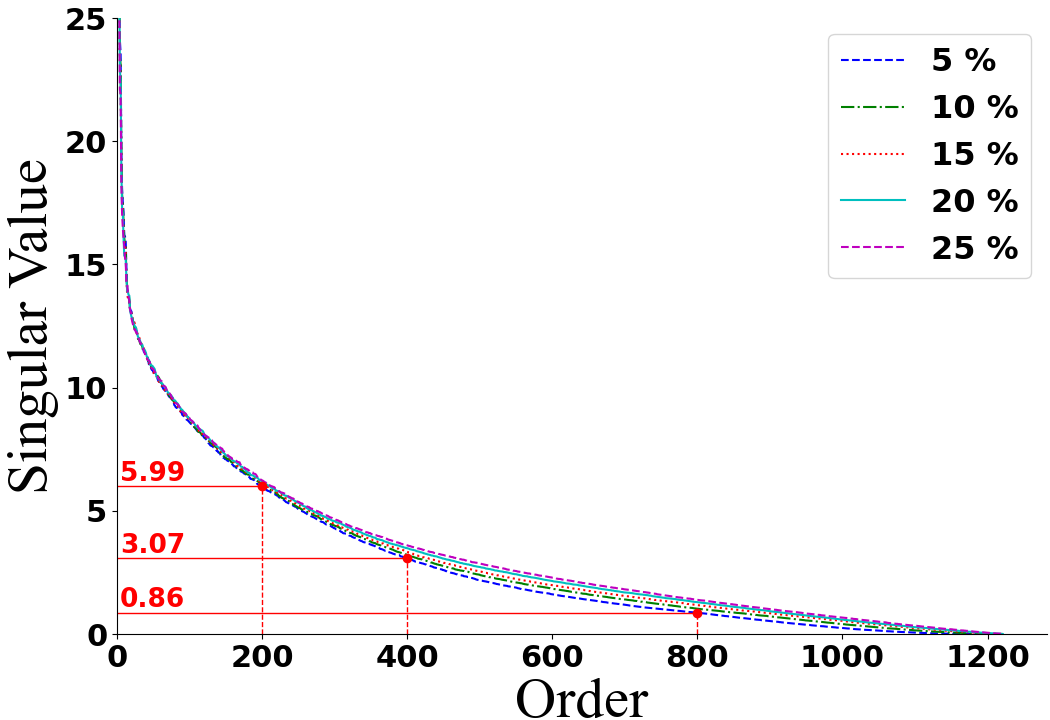}
					\label{Fig: PGD} 
				\end{minipage}
		}
\vspace{0.02cm}
\subfigure[S-V decrease (Metattack)]{
	\begin{minipage}[t]{0.3\columnwidth}
		\centering
		\includegraphics[width=1.7in]{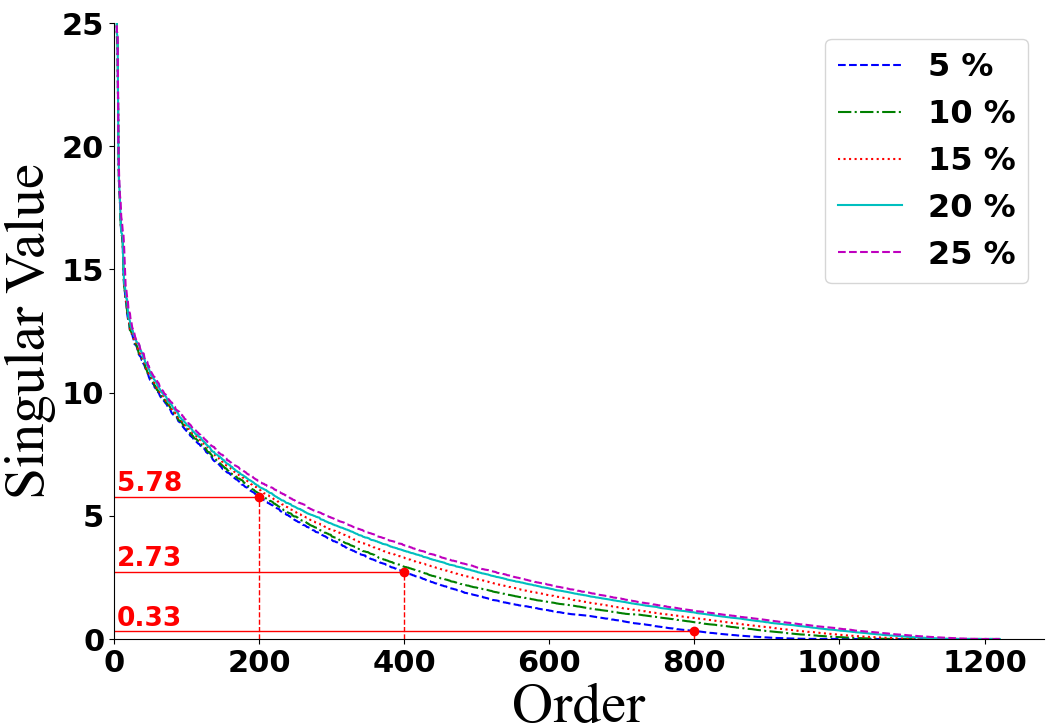}
		\label{Fig: PGD} 
	\end{minipage}
}
\vspace{0.02cm}
\subfigure[S-V decrease (CE-PGD)]{
	\begin{minipage}[t]{0.3\columnwidth}
		\centering
		\includegraphics[width=1.7in]{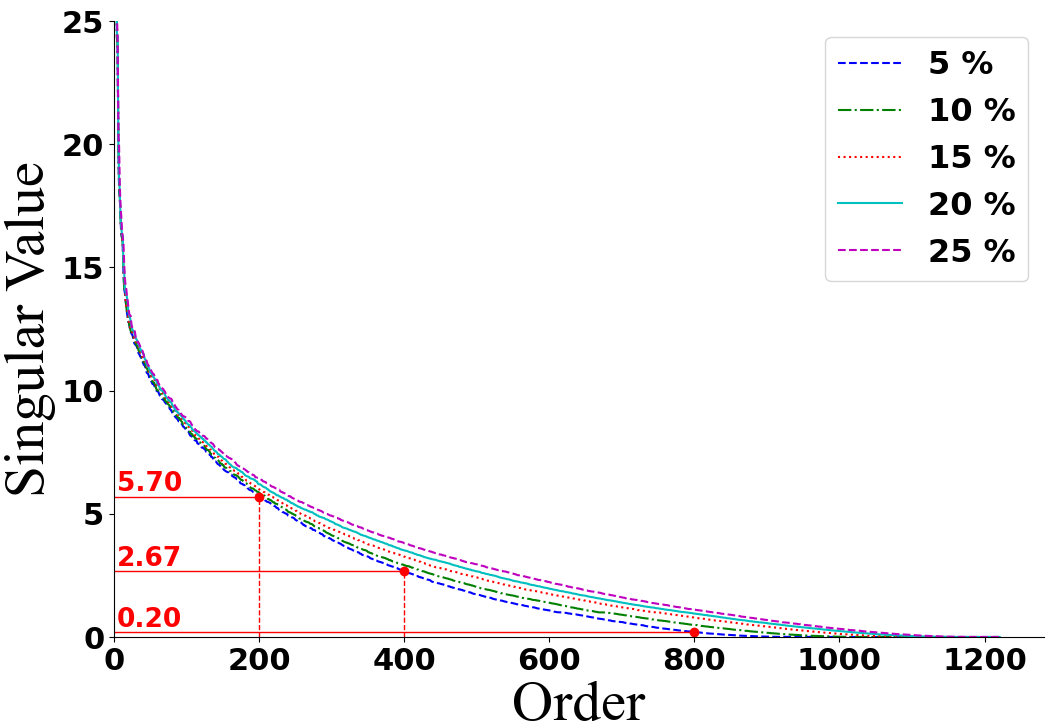}
		\label{Fig: PGD} 
	\end{minipage}
}
	\vspace{0.02cm}
	\subfigure[S-V decrease (DICE)]{
			\begin{minipage}[t]{0.3\columnwidth}
					\centering
					\includegraphics[width=1.7in]{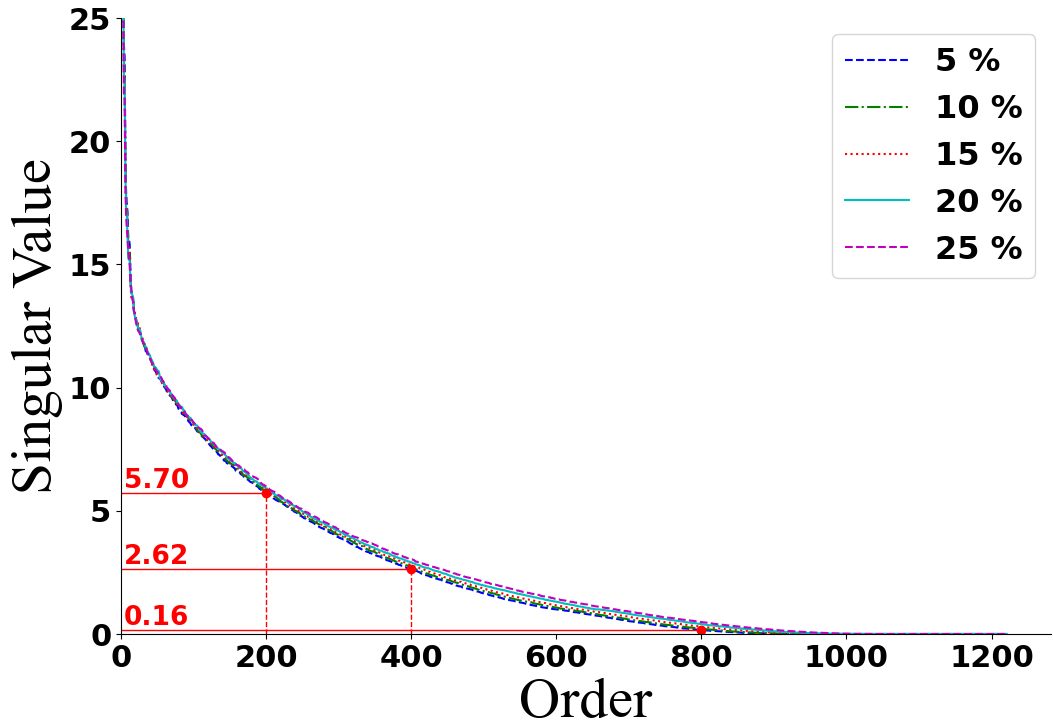}
					\label{Fig: PGD} 
				\end{minipage}
		}
\caption{Singular-Value (S-V) variation of Polblogs with EquiliRes.}
\label{Fig: MatrixSingularPolblogs}
\end{figure*}

\begin{figure*} [hbtp]
\centering
\subfigure[S-V (Metattack)]{
	\begin{minipage}[t]{0.3\columnwidth}
		\centering
		\includegraphics[width=1.8in]{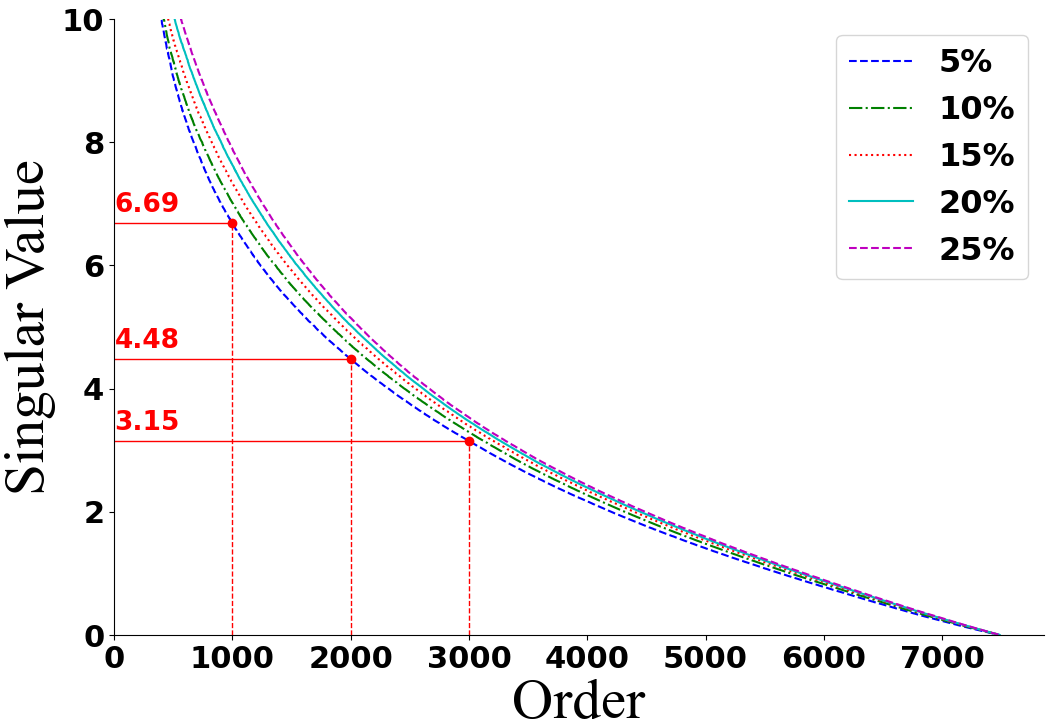}
		\label{Fig: Metattack}
	\end{minipage}
}
\vspace{0.02cm}
\subfigure[S-V (CE-PGD)]{
	\begin{minipage}[t]{0.3\columnwidth}
		\centering
		\includegraphics[width=1.8in]{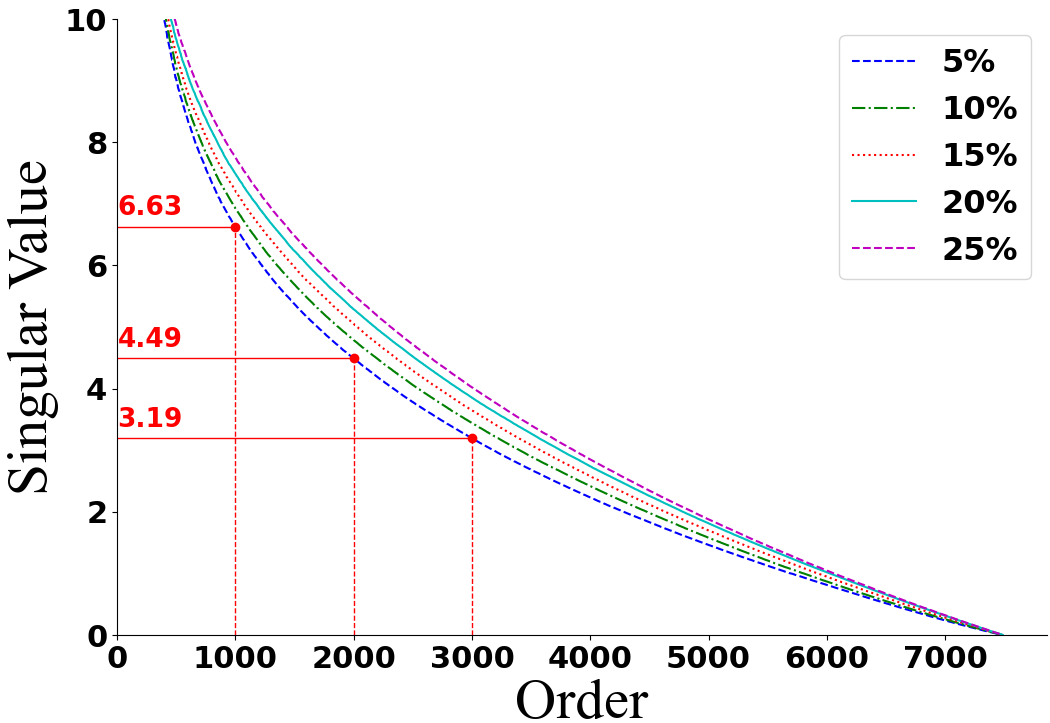}
		\label{Fig: PGD} 
	\end{minipage}
}
	\vspace{0.02cm}
	\subfigure[S-V (DICE)]{
			\begin{minipage}[t]{0.3\columnwidth}
					\centering
					\includegraphics[width=1.8in]{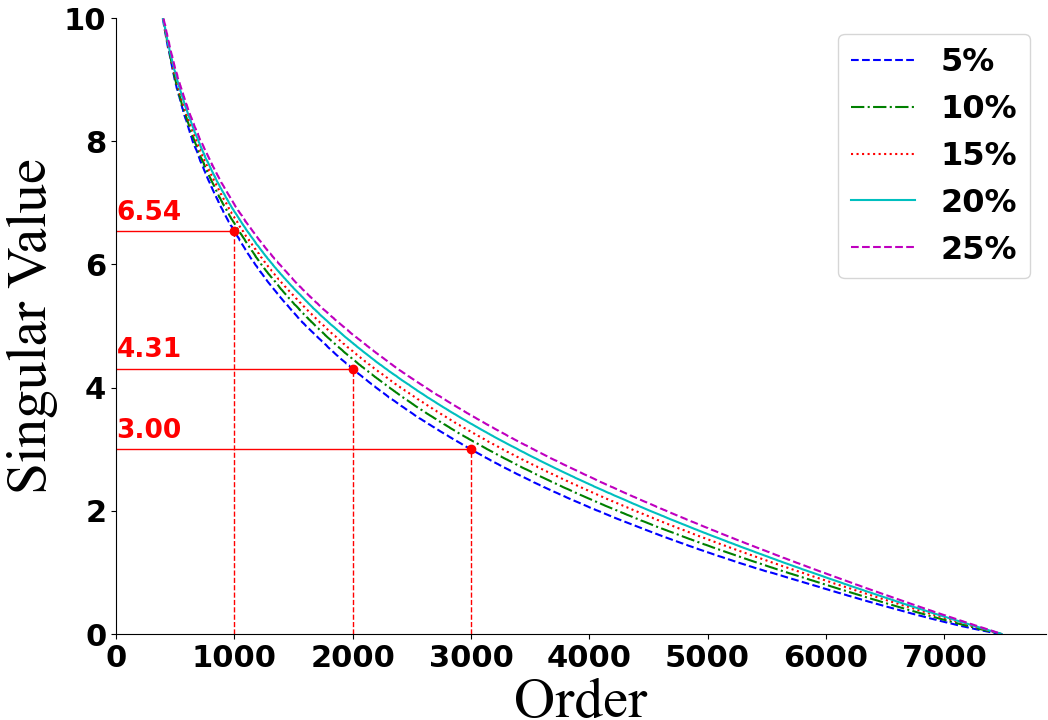}
					\label{Fig: PGD} 
				\end{minipage}
		}
\vspace{0.02cm}
\subfigure[S-V decrease (Metattack)]{
	\begin{minipage}[t]{0.3\columnwidth}
		\centering
		\includegraphics[width=1.8in]{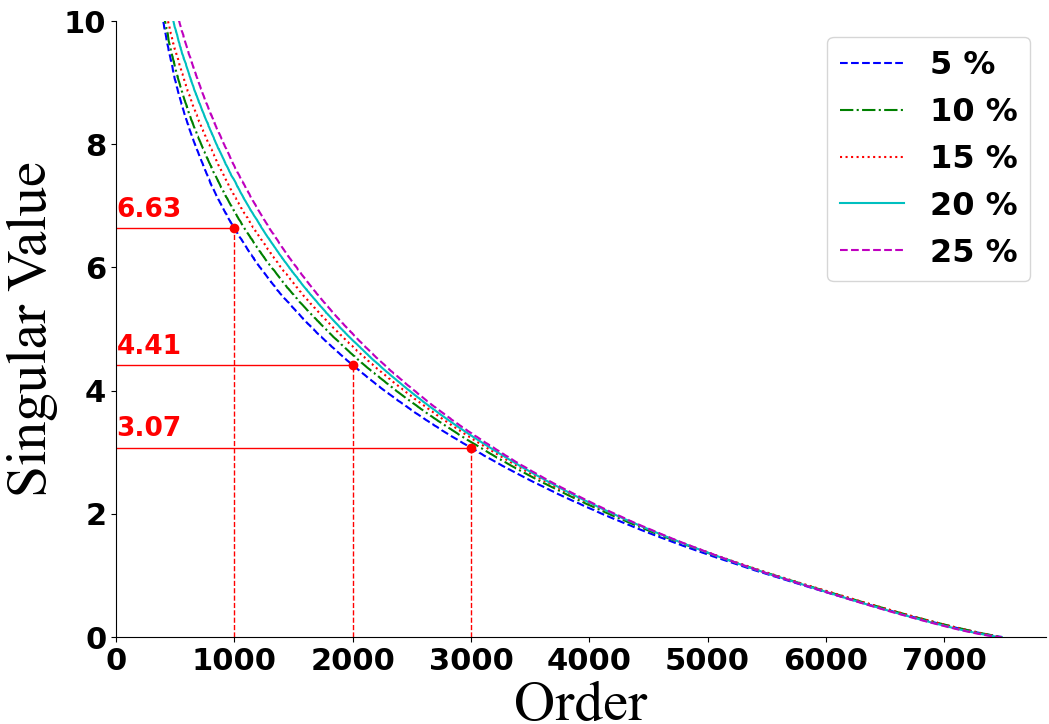}
		\label{Fig: PGD} 
	\end{minipage}
}
\vspace{0.02cm}
\subfigure[S-V decrease (CE-PGD)]{
	\begin{minipage}[t]{0.3\columnwidth}
		\centering
		\includegraphics[width=1.8in]{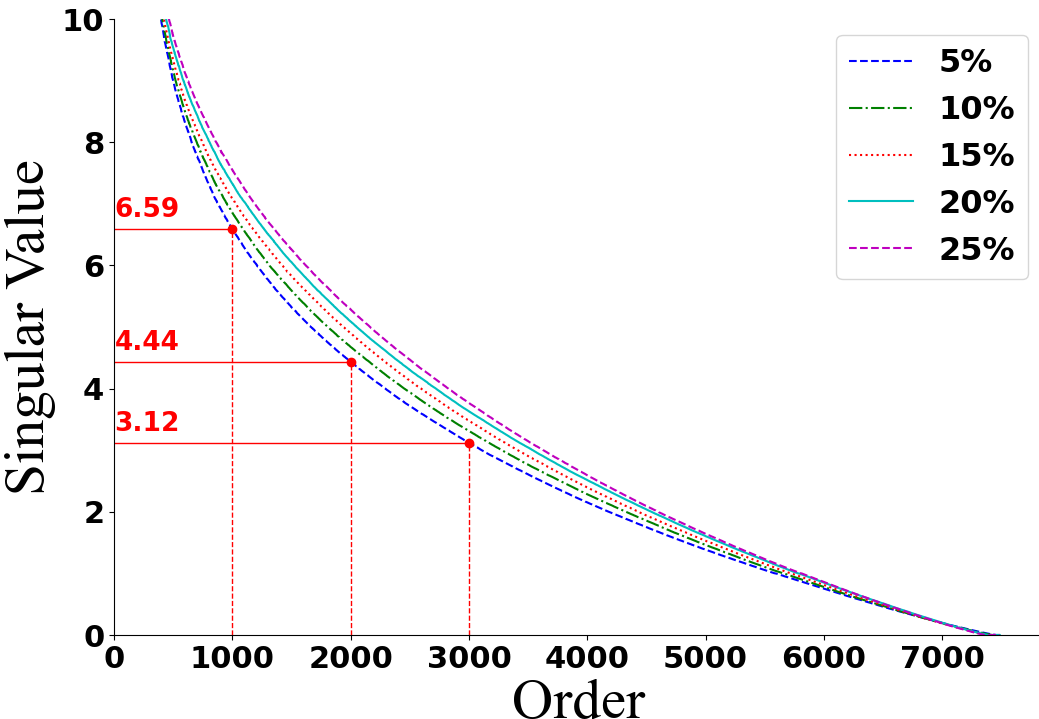}
		\label{Fig: PGD} 
	\end{minipage}
}
	\vspace{0.02cm}
	\subfigure[S-V decrease (DICE)]{
			\begin{minipage}[t]{0.3\columnwidth}
					\centering
					\includegraphics[width=1.8in]{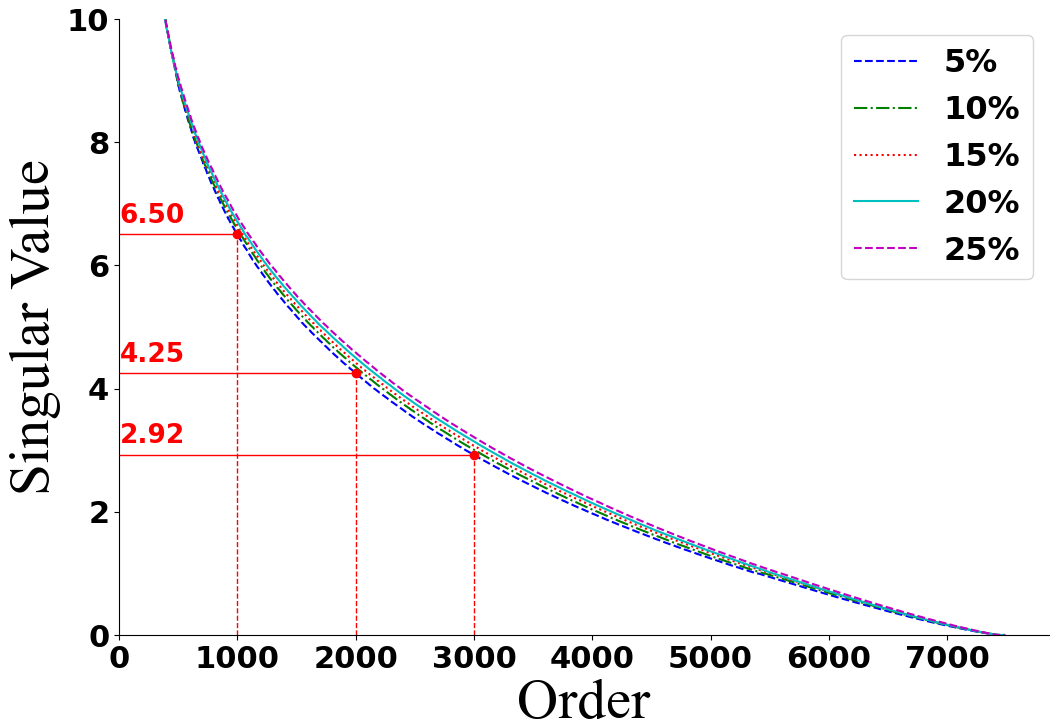}
					\label{Fig: PGD} 
				\end{minipage}
		}
\caption{Singular-Value (S-V) variation of Amazon Photo with EquiliRes.}
\label{Fig: MatrixSingularPhoto}
\end{figure*}

The investigation \cite{JinMa20} discloses the attack Metattack can enlarge the singular values of adjacency matrix, thus we run a set of experiments to showcase the variation of singular values of adjacency matrix after using our EquiliRes on moderate-scale Polblogs and large-scale Amazon Photo datasets, and depict the results in Fig. \ref{Fig: MatrixSingularPolblogs} and Fig. \ref{Fig: MatrixSingularPhoto}, from which the following observations can be obtained. 

First, our EquiliRes indeed reduces the singular values of adjacency matrices under the three adversarial attacks on both datasets, which indicates our proposed equilibrium-point based modification fashion is reasonable and valid to remove adversarial edges w.r.t. different-level adversarial perturbations. In other words, the process of modifying the adjacency matrix to approach to the equilibrium state exactly corresponds to the manipulation of declining singular values. Second, different types of graph adversarial attacks with different attack strengths cause the singular values to have different-level lifts. Correspondingly, our EquiliRes also results in different-level descent in parallel.  
			
\subsection{Computation overhead} \label{Sec:computationOverhead}
\begin{figure} [tbhp]
	\centering
	\subfigure[Polblogs]{
		\begin{minipage}[t]{0.98\columnwidth}
			\centering
			\includegraphics[width=5.4in]{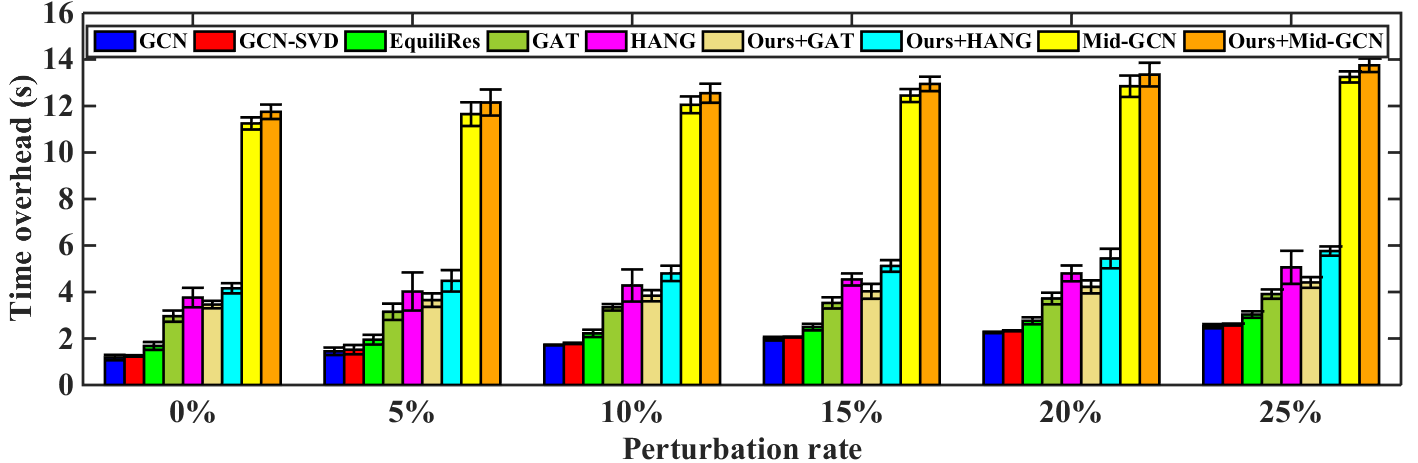}
		\end{minipage}
	}
	\vspace{0.01cm}
	\subfigure[Cora\_ML]{
		\begin{minipage}[t]{0.98\columnwidth}
			\centering
			\includegraphics[width=5.4in]{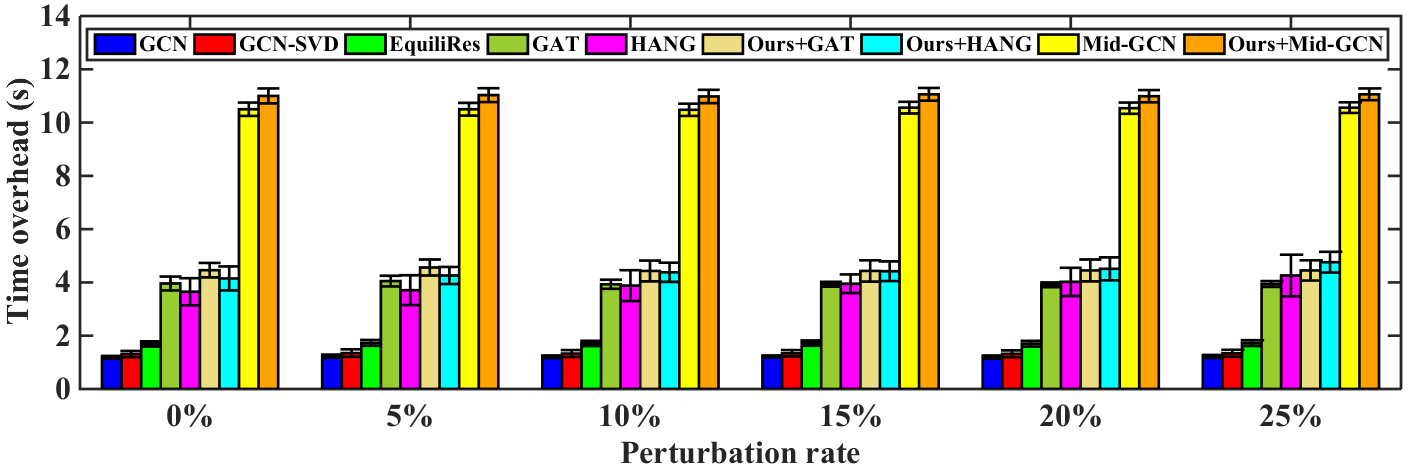}
		\end{minipage}
	}
	\vspace{0.01cm}
	\subfigure[Cora]{
		\begin{minipage}[t]{0.98\columnwidth}
			\centering
			\includegraphics[width=5.4in]{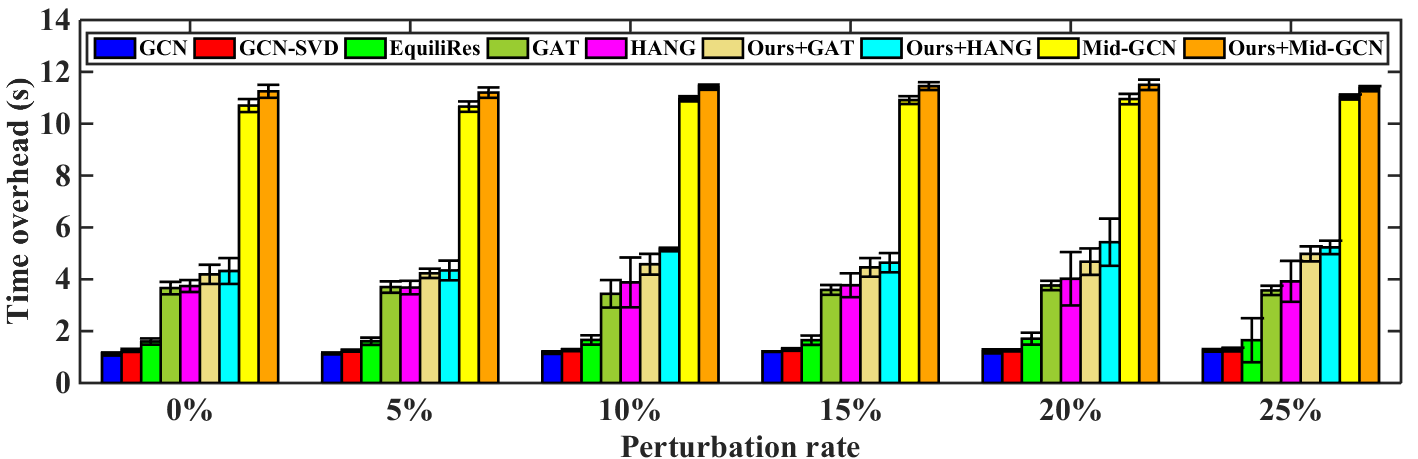}
		\end{minipage}
	}
	\vspace{0.01cm}
    \subfigure[Citeseer]{
	\begin{minipage}[t]{0.98\columnwidth}
		\centering
		\includegraphics[width=5.4in]{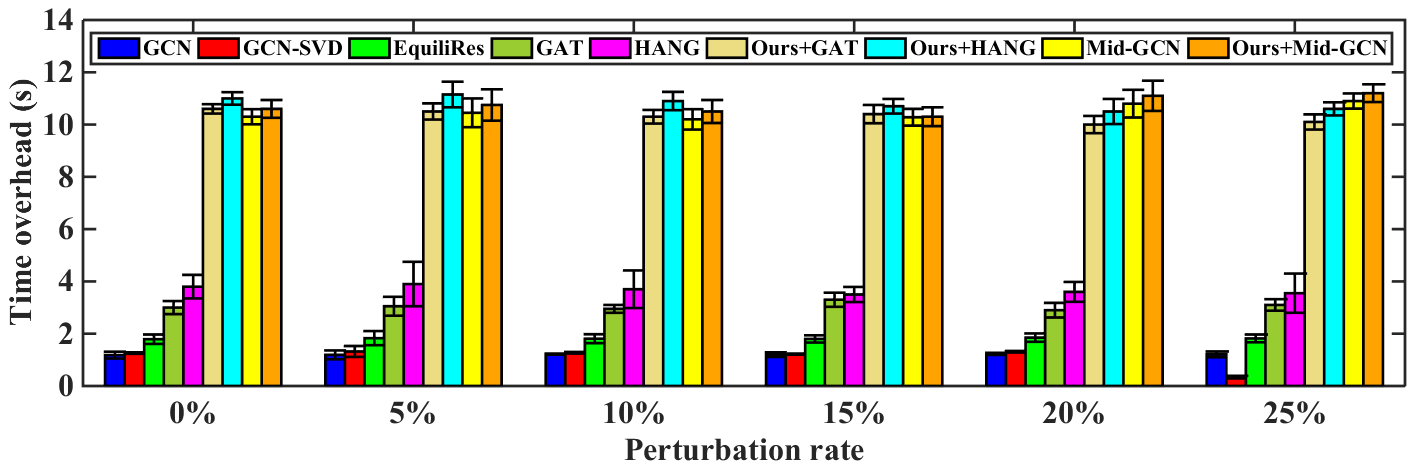}
	\end{minipage}
   }
	\caption{Computation overhead under CE-PGD.}
	\label{Fig: TimeOverheadunderCEPGD}
\end{figure}

One advantages of our EquiliRes lies in post-learning that can be obtained from Laplace Transfer without executing deep-learning manipulation in reality, i.e., the critical state of adversarial resilience of graph regime, stemming from the theoretic guarantee of asymptotic stability. Furthermore, our method employs the condensed one-dimensional function to sketch the entire graph' dynamics via degree centrality, rather than other sophisticated multi-dimensional functions, which also decreases the running time to a large extent. To comparatively analyze the efficiency, we calculate the time overhead from the training input to defense output with the experiment settings: APR is set from 0\% to 25\%, the designated running epoch is set as 200 rounds, and the time overhead is counted on average for 5 times for each experiment. The results on the four datasets Polblogs, Cora\_Ml, Cora, and Citeseer under the attacks CE-PGD are plotted in Fig. \ref{Fig: TimeOverheadunderCEPGD}.

\begin{figure} [tbhp]
	\centering
	\subfigure[Polblogs]{
		\begin{minipage}[t]{0.45\columnwidth}
			\centering
			\includegraphics[width=2.2in, height=1.4in]{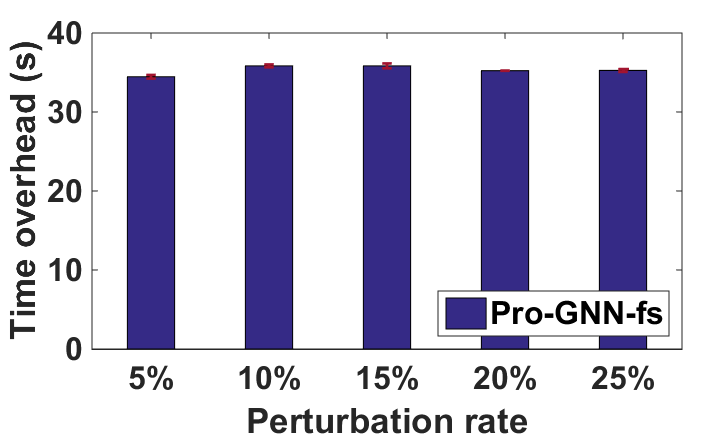}
		\end{minipage}
	}
	\vspace{0.01cm}
	\subfigure[Cora\_ML]{
		\begin{minipage}[t]{0.45\columnwidth}
			\centering
			\includegraphics[width=2.2in, height=1.4in]{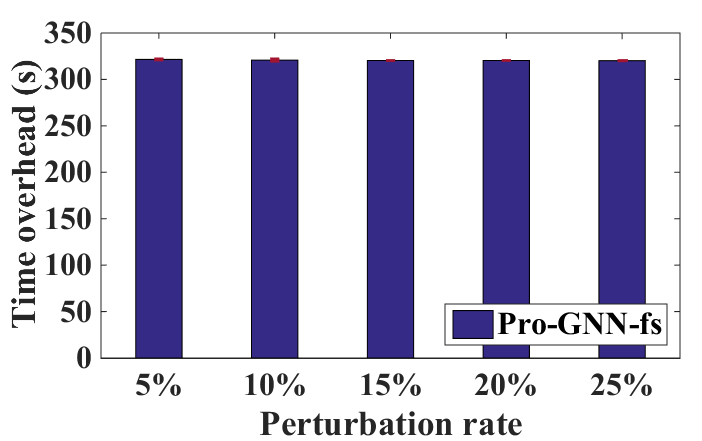}
		\end{minipage}
	}
	\vspace{0.01cm}
	\subfigure[Cora]{
		\begin{minipage}[t]{0.45\columnwidth}
			\centering
			\includegraphics[width=2.2in, height=1.4in]{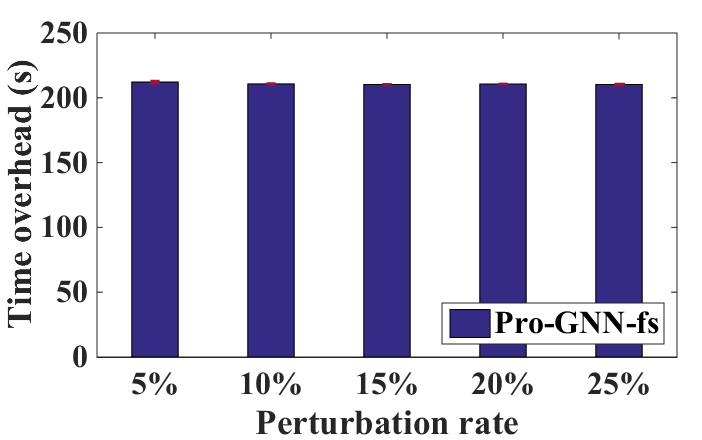}
		\end{minipage}
	}
	\subfigure[Citeseer]{
	\begin{minipage}[t]{0.45\columnwidth}
		\centering
		\includegraphics[width=2.2in, height=1.4in]{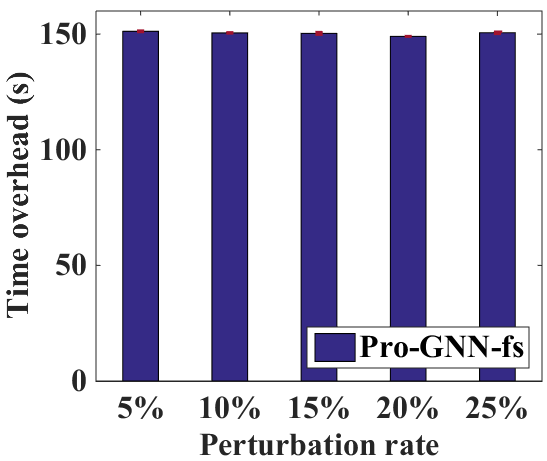}
	\end{minipage}
}
	\caption{Time overhead of Pro-GNN-fs under CE-PGD.}
	\label{Fig: Pro-GNN-fs-TimeOverheadunderCEPGD}
\end{figure}
								
From the experimental results, we can observe that: i) compared to GAT, HANG, Ours+GAT, and Ours+HANG, the basic GCN, GCN-SVD and our EquiliRes take less time cost on the datasets, this is because GCN only needs to capture the graph structure information without other handling. The singular value decomposition makes GCN-SVD cost similar or a little more overhead than GCN, this is because the decomposition only causes the graph to remove some edges. For our EquiliRes, due to the extra location on equilibrium points, thus it costs a little more than GCN-SVD; 
ii) Compared with other methods, Mid-GCN and Ours+Mid-GCN consume more time remarkably, we think this is because they need capture the mid-frequency signals from the high-frequency and low-frequency signals through leveraging a filter for graphs, which costs relatively-long time;  
and iii) although Ours+GAT, Ours+HANG, and Ours+Mid-GCN need a little more overhead than respective individuals, nevertheless, the introduction of our method would not bring too much cost because of the pre-location of inferred equilibrium-point in theory. Overall, these methods are all lightweight compared to those deep learning-based adjacency-matrix iterative optimization mechanisms, such as Pro-GNN-fs \cite{JinMa20}.    
As shown in Fig. \ref{Fig: Pro-GNN-fs-TimeOverheadunderCEPGD}, Pro-GNN-fs costs much more time than the others, deriving from the repeatedly search for the optimal low-rank adjacency matrix. For instance, the maximum overhead of Pro-GNN-fs exceeds 300 seconds on dataset Cora\_ML, while the maximum of the others is no more than 14 seconds. Comparatively analyzing from the two groups of computation overhead, we know that all these adjacency matrix-based and mid-frequency signal-captured defense approaches cost much less time than the deep learning-based adjacency-matrix iterative optimization methods. 
			
\section{Related work} \label{RelatedWork}
\subsection{Graph adversarial attack}
Currently, extensive research has witnessed the graph data is suffering from severe adversarial attacks \cite{WangNeil19, LiuSi19, Bojchevski19}, partially deriving from the sophisticated dependency (links) among graph nodes. For instance, from the viewpoint of global graph, the work \cite{WangNeil19} manipulates graph structure towards the collective classification question. Metattack \cite{Daniel19} leverages meta-learning to strategically attack the edges, in addition to modify weight, and the reference \cite{Daniel19} reported that Metattack could quickly boost the rank of adjacency matrix, and tended to establish connections between dissimilar nodes. Xu et al. \cite{XuChen19} employ negative cross entropy-based PGD (CE-PGD) and CW attack-based PGD (CW-PGD) to generate topology perturbations. 

More seriously, towards the training graph data, even inaccessible, some crucial properties (node degree, subgraph structure) still can be explored by property inference attack (PIA) \cite{WangWendy22} and structure membership inference attack (SMIA) \cite{WangWendy24}. He et al. \cite{HeMichael21} propose how to steal graph links from training dataset given a non-access scenario to GNN model. Generally speaking, large graph regime can be usually deemed as a sophisticated system, the complexity of interaction (information-passing) fashion among a large number of nodes makes it difficult to defend against continuous and imperceivable adversarial perturbations, especially for large-scale graph regimes. 
\subsection{Graph adversarial defense}
At present, the adversarial defense is studied mainly from two lines: graph per se and graph neural flow. The former mainly improves the adversarial resilience through modifying topology and node feature, for example, from the preprocessing perspective, Wu et al. \cite{Chamberlain21} firstly remove the dissimilar edges resorting to node feature-based Jaccard-similarity comparison with a preset threshold. GCN-SVD \cite{Chamberlain21}, serving as a preprocessing means as well, utilizes SVD to decompose adjacency matrix of the perturbed graph and acquires a low-rank adjacency-matrix approximation to represent a cleaner graph. Pro-GCN-fs \cite{JinMa20} tries to learn a low-rank and close adjacency matrix to the original. Obviously, the above methods all aim to modify the graph topology and/or features to promote adversarial resilience. Our work falls into this category due to the consideration of finding out the intrinsic critical state of adversarial resilience through modifying topology. More concretely, our work only focuses on graph structure without considering node feature.

The other line, i.e. neural ODE network, aims to improve the adversarial resilience through studying the graph neural flow (GNF), that is, neural network constrains the input and output to adhere to certain physical laws, and the transmission of such constrained data within the neural network is referred to as GNF. Typically, these constraints are implemented through system equations from physics. In other words, GNNs can be instantiated with system equations to learn graph data. In recent years, the neural ODE has been successfully applied to GNN through modeling information exchange. The works \cite{Chamberlain21, Thorpe22} deem message-passing process as the heat diffusion model, while the literature \cite{ChamberlainJames21, SongKang22} regards the message-passing process as Beltrami diffusion. In addition, towards graph nodes, Rusch et al. \cite{Rusch22} adopt the oscillators to model nodes and conduct the message-passing process under the guidance of a coupled oscillating ODE. Recently, HANG \cite{ZhaoKang23} imposes an energy-conservation constraint on GNF, endowed with Lyapunov stability, to enhance the adversarial robustness of GNNs. Differently, our approach models graph adversarial perturbation by adhering to the asymptotic stability of Lyapunov theory, simulating the graph state change under adversarial attacks. By observing the phenomena of dynamic variation, we propose a preprocessing graph optimization method tailored for graph data through resorting to equilibrium-point theory in dynamic systems. This category of GNF-based defense methods is designed to construct graph neural flows for GNNs, while our work is oriented toward graph data processing. 


\section{Motivation and open issues} \label{Sec: OpenIssue}
\textbf{Motivation.} Towards why we need dynamic-system findings to study adversarial learning, We think there are at least three reasons: i) the oscillation in dynamic system and the perturbation in adversarial learning have common characteristic-dynamics. Considering the difficulty of measuring adversarial attacks (To our best knowledge, no solution to quantify adversarial attacks to date), we can employ Laplace Transform to measure adversarial attack quantitatively, especially for the scenarios where we have no idea about the behavior of GAAs. Hence, resorting to the findings in dynamic system to bridge connection to adversarial learning is proper and worth in-depth exploration, and Fig. \ref{Fig: PolblogsTrajec}-\ref{Fig:AmazonPhotoTrajec} show the consistency between the equilibrium-point trajectory and adversarial attack effect; ii) using the stability theory in dynamic systems, we can infer an optimal robust (equilibrium-point) state in theory for each graph regime in a simple yet valid manner; and iii) in the light of equilibrium point, we can directly purify the perturbed adjacency matrix in a light-weight way, resulting in a lower time overhead compared to the time-consuming model-retraining methods (e.g. Pro-GNN-fs) as shown in Fig. \ref{Fig: TimeOverheadunderCEPGD} and Fig. \ref{Fig: Pro-GNN-fs-TimeOverheadunderCEPGD}. 

\textbf{Open Issues.} Our work is completely a new attempt to study graph adversarial resilience, and the following issues are well worthy discussing: i) \textbf{Condensed One-Dimensional Mapping Function.} In our work, a simple condensed one-dimensional function is developed to map the dynamics of graph regime as a whole, in fact, two-dimensional mapping or multi-dimensional mapping can be explored as well for distinct graph regimes, this is because our proposed asymptotically-stable theoretic framework fully supports multi-dimensional dynamic-variation function. The more accurate the modeling on GAAs, the higher the adversarial resilience would be; ii) \textbf{Adversarial Resilience on Multi-Modal Data.} This paper only concentrates graph data, a natural question emerges, i.e. whether other modal data also has such equilibrium state to maintain the adversarial resilience, such as image, audio, text, etc. If exist, whether we can tackle in the same way in technique given the consideration of totally different data structures; and iii) \textbf{Parameters of Adversarial Resilience-Declining Function.} As shown in Fig. \ref{Fig: PolblogsTrajec}-\ref{Fig:AmazonPhotoTrajec}, different graph regimes usually have distinct adversarial resilience-declining tendencies under different-level perturbations, hence, adequate parameters are needed to well-fit the variation of adverseness by various graph adversarial attacks. Only the well-fitting between theoretical equilibrium-point trajectory and adversarial-effect dots can truthfully unveil the behavior of adversarial attack.

\end{document}